\documentclass[12pt]{article}
\usepackage{restrained,ling1}
\usepackage{fullpage,times,mathptm,latexsym,doublespace,epsfig,apalike}

\makeatletter
\long\def\@footnotetext#1{\insert\footins{\def\baselinestretch{1}\footnotesize
    \interlinepenalty\interfootnotelinepenalty 
    \splittopskip\footnotesep
    \splitmaxdepth \dp\strutbox \floatingpenalty \@MM
    \hsize\columnwidth \@parboxrestore
   \edef\@currentlabel{\csname p@footnote\endcsname\@thefnmark}\@makefntext
    {\rule{\z@}{\footnotesep}\ignorespaces
      #1\strut}}}
\newcommand{\dspftnt}[1]{\footnote{#1}}
\makeatother

\newcommand{\term}[1]{\textsc{#1}}
\newcommand{\ling}[1]{\textit{#1}}

\newcommand{\m}[1]{\mbox{\textit{#1}}}
\newcommand{\spud}{\mbox{\textsc{spud}}}
\newcommand{\speaker}{\mbox{[\textsc{s}]}}
\newcommand{\user}{\mbox{[\textsc{u}]}}
\newcommand{\shared}{\mbox{[\textsc{cr}]}}
\newcommand{\meanings}{\mbox{[\textsc{mp}]}}
\newcommand{\sharedinst}[1]{\mbox{\underline{\ensuremath{#1}}}}
\newcommand{\goalinst}[1]{\fbox{\ensuremath{#1}}}
\newcommand{\distractors}{\mbox{\em D}}
\newcommand{\entity}{\mbox{\em e}}
\newcommand{\entityA}{\mbox{\em a}}
\newcommand{\entityB}{\mbox{\em b}}

\newcommand{\imp}{\supset}
\newcommand{\proves}{\longrightarrow}
\newcommand{\footnodemark}{\ensuremath{_{\mbox{*}}}}

\newcommand{\lxstructure}[1]{
\begin{tabular}{|l|}
\hline \textbf{Structure:} \\
\hline #1 \\
\hline
\end{tabular}}
   
\newcommand{\alinks}[1]{
\begin{tabular}{|l|}
\hline \textbf{Assert:} \\
\hline #1 \\
\hline
\end{tabular}}

\newcommand{\pslinks}[1]{
\begin{tabular}{|l|}
\hline \textbf{Presuppose:} \\
\hline  #1 \\
\hline
\end{tabular}}

\newcommand{\pglinks}[1]{
\begin{tabular}{|l|}
\hline \textbf{Pragmatics:} \\
\hline #1 \\
\hline
\end{tabular}}

\newcommand{\thingstodo}[1]{
\begin{tabular}{|l|}
\hline \textbf{Requirements:} \\
\hline #1 \\
\hline
\end{tabular}}

\newcommand{\lxstructuret}[1]{
\begin{tabular}{|l|}
\hline \textbf{Structure:} \\
\hline \\[-1ex] #1 \\[3ex]
\hline
\end{tabular}}
   
\newcommand{\alinkst}[1]{
\begin{tabular}{|l|}
\hline \textbf{Assert:} \\
\hline \\[-1ex] #1 \\[3ex]
\hline
\end{tabular}}

\newcommand{\pslinkst}[1]{
\begin{tabular}{|l|}
\hline \textbf{Presuppose:} \\
\hline \\[-1ex] #1 \\[3ex]
\hline
\end{tabular}}

\newcommand{\pglinkst}[1]{
\begin{tabular}{|l|}
\hline \textbf{Pragmatics:} \\
\hline \\[-1ex] #1 \\[3ex]
\hline
\end{tabular}}

\newcounter{treecount}
\newcounter{branchcount}
\newsavebox{\parentbox}
\newsavebox{\treebox}
\newsavebox{\treeboxone}
\newsavebox{\treeboxtwo}
\newsavebox{\treeboxthree}
\newsavebox{\treeboxfour}
\newsavebox{\treeboxfive}
\newsavebox{\treeboxsix}
\newsavebox{\treeboxseven}
\newsavebox{\treeboxeight}
\newsavebox{\treeboxnine}
\newsavebox{\treeboxten}
\newsavebox{\treeboxeleven}
\newsavebox{\treeboxtwelve}
\newsavebox{\treeboxthirteen}
\newsavebox{\treeboxfourteen}
\newsavebox{\treeboxfifteen}
\newsavebox{\treeboxsixteen}
\newsavebox{\treeboxseventeen}
\newsavebox{\treeboxeighteen}
\newsavebox{\treeboxnineteen}
\newsavebox{\treeboxtwenty}
\newlength{\treeoffsetone}
\newlength{\treeoffsettwo}
\newlength{\treeoffsetthree}
\newlength{\treeoffsetfour}
\newlength{\treeoffsetfive}
\newlength{\treeoffsetsix}
\newlength{\treeoffsetseven}
\newlength{\treeoffseteight}
\newlength{\treeoffsetnine}
\newlength{\treeoffsetten}
\newlength{\treeoffseteleven}
\newlength{\treeoffsettwelve}
\newlength{\treeoffsetthirteen}
\newlength{\treeoffsetfourteen}
\newlength{\treeoffsetfifteen}
\newlength{\treeoffsetsixteen}
\newlength{\treeoffsetseventeen}
\newlength{\treeoffseteighteen}
\newlength{\treeoffsetnineteen}
\newlength{\treeoffsettwenty}

\newlength{\treeshiftone}
\newlength{\treeshifttwo}
\newlength{\treeshiftthree}
\newlength{\treeshiftfour}
\newlength{\treeshiftfive}
\newlength{\treeshiftsix}
\newlength{\treeshiftseven}
\newlength{\treeshifteight}
\newlength{\treeshiftnine}
\newlength{\treeshiftten}
\newlength{\treeshifteleven}
\newlength{\treeshifttwelve}
\newlength{\treeshiftthirteen}
\newlength{\treeshiftfourteen}
\newlength{\treeshiftfifteen}
\newlength{\treeshiftsixteen}
\newlength{\treeshiftseventeen}
\newlength{\treeshifteighteen}
\newlength{\treeshiftnineteen}
\newlength{\treeshifttwenty}
\newlength{\treewidthone}
\newlength{\treewidthtwo}
\newlength{\treewidththree}
\newlength{\treewidthfour}
\newlength{\treewidthfive}
\newlength{\treewidthsix}
\newlength{\treewidthseven}
\newlength{\treewidtheight}
\newlength{\treewidthnine}
\newlength{\treewidthten}
\newlength{\treewidtheleven}
\newlength{\treewidthtwelve}
\newlength{\treewidththirteen}
\newlength{\treewidthfourteen}
\newlength{\treewidthfifteen}
\newlength{\treewidthsixteen}
\newlength{\treewidthseventeen}
\newlength{\treewidtheighteen}
\newlength{\treewidthnineteen}
\newlength{\treewidthtwenty}
\newlength{\daughteroffsetone}
\newlength{\daughteroffsettwo}
\newlength{\daughteroffsetthree}
\newlength{\daughteroffsetfour}
\newlength{\branchwidthone}
\newlength{\branchwidthtwo}
\newlength{\branchwidththree}
\newlength{\branchwidthfour}
\newlength{\parentoffset}
\newlength{\treeoffset}
\newlength{\daughteroffset}
\newlength{\branchwidth}
\newlength{\parentwidth}
\newlength{\treewidth}
\newcommand{\ontop}[1]{\begin{tabular}{c}#1\end{tabular}}
\newcommand{\poptree}{%
\ifnum\value{treecount}=0\typeout{QobiTeX warning---Tree stack underflow}\fi%
\addtocounter{treecount}{-1}%
\setlength{\treeoffsettwo}{\treeoffsetthree}%
\setlength{\treeoffsetthree}{\treeoffsetfour}%
\setlength{\treeoffsetfour}{\treeoffsetfive}%
\setlength{\treeoffsetfive}{\treeoffsetsix}%
\setlength{\treeoffsetsix}{\treeoffsetseven}%
\setlength{\treeoffsetseven}{\treeoffseteight}%
\setlength{\treeoffseteight}{\treeoffsetnine}%
\setlength{\treeoffsetnine}{\treeoffsetten}%
\setlength{\treeoffsetten}{\treeoffseteleven}%
\setlength{\treeoffseteleven}{\treeoffsettwelve}%
\setlength{\treeoffsettwelve}{\treeoffsetthirteen}%
\setlength{\treeoffsetthirteen}{\treeoffsetfourteen}%
\setlength{\treeoffsetfourteen}{\treeoffsetfifteen}%
\setlength{\treeoffsetfifteen}{\treeoffsetsixteen}%
\setlength{\treeoffsetsixteen}{\treeoffsetseventeen}%
\setlength{\treeoffsetseventeen}{\treeoffseteighteen}%
\setlength{\treeoffseteighteen}{\treeoffsetnineteen}%
\setlength{\treeoffsetnineteen}{\treeoffsettwenty}%
\setlength{\treeshifttwo}{\treeshiftthree}%
\setlength{\treeshiftthree}{\treeshiftfour}%
\setlength{\treeshiftfour}{\treeshiftfive}%
\setlength{\treeshiftfive}{\treeshiftsix}%
\setlength{\treeshiftsix}{\treeshiftseven}%
\setlength{\treeshiftseven}{\treeshifteight}%
\setlength{\treeshifteight}{\treeshiftnine}%
\setlength{\treeshiftnine}{\treeshiftten}%
\setlength{\treeshiftten}{\treeshifteleven}%
\setlength{\treeshifteleven}{\treeshifttwelve}%
\setlength{\treeshifttwelve}{\treeshiftthirteen}%
\setlength{\treeshiftthirteen}{\treeshiftfourteen}%
\setlength{\treeshiftfourteen}{\treeshiftfifteen}%
\setlength{\treeshiftfifteen}{\treeshiftsixteen}%
\setlength{\treeshiftsixteen}{\treeshiftseventeen}%
\setlength{\treeshiftseventeen}{\treeshifteighteen}%
\setlength{\treeshifteighteen}{\treeshiftnineteen}%
\setlength{\treeshiftnineteen}{\treeshifttwenty}%
\setlength{\treewidthtwo}{\treewidththree}%
\setlength{\treewidththree}{\treewidthfour}%
\setlength{\treewidthfour}{\treewidthfive}%
\setlength{\treewidthfive}{\treewidthsix}%
\setlength{\treewidthsix}{\treewidthseven}%
\setlength{\treewidthseven}{\treewidtheight}%
\setlength{\treewidtheight}{\treewidthnine}%
\setlength{\treewidthnine}{\treewidthten}%
\setlength{\treewidthten}{\treewidtheleven}%
\setlength{\treewidtheleven}{\treewidthtwelve}%
\setlength{\treewidthtwelve}{\treewidththirteen}%
\setlength{\treewidththirteen}{\treewidthfourteen}%
\setlength{\treewidthfourteen}{\treewidthfifteen}%
\setlength{\treewidthfifteen}{\treewidthsixteen}%
\setlength{\treewidthsixteen}{\treewidthseventeen}%
\setlength{\treewidthseventeen}{\treewidtheighteen}%
\setlength{\treewidtheighteen}{\treewidthnineteen}%
\setlength{\treewidthnineteen}{\treewidthtwenty}%
\sbox{\treeboxtwo}{\usebox{\treeboxthree}}%
\sbox{\treeboxthree}{\usebox{\treeboxfour}}%
\sbox{\treeboxfour}{\usebox{\treeboxfive}}%
\sbox{\treeboxfive}{\usebox{\treeboxsix}}%
\sbox{\treeboxsix}{\usebox{\treeboxseven}}%
\sbox{\treeboxseven}{\usebox{\treeboxeight}}%
\sbox{\treeboxeight}{\usebox{\treeboxnine}}%
\sbox{\treeboxnine}{\usebox{\treeboxten}}%
\sbox{\treeboxten}{\usebox{\treeboxeleven}}%
\sbox{\treeboxeleven}{\usebox{\treeboxtwelve}}%
\sbox{\treeboxtwelve}{\usebox{\treeboxthirteen}}%
\sbox{\treeboxthirteen}{\usebox{\treeboxfourteen}}%
\sbox{\treeboxfourteen}{\usebox{\treeboxfifteen}}%
\sbox{\treeboxfifteen}{\usebox{\treeboxsixteen}}%
\sbox{\treeboxsixteen}{\usebox{\treeboxseventeen}}%
\sbox{\treeboxseventeen}{\usebox{\treeboxeighteen}}%
\sbox{\treeboxeighteen}{\usebox{\treeboxnineteen}}%
\sbox{\treeboxnineteen}{\usebox{\treeboxtwenty}}}
\newcommand{\leaf}[1]{%
\ifnum\value{treecount}=20\typeout{QobiTeX warning---Tree stack overflow}\fi%
\addtocounter{treecount}{1}%
\sbox{\treeboxtwenty}{\usebox{\treeboxnineteen}}%
\sbox{\treeboxnineteen}{\usebox{\treeboxeighteen}}%
\sbox{\treeboxeighteen}{\usebox{\treeboxseventeen}}%
\sbox{\treeboxseventeen}{\usebox{\treeboxsixteen}}%
\sbox{\treeboxsixteen}{\usebox{\treeboxfifteen}}%
\sbox{\treeboxfifteen}{\usebox{\treeboxfourteen}}%
\sbox{\treeboxfourteen}{\usebox{\treeboxthirteen}}%
\sbox{\treeboxthirteen}{\usebox{\treeboxtwelve}}%
\sbox{\treeboxtwelve}{\usebox{\treeboxeleven}}%
\sbox{\treeboxeleven}{\usebox{\treeboxten}}%
\sbox{\treeboxten}{\usebox{\treeboxnine}}%
\sbox{\treeboxnine}{\usebox{\treeboxeight}}%
\sbox{\treeboxeight}{\usebox{\treeboxseven}}%
\sbox{\treeboxseven}{\usebox{\treeboxsix}}%
\sbox{\treeboxsix}{\usebox{\treeboxfive}}%
\sbox{\treeboxfive}{\usebox{\treeboxfour}}%
\sbox{\treeboxfour}{\usebox{\treeboxthree}}%
\sbox{\treeboxthree}{\usebox{\treeboxtwo}}%
\sbox{\treeboxtwo}{\usebox{\treeboxone}}%
\sbox{\treeboxone}{\ontop{#1}}%
\sbox{\treeboxone}{\raisebox{-\ht\treeboxone}{\usebox{\treeboxone}}}%
\setlength{\treeoffsettwenty}{\treeoffsetnineteen}%
\setlength{\treeoffsetnineteen}{\treeoffseteighteen}%
\setlength{\treeoffseteighteen}{\treeoffsetseventeen}%
\setlength{\treeoffsetseventeen}{\treeoffsetsixteen}%
\setlength{\treeoffsetsixteen}{\treeoffsetfifteen}%
\setlength{\treeoffsetfifteen}{\treeoffsetfourteen}%
\setlength{\treeoffsetfourteen}{\treeoffsetthirteen}%
\setlength{\treeoffsetthirteen}{\treeoffsettwelve}%
\setlength{\treeoffsettwelve}{\treeoffseteleven}%
\setlength{\treeoffseteleven}{\treeoffsetten}%
\setlength{\treeoffsetten}{\treeoffsetnine}%
\setlength{\treeoffsetnine}{\treeoffseteight}%
\setlength{\treeoffseteight}{\treeoffsetseven}%
\setlength{\treeoffsetseven}{\treeoffsetsix}%
\setlength{\treeoffsetsix}{\treeoffsetfive}%
\setlength{\treeoffsetfive}{\treeoffsetfour}%
\setlength{\treeoffsetfour}{\treeoffsetthree}%
\setlength{\treeoffsetthree}{\treeoffsettwo}%
\setlength{\treeoffsettwo}{\treeoffsetone}%
\setlength{\treeoffsetone}{0.5\wd\treeboxone}%
\setlength{\treeshifttwenty}{\treeshiftnineteen}%
\setlength{\treeshiftnineteen}{\treeshifteighteen}%
\setlength{\treeshifteighteen}{\treeshiftseventeen}%
\setlength{\treeshiftseventeen}{\treeshiftsixteen}%
\setlength{\treeshiftsixteen}{\treeshiftfifteen}%
\setlength{\treeshiftfifteen}{\treeshiftfourteen}%
\setlength{\treeshiftfourteen}{\treeshiftthirteen}%
\setlength{\treeshiftthirteen}{\treeshifttwelve}%
\setlength{\treeshifttwelve}{\treeshifteleven}%
\setlength{\treeshifteleven}{\treeshiftten}%
\setlength{\treeshiftten}{\treeshiftnine}%
\setlength{\treeshiftnine}{\treeshifteight}%
\setlength{\treeshifteight}{\treeshiftseven}%
\setlength{\treeshiftseven}{\treeshiftsix}%
\setlength{\treeshiftsix}{\treeshiftfive}%
\setlength{\treeshiftfive}{\treeshiftfour}%
\setlength{\treeshiftfour}{\treeshiftthree}%
\setlength{\treeshiftthree}{\treeshifttwo}%
\setlength{\treeshifttwo}{\treeshiftone}%
\setlength{\treeshiftone}{0pt}%
\setlength{\treewidthtwenty}{\treewidthnineteen}%
\setlength{\treewidthnineteen}{\treewidtheighteen}%
\setlength{\treewidtheighteen}{\treewidthseventeen}%
\setlength{\treewidthseventeen}{\treewidthsixteen}%
\setlength{\treewidthsixteen}{\treewidthfifteen}%
\setlength{\treewidthfifteen}{\treewidthfourteen}%
\setlength{\treewidthfourteen}{\treewidththirteen}%
\setlength{\treewidththirteen}{\treewidthtwelve}%
\setlength{\treewidthtwelve}{\treewidtheleven}%
\setlength{\treewidtheleven}{\treewidthten}%
\setlength{\treewidthten}{\treewidthnine}%
\setlength{\treewidthnine}{\treewidtheight}%
\setlength{\treewidtheight}{\treewidthseven}%
\setlength{\treewidthseven}{\treewidthsix}%
\setlength{\treewidthsix}{\treewidthfive}%
\setlength{\treewidthfive}{\treewidthfour}%
\setlength{\treewidthfour}{\treewidththree}%
\setlength{\treewidththree}{\treewidthtwo}%
\setlength{\treewidthtwo}{\treewidthone}%
\setlength{\treewidthone}{\wd\treeboxone}}
\newcommand{\branch}[2]{%
\setcounter{branchcount}{#1}%
\ifnum\value{branchcount}=1\sbox{\parentbox}{\ontop{#2}}%
\setlength{\parentoffset}{\treeoffsetone}%
\addtolength{\parentoffset}{-0.5\wd\parentbox}%
\setlength{\daughteroffset}{0in}%
\ifdim\parentoffset<0in%
\setlength{\daughteroffset}{-\parentoffset}%
\setlength{\parentoffset}{0in}\fi%
\setlength{\parentwidth}{\parentoffset}%
\addtolength{\parentwidth}{\wd\parentbox}%
\setlength{\treeoffset}{\daughteroffset}%
\addtolength{\treeoffset}{\treeoffsetone}%
\setlength{\treewidth}{\wd\treeboxone}%
\addtolength{\treewidth}{\daughteroffset}%
\ifdim\treewidth<\parentwidth\setlength{\treewidth}{\parentwidth}\fi%
\sbox{\treebox}{\begin{minipage}{\treewidth}%
\begin{flushleft}%
\hspace*{\parentoffset}\usebox{\parentbox}\\
{\setlength{\unitlength}{2ex}%
\hspace*{\treeoffset}\begin{picture}(0,1)%
\put(0,0){\line(0,1){1}}%
\end{picture}}\\
\vspace{-\baselineskip}
\hspace*{\daughteroffset}%
\raisebox{-\ht\treeboxone}{\usebox{\treeboxone}}%
\end{flushleft}%
\end{minipage}}%
\setlength{\treeoffsetone}{\parentoffset}%
\addtolength{\treeoffsetone}{0.5\wd\parentbox}%
\setlength{\treeshiftone}{0pt}%
\setlength{\treewidthone}{\treewidth}%
\sbox{\treeboxone}{\usebox{\treebox}}%
\else\ifnum\value{branchcount}=2\sbox{\parentbox}{\ontop{#2}}%
\setlength{\branchwidthone}{\treewidthtwo}%
\addtolength{\branchwidthone}{\treeoffsetone}%
\addtolength{\branchwidthone}{-\treeshiftone}%
\addtolength{\branchwidthone}{-\treeoffsettwo}%
\setlength{\branchwidth}{\branchwidthone}%
\setlength{\daughteroffsetone}{\branchwidth}%
\addtolength{\daughteroffsetone}{-\branchwidthone}%
\addtolength{\daughteroffsetone}{-\treeshiftone}%
\setlength{\parentoffset}{-0.5\wd\parentbox}%
\addtolength{\parentoffset}{\treeoffsettwo}%
\addtolength{\parentoffset}{0.5\branchwidth}%
\setlength{\daughteroffset}{0in}%
\ifdim\parentoffset<0in%
\setlength{\daughteroffset}{-\parentoffset}%
\setlength{\parentoffset}{0in}\fi%
\setlength{\parentwidth}{\parentoffset}%
\addtolength{\parentwidth}{\wd\parentbox}%
\setlength{\treeoffset}{\daughteroffset}%
\addtolength{\treeoffset}{\treeoffsettwo}%
\setlength{\treewidth}{\wd\treeboxone}%
\addtolength{\treewidth}{\daughteroffsetone}%
\addtolength{\treewidth}{\treewidthtwo}%
\addtolength{\treewidth}{\daughteroffset}%
\ifdim\treewidth<\parentwidth\setlength{\treewidth}{\parentwidth}\fi%
\sbox{\treebox}{\begin{minipage}{\treewidth}%
\begin{flushleft}%
\hspace*{\parentoffset}\usebox{\parentbox}\\
{\setlength{\unitlength}{0.5\branchwidth}%
\hspace*{\treeoffset}\begin{picture}(2,0.5)%
\put(0,0){\line(2,1){1}}%
\put(2,0){\line(-2,1){1}}%
\end{picture}}\\
\vspace{-\baselineskip}
\hspace*{\daughteroffset}%
\makebox[\treewidthtwo][l]%
{\raisebox{-\ht\treeboxtwo}{\usebox{\treeboxtwo}}}%
\hspace*{\daughteroffsetone}%
\raisebox{-\ht\treeboxone}{\usebox{\treeboxone}}%
\end{flushleft}%
\end{minipage}}%
\setlength{\treeoffsetone}{\parentoffset}%
\addtolength{\treeoffsetone}{0.5\wd\parentbox}%
\setlength{\treeshiftone}{0pt}%
\setlength{\treewidthone}{\treewidth}%
\sbox{\treeboxone}{\usebox{\treebox}}\poptree%
\else\ifnum\value{branchcount}=3\sbox{\parentbox}{\ontop{#2}}%
\setlength{\branchwidthone}{\treewidthtwo}%
\addtolength{\branchwidthone}{\treeoffsetone}%
\addtolength{\branchwidthone}{-\treeshiftone}%
\addtolength{\branchwidthone}{-\treeoffsettwo}%
\setlength{\branchwidthtwo}{\treewidththree}%
\addtolength{\branchwidthtwo}{\treeoffsettwo}%
\addtolength{\branchwidthtwo}{-\treeshifttwo}%
\addtolength{\branchwidthtwo}{-\treeoffsetthree}%
\setlength{\branchwidth}{\branchwidthone}%
\ifdim\branchwidthtwo>\branchwidth%
\setlength{\branchwidth}{\branchwidthtwo}\fi%
\setlength{\daughteroffsetone}{\branchwidth}%
\addtolength{\daughteroffsetone}{-\branchwidthone}%
\addtolength{\daughteroffsetone}{-\treeshiftone}%
\setlength{\daughteroffsettwo}{\branchwidth}%
\addtolength{\daughteroffsettwo}{-\branchwidthtwo}%
\addtolength{\daughteroffsettwo}{-\treeshifttwo}%
\setlength{\parentoffset}{-0.5\wd\parentbox}%
\addtolength{\parentoffset}{\treeoffsetthree}%
\addtolength{\parentoffset}{\branchwidth}%
\setlength{\daughteroffset}{0in}%
\ifdim\parentoffset<0in%
\setlength{\daughteroffset}{-\parentoffset}%
\setlength{\parentoffset}{0in}\fi%
\setlength{\parentwidth}{\parentoffset}%
\addtolength{\parentwidth}{\wd\parentbox}%
\setlength{\treeoffset}{\daughteroffset}%
\addtolength{\treeoffset}{\treeoffsetthree}%
\setlength{\treewidth}{\wd\treeboxone}%
\addtolength{\treewidth}{\daughteroffsetone}%
\addtolength{\treewidth}{\treewidthtwo}%
\addtolength{\treewidth}{\daughteroffsettwo}%
\addtolength{\treewidth}{\treewidththree}%
\addtolength{\treewidth}{\daughteroffset}%
\ifdim\treewidth<\parentwidth\setlength{\treewidth}{\parentwidth}\fi%
\sbox{\treebox}{\begin{minipage}{\treewidth}%
\begin{flushleft}%
\hspace*{\parentoffset}\usebox{\parentbox}\\
{\setlength{\unitlength}{0.5\branchwidth}%
\hspace*{\treeoffset}\begin{picture}(4,1)%
\put(0,0){\line(2,1){2}}%
\put(2,0){\line(0,1){1}}%
\put(4,0){\line(-2,1){2}}%
\end{picture}}\\
\vspace{-\baselineskip}
\hspace*{\daughteroffset}%
\makebox[\treewidththree][l]%
{\raisebox{-\ht\treeboxthree}{\usebox{\treeboxthree}}}%
\hspace*{\daughteroffsettwo}%
\makebox[\treewidthtwo][l]%
{\raisebox{-\ht\treeboxtwo}{\usebox{\treeboxtwo}}}%
\hspace*{\daughteroffsetone}%
\raisebox{-\ht\treeboxone}{\usebox{\treeboxone}}%
\end{flushleft}%
\end{minipage}}%
\setlength{\treeoffsetone}{\parentoffset}%
\addtolength{\treeoffsetone}{0.5\wd\parentbox}%
\setlength{\treeshiftone}{0pt}%
\setlength{\treewidthone}{\treewidth}%
\sbox{\treeboxone}{\usebox{\treebox}}\poptree\poptree%
\else\ifnum\value{branchcount}=4\sbox{\parentbox}{\ontop{#2}}%
\setlength{\branchwidthone}{\treewidthtwo}%
\addtolength{\branchwidthone}{\treeoffsetone}%
\addtolength{\branchwidthone}{-\treeshiftone}%
\addtolength{\branchwidthone}{-\treeoffsettwo}%
\setlength{\branchwidthtwo}{\treewidththree}%
\addtolength{\branchwidthtwo}{\treeoffsettwo}%
\addtolength{\branchwidthtwo}{-\treeshifttwo}%
\addtolength{\branchwidthtwo}{-\treeoffsetthree}%
\setlength{\branchwidththree}{\treewidthfour}%
\addtolength{\branchwidththree}{\treeoffsetthree}%
\addtolength{\branchwidththree}{-\treeshiftthree}%
\addtolength{\branchwidththree}{-\treeoffsetfour}%
\setlength{\branchwidth}{\branchwidthone}%
\ifdim\branchwidthtwo>\branchwidth%
\setlength{\branchwidth}{\branchwidthtwo}\fi%
\ifdim\branchwidththree>\branchwidth%
\setlength{\branchwidth}{\branchwidththree}\fi%
\setlength{\daughteroffsetone}{\branchwidth}%
\addtolength{\daughteroffsetone}{-\branchwidthone}%
\addtolength{\daughteroffsetone}{-\treeshiftone}%
\setlength{\daughteroffsettwo}{\branchwidth}%
\addtolength{\daughteroffsettwo}{-\branchwidthtwo}%
\addtolength{\daughteroffsettwo}{-\treeshifttwo}%
\setlength{\daughteroffsetthree}{\branchwidth}%
\addtolength{\daughteroffsetthree}{-\branchwidththree}%
\addtolength{\daughteroffsetthree}{-\treeshiftthree}%
\setlength{\parentoffset}{-0.5\wd\parentbox}%
\addtolength{\parentoffset}{\treeoffsetfour}%
\addtolength{\parentoffset}{1.5\branchwidth}%
\setlength{\daughteroffset}{0in}%
\ifdim\parentoffset<0in%
\setlength{\daughteroffset}{-\parentoffset}%
\setlength{\parentoffset}{0in}\fi%
\setlength{\parentwidth}{\parentoffset}%
\addtolength{\parentwidth}{\wd\parentbox}%
\setlength{\treeoffset}{\daughteroffset}%
\addtolength{\treeoffset}{\treeoffsetfour}%
\setlength{\treewidth}{\wd\treeboxone}%
\addtolength{\treewidth}{\daughteroffsetone}%
\addtolength{\treewidth}{\treewidthtwo}%
\addtolength{\treewidth}{\daughteroffsettwo}%
\addtolength{\treewidth}{\treewidththree}%
\addtolength{\treewidth}{\daughteroffsetthree}%
\addtolength{\treewidth}{\treewidthfour}%
\addtolength{\treewidth}{\daughteroffset}%
\ifdim\treewidth<\parentwidth\setlength{\treewidth}{\parentwidth}\fi%
\sbox{\treebox}{\begin{minipage}{\treewidth}%
\begin{flushleft}%
\hspace*{\parentoffset}\usebox{\parentbox}\\
{\setlength{\unitlength}{0.5\branchwidth}%
\hspace*{\treeoffset}\begin{picture}(6,1)%
\put(0,0){\line(3,1){3}}%
\put(2,0){\line(1,1){1}}%
\put(4,0){\line(-1,1){1}}%
\put(6,0){\line(-3,1){3}}%
\end{picture}}\\
\vspace{-\baselineskip}
\hspace*{\daughteroffset}%
\makebox[\treewidthfour][l]%
{\raisebox{-\ht\treeboxfour}{\usebox{\treeboxfour}}}%
\hspace*{\daughteroffsetthree}%
\makebox[\treewidththree][l]%
{\raisebox{-\ht\treeboxthree}{\usebox{\treeboxthree}}}%
\hspace*{\daughteroffsettwo}%
\makebox[\treewidthtwo][l]%
{\raisebox{-\ht\treeboxtwo}{\usebox{\treeboxtwo}}}%
\hspace*{\daughteroffsetone}%
\raisebox{-\ht\treeboxone}{\usebox{\treeboxone}}%
\end{flushleft}%
\end{minipage}}%
\setlength{\treeoffsetone}{\parentoffset}%
\addtolength{\treeoffsetone}{0.5\wd\parentbox}%
\setlength{\treeshiftone}{0pt}%
\setlength{\treewidthone}{\treewidth}%
\sbox{\treeboxone}{\usebox{\treebox}}\poptree\poptree\poptree%
\else\ifnum\value{branchcount}=5\sbox{\parentbox}{\ontop{#2}}%
\setlength{\branchwidthone}{\treewidthtwo}%
\addtolength{\branchwidthone}{\treeoffsetone}%
\addtolength{\branchwidthone}{-\treeshiftone}%
\addtolength{\branchwidthone}{-\treeoffsettwo}%
\setlength{\branchwidthtwo}{\treewidththree}%
\addtolength{\branchwidthtwo}{\treeoffsettwo}%
\addtolength{\branchwidthtwo}{-\treeshifttwo}%
\addtolength{\branchwidthtwo}{-\treeoffsetthree}%
\setlength{\branchwidththree}{\treewidthfour}%
\addtolength{\branchwidththree}{\treeoffsetthree}%
\addtolength{\branchwidththree}{-\treeshiftthree}%
\addtolength{\branchwidththree}{-\treeoffsetfour}%
\setlength{\branchwidthfour}{\treewidthfive}%
\addtolength{\branchwidthfour}{\treeoffsetfour}%
\addtolength{\branchwidthfour}{-\treeshiftfour}%
\addtolength{\branchwidthfour}{-\treeoffsetfive}%
\setlength{\branchwidth}{\branchwidthone}%
\ifdim\branchwidthtwo>\branchwidth%
\setlength{\branchwidth}{\branchwidthtwo}\fi%
\ifdim\branchwidththree>\branchwidth%
\setlength{\branchwidth}{\branchwidththree}\fi%
\ifdim\branchwidthfour>\branchwidth%
\setlength{\branchwidth}{\branchwidthfour}\fi%
\setlength{\daughteroffsetone}{\branchwidth}%
\addtolength{\daughteroffsetone}{-\branchwidthone}%
\addtolength{\daughteroffsetone}{-\treeshiftone}%
\setlength{\daughteroffsettwo}{\branchwidth}%
\addtolength{\daughteroffsettwo}{-\branchwidthtwo}%
\addtolength{\daughteroffsettwo}{-\treeshifttwo}%
\setlength{\daughteroffsetthree}{\branchwidth}%
\addtolength{\daughteroffsetthree}{-\branchwidththree}%
\addtolength{\daughteroffsetthree}{-\treeshiftthree}%
\setlength{\daughteroffsetfour}{\branchwidth}%
\addtolength{\daughteroffsetfour}{-\branchwidthfour}%
\addtolength{\daughteroffsetfour}{-\treeshiftfour}%
\setlength{\parentoffset}{-0.5\wd\parentbox}%
\addtolength{\parentoffset}{\treeoffsetfive}%
\addtolength{\parentoffset}{2\branchwidth}%
\setlength{\daughteroffset}{0in}%
\ifdim\parentoffset<0in%
\setlength{\daughteroffset}{-\parentoffset}%
\setlength{\parentoffset}{0in}\fi%
\setlength{\parentwidth}{\parentoffset}%
\addtolength{\parentwidth}{\wd\parentbox}%
\setlength{\treeoffset}{\daughteroffset}%
\addtolength{\treeoffset}{\treeoffsetfive}%
\setlength{\treewidth}{\wd\treeboxone}%
\addtolength{\treewidth}{\daughteroffsetone}%
\addtolength{\treewidth}{\treewidthtwo}%
\addtolength{\treewidth}{\daughteroffsettwo}%
\addtolength{\treewidth}{\treewidththree}%
\addtolength{\treewidth}{\daughteroffsetthree}%
\addtolength{\treewidth}{\treewidthfour}%
\addtolength{\treewidth}{\daughteroffsetfour}%
\addtolength{\treewidth}{\treewidthfive}%
\addtolength{\treewidth}{\daughteroffset}%
\ifdim\treewidth<\parentwidth\setlength{\treewidth}{\parentwidth}\fi%
\sbox{\treebox}{\begin{minipage}{\treewidth}%
\begin{flushleft}%
\hspace*{\parentoffset}\usebox{\parentbox}\\
{\setlength{\unitlength}{0.5\branchwidth}%
\hspace*{\treeoffset}\begin{picture}(8,1)%
\put(0,0){\line(4,1){4}}%
\put(2,0){\line(2,1){2}}%
\put(4,0){\line(0,1){1}}%
\put(6,0){\line(-2,1){2}}%
\put(8,0){\line(-4,1){4}}%
\end{picture}}\\
\vspace{-\baselineskip}
\hspace*{\daughteroffset}%
\makebox[\treewidthfive][l]%
{\raisebox{-\ht\treeboxfour}{\usebox{\treeboxfive}}}%
\hspace*{\daughteroffsetfour}%
\makebox[\treewidthfour][l]%
{\raisebox{-\ht\treeboxfour}{\usebox{\treeboxfour}}}%
\hspace*{\daughteroffsetthree}%
\makebox[\treewidththree][l]%
{\raisebox{-\ht\treeboxthree}{\usebox{\treeboxthree}}}%
\hspace*{\daughteroffsettwo}%
\makebox[\treewidthtwo][l]%
{\raisebox{-\ht\treeboxtwo}{\usebox{\treeboxtwo}}}%
\hspace*{\daughteroffsetone}%
\raisebox{-\ht\treeboxone}{\usebox{\treeboxone}}%
\end{flushleft}%
\end{minipage}}%
\setlength{\treeoffsetone}{\parentoffset}%
\addtolength{\treeoffsetone}{0.5\wd\parentbox}%
\setlength{\treeshiftone}{0pt}%
\setlength{\treewidthone}{\treewidth}%
\sbox{\treeboxone}{\usebox{\treebox}}\poptree\poptree\poptree\poptree%
\else\typeout{QobiTeX warning--- Can't handle #1 branching}\fi\fi\fi\fi\fi}
\newcommand{\tree}{%
\usebox{\treeboxone}
\setlength{\treeoffsetone}{\treeoffsettwo}%
\sbox{\treeboxone}{\usebox{\treeboxtwo}}%
\poptree}

\begin{document}

\begin{singlespace}
\begin{center}
\textbf{\large Microplanning with Communicative Intentions: \\
	The SPUD System} \\
\begin{tabular}{ccccc}
Matthew Stone & 
Christine Doran &
Bonnie Webber & 
Tonia Bleam &
Martha Palmer \\
Rutgers &
MITRE &
Edinburgh &
Pennsylvania &
Pennsylvania 
\end{tabular}
\end{center}

\sloppy

\section*{Abstract}
\emph{The process of microplanning encompasses a range of problems in
   Natural Language Generation (NLG), such as referring expression
   generation, lexical choice, and aggregation, problems in which a
   generator must bridge underlying domain-specific representations
   and general linguistic representations.  In this paper, we describe
   a uniform approach to microplanning based on declarative
   representations of a generator's communicative intent.  These
   representations describe the \term{results} of NLG: communicative
   intent associates the concrete linguistic structure planned by the
   generator with inferences that show how the meaning of that
   structure communicates needed information about some application
   domain in the current discourse context.  Our approach, implemented
   in the \spud\ (sentence planning using description) microplanner,
   uses the lexicalized tree-adjoining grammar formalism (LTAG) to
   connect structure to meaning and uses modal logic programming to
   connect meaning to context.  At the same time, communicative intent
   representations provide a \term{resource} for the \term{process} of
   NLG.  Using representations of communicative intent, a generator
   can augment the syntax, semantics and pragmatics of an incomplete
   sentence simultaneously, and can assess its progress on the various
   problems of microplanning incrementally.  The declarative
   formulation of communicative intent translates into a well-defined
   methodology for designing grammatical and conceptual resources
   which the generator can use to achieve desired microplanning
   behavior in a specified domain.
}
\tableofcontents
\vspace*{3em}
\hspace*{\fill}
\begin{tabular}{l}
Contact Address \\
Matthew Stone \\
Department of Computer Science \\
and Center for Cognitive Science \\
Rutgers, the State University of New Jersey \\
110 Frelinghuysen Road \\
Piscataway NJ 08854-8019 \\
mdstone@cs.rutgers.edu \\
\today 
\end{tabular}
\end{singlespace}
\pagebreak

\pagebreak

\section{Motivation}
\label{introduction-sec}

   	Success in Natural Language Generation (NLG) requires
   connecting domain knowledge and linguistic representations.  After
   all, an agent must have substantive and correct knowledge for
   others to benefit from the information it provides.  And an agent
   must communicate this information in a concise and natural form, if
   people are to understand it.  The instruction in
   \sref{reposition-eg} from an aircraft maintenance manual suggests
   the challenge involved in reconciling these two kinds of
   representation.
\sentence{reposition-eg}{
	Reposition coupling nut.
}
	The domain knowledge behind \sref{reposition-eg} must specify
   a definite location where the coupling nut goes, and a definite
   function in an overall repair that the nut fulfills there.
   However, the linguistic form does not indicate this location or
   function explicitly; instead, its precise vocabulary and structure
   allows one to draw on one's existing understanding of the repair to
   fill in these details for oneself.
	
\begin{figure}
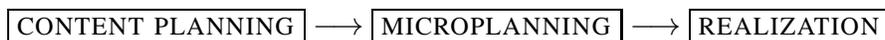

\begin{center}
\fbox{\textsc{content planning}}
$\longrightarrow$
\fbox{\textsc{microplanning}}
$\longrightarrow$
\fbox{\textsc{realization}}
\end{center}
\caption{Microplanning in the NLG pipeline.}
\label{arch-fig}
\end{figure}

	In the architecture typical of most NLG systems, and in many
   psycholinguistic models of speaking, a distinctive process of
   \term{microplanning} is responsible for making the connection
   between domain knowledge and linguistic representations.%
\dspftnt{
	The name microplanning originates in Levelt's psycholinguistic
   model of language production \cite{levelt:speaking}, and is adopted
   in Reiter and Dale's overview of NLG systems
   \cite{reiter/dale:bnlg}.  The process has also been termed
   \term{sentence planning}, beginning with \cite{rambow/korelsky:sp}.
}
	Microplanning intervenes between a process of \term{content
   planning}, in which the agent assembles information to provide in
   conversation by drawing on knowledge and conventions from a
   particular domain, and the domain-independent process of
   \term{realization} through which a concrete presentation is
   actually delivered to a conversational partner.  These processes
   are frequently implemented in a pipeline architecture, as shown in
   Figure~\ref{arch-fig}.
	Concretely, the content planner is typically responsible for
   responding to the information goals of the conversation by
   identifying a body of domain facts to present, and by organizing
   those facts into a rhetorical structure that represents a coherent
   and potentially convincing argument.  Microplanning takes these
   domain facts and recodes them in suitable linguistic terms.
   Finally, realization is responsible for a variety of low-level
   linguistic tasks (including certain syntactic and morphological
   processes), as well as such formatting tasks as laying out a
   presentation on a page or a screen or performing speech synthesis.
   See Reiter and Dale for a thorough overview of these different
   stages in NLG systems \cite{reiter/dale:bnlg}.

	Microplanning often looks like a grab-bag of idiosyncratic
   tasks, each of which calls for its own representations and
   algorithms.  For example, consider the three microplanning tasks
   that Reiter and Dale survey: referring expression generation,
   lexical choice, and aggregation.  
\begin{itemize}
\item
	In referring expression generation, the task is to derive an
   identifying description to take the place of the internal
   representation of some discourse referent.  To carry out this task,
   generators often execute rules to elaborate an incomplete semantic
   specification of an utterance (\ling{the rabbit}, say) by
   incorporating additional descriptive concepts (for instance
   \ling{white}, to yield \emph{the white rabbit})
   \cite{dale/haddock:referring,dale:expression/book,dale/reiter:cogsci}.

\item
	In lexical choice, the task is to select a word from among the
   many that describe an object or event.  To perform lexical choice,
   generators often invoke a pattern-matching process that rewrites
   domain information (that there is a caused event of motion along a
   surface, say) in terms of available language-specific meanings (to
   recognize that there is \emph{sliding}, for example)
   \cite{nogier/zock:91,elhadad:constraints,stede:cl98}.

\item
	In aggregation, the task is to use modifiers, conjoined
   phrases, and other linguistic constructions to pack information
   concisely into fewer (but more complex) sentences.  Aggregation
   depends on applying operators that detect relationships within the
   information to be expressed, such as repeated reference to common
   participants (that Doe is a patient and that Doe is female, say),
   and then reorganize related semantic material into a nested
   structure (to obtain \emph{Doe is a female patient}, for example)
   \cite{dalianis:phd,shaw:inlg98}.
\end{itemize}
	But tasks like referring expression generation, lexical choice
   and aggregation interact in systematic and intricate ways
   \cite{wanner/hovy:inlg96}.  These interactions represent a major
   challenge to integrating heterogeneous microplanning
   processes---all the more so in that NLG systems adopt widely
   divergent, often application-specific methods for sequencing these
   operations and combining their results \cite{cahill/reape:99-05}.

	In contrast to this heterogeneity, we advocate a
   \term{uniform} approach to microplanning.  Our generator, called
   \spud\ (for sentence planning using description), maintains a
   common representation of its provisional utterance during
   microplanning and carries out a single decision-making strategy
   using this representation.  In what follows, we draw on and extend
   our preliminary presentations of \spud\ in
   \cite{colloc-paper,gen-paper,genw-paper,stone:tag+00} to describe
   this approach in more detail.

	The key to our framework is our generator's representation of
   the \term{interpretation} of its provisional utterances.  We call
   this representation \term{communicative intent}.  In doing so, we
   emphasize that language use involves a \term{ladder of related
   intentions} \cite{clark:book}, from uttering particular words,
   through referring to shared individuals from the context and
   contributing new information, to answering open questions in the
   conversation.  (Clark's ladder metaphor particularly suits the
   graphical presentation of communicative intent that we introduce in
   Section~\ref{problem-sec}.)  Since many of these intentions are
   adopted during the course of microplanning, communicative intent
   represents the \term{results} of generation.  At the same time, we
   emphasize that microplanning is a deliberative process like any
   other, in which the provisional intentions that an agent is
   committed to can guide and constrain further reasoning
   \cite{bratman:book,pollack:uses}.  Thus, communicative intent also
   serves as a key resource for the \term{process} of generation.

  	Our specific representation of communicative intent, described
   in Sections~\ref{problem-sec}--\ref{inference-section}, associates
   a concrete linguistic structure with inferences about its meaning
   that show how, in the current discourse context, that structure
   describes a variety of generalized individuals%
\dspftnt{
	meaning not only objects but also actions, events, and any
   other constituents of a rich ontology for natural language, as
   described in \cite{bach:lectures} and advocated in
   \cite{hobbs:promiscuity}
}
	and thereby communicates specific information about the
   application domain.  As argued in
   Sections~\ref{algorithm-section}--\ref{example-section}, this
   representation has all the information required to make decisions
   in microplanning.  For example, it records progress towards
   unambiguous formulation of referring expressions; it shows how
   alternative choices of words and syntactic constructions suit an
   ongoing generation task to different degrees because they
   encapsulate different constellations of domain information or set
   up different links with the context; and it indicates how given
   structure and meaning may be elaborated with modifiers so that
   multiple pieces of information can be organized for expression in a
   single sentence.  Thus, with a model of communicative intent,
   \spud\ can augment the syntax, semantics and pragmatics of an
   incomplete sentence simultaneously, and can assess its progress on
   the various interacting subproblems of microplanning incrementally.

	In communicative intent, the pairing between structure and
   meaning is specified by a grammar which describes linguistic
   analyses in formal terms.  Likewise, links between domain knowledge
   and linguistic meanings are formalized in terms of logical
   relationships among concepts.  To construct communicative intent,
   we draw conclusions about interpretation by reasoning from these
   specifications.  Thus, communicative intent is a \term{declarative}
   representation; it enjoys the numerous advantages of declarative
   programming in Natural Language Processing
   \cite{pereira/shieber:pnlp}.  In particular, as we discuss in
   Section~\ref{build-sec}, the declarative use of grammatical
   resources leads to a concrete methodology for designing grammars
   that allow \spud\ to achieve desired behavior in a specified
   domain.

   	Performing microplanning using communicative intent means
   searching through derivations of a grammar to construct an
   utterance and its interpretation simultaneously.  This search is
   facilitated with a grammar formalism that packages meaningful
   decisions together and allows those decisions to be assessed
   incrementally; \spud\ uses the lexicalized tree-adjoining grammar
   formalism.  Meanwhile, the use of techniques such as logic
   programming and constraint satisfaction leads to efficient methods
   to determine the communicative intent for a given linguistic form
   and evaluate progress on a microplanning problem.  These design
   decisions, combined for the first time in \spud, lend considerable
   promise to communicative-intent--based microplanning as an
   efficient and manageable framework for practical NLG.

\section{Introduction to Microplanning Based on Communicative Intent}
\label{problem-sec}

   	We begin with an extended illustration of communicative intent
   and motivation for its use in microplanning.  In
   Section~\ref{converse-intro-subsec}, we situate representations of
   communicative intent more broadly within research on the cognitive
   science of contributing to conversation, and we use a high-level
   case-study of communicative intent to discuss more precisely how
   such representations may be constructed from linguistic and domain
   knowledge.  In Section~\ref{rep-use-intro-subsec}, we show how such
   representations could be used to guide reasoning in conversational
   systems, particularly to support microplanning decisions.  Finally,
   in Section~\ref{spud-intro-subsec}, we identify the key assumptions
   that we have made in \spud, in order to construct an effective NLG
   system that implements a model of communicative intent.

\subsection{Representing Communicative Intent}
\label{converse-intro-subsec}
\label{rep-intro-subsec}

	Communicative intent responds to a view of contributing to
   conversation whose antecedents are Grice's description of
   communication in terms of intention recognition
   (\cite{grice:meaning}, as updated by Thomason
   \cite{thomason:intentions}) and Clark's approach to language use as
   joint activity \cite{clark:book}.

	According to this view, conversation consists of joint
   activity undertaken in support of common goals.  Participants take
   actions publicly; they coordinate so that all agree on how each
   action is intended to advance the goals of the conversation, and so
   that all agree on whether the action succeeds in its intended
   effects.  This joint activity defines a \term{conversational
   process} which people engage in intentionally and collaboratively,
   and, we might even suppose, rationally.  The fundamental component
   of conversational process is the coordination by which speakers
   manifest and hearers recognize communicative intentions carried by
   linguistic actions.  But there are many other aspects of
   conversational process: acknowledgment, grounding and backchannels;
   clarification and repair; and even regulation of turn-taking.  (See
   \cite{clark:book} and references therein.)  Dialogue systems
   increasingly implement rich models of conversational process; see
   e.g. \cite{cassell:cacm}.  This makes it vital that a sentence
   planning module interface with and support a system's
   conversational process.

	Like any deliberative process \cite{pollack:uses}, this
   conversational process depends on plans, which provide resources
   for decision-making.  In conversation, these plans map out how the
   respondent might use certain words to convey certain information:
   they describe the utterances of words and linguistic constructions,
   spell out the meanings of those utterances, and show how these
   utterances, with these meanings, could contribute structure,
   representing propositions and intentions, to the
   \term{conversational record}, an evolving abstract model of the the
   dialogue \cite{thomason:intentions}.  In other words, a
   communicative plan is a structure, built by reasoning from a
   grammar, which summarizes the interpretation of an utterance in
   context.  Such plans constitute our \term{abstract level of
   representation of communicative intent}.  Note that this level of
   representation presupposes, and thereby suppresses, the specific
   collaborative activity that determines how meaning is actually
   recognized and ratified.  Communicative intent is a resource for
   these processes in conversation, not a description of them.

   	We can develop these ideas in more detail by considering an
   illustrative example.
\bsentence{\myinstruction}{instruction-eg}{
	Slide coupling nut onto elbow to uncover fuel-line sealing
   ring.
}

\begin{figure}
\begin{center}
\psfig{file=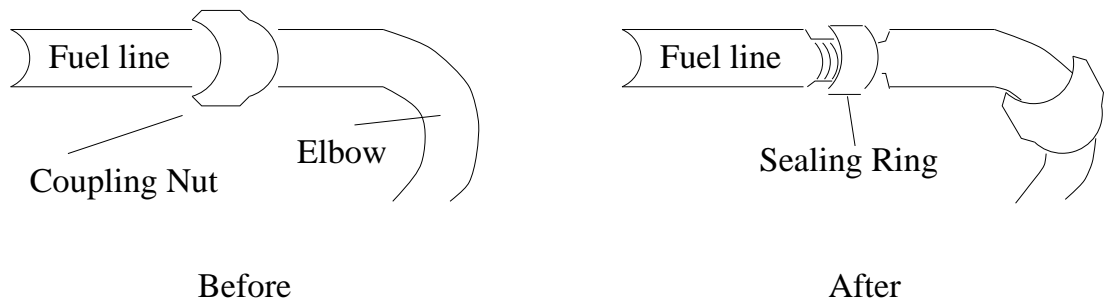,scale=.8,silent=}
\end{center}
\caption{Carrying out instruction \sref{instruction-eg}
	 in an aircraft fuel system.}
\label{instruct:fig1}
\end{figure}

	We draw \sref{instruction-eg} from an aircraft maintenance
   domain that we have studied in detail and report on fully in
   Section~\ref{build-sec}; Figure~\ref{instruct:fig1} shows the
   effect of the action on the aircraft fuel system.
 	In this system, pipes are joined together using sealing rings
   that fit snugly over the ends of adjacent pipes.  Sometimes these
   joints are secured by positioning a coupling nut around the seal to
   keep it tight and then installing a retainer to keep the coupling
   nut and seal in place.  In checking (and, if necessary, replacing)
   such sealing rings, personnel must gain access to them by first
   removing the retainer and then sliding the coupling nut away.
   Figure~\ref{instruct:fig1} illustrates part of this process for a
   case where an instructor could use \sref{instruction-eg} to direct
   an actor to perform the step of sliding the coupling nut clear.

	We draw on this account of the domain in which
   \sref{instruction-eg} is used, to describe the communicative intent
   with which we represent \sref{instruction-eg}.  We consider three
   components of communicative intent in turn.
\begin{itemize}
\item
	The first derives from the update to the conversational record
   that the instruction is meant to achieve.  This update includes the
   fact that the actor is to carry out a motion specified in terms of
   the given objects and landmarks---namely, the actor is to move the
   coupling nut smoothly along the surface of the fuel line from its
   current position onto the elbow.  But the update also spells out
   the intended purpose of this action: the action is to uncover the
   sealing ring and, we may presume, thereby enable subsequent
   maintenance steps.  So communicative intent must show how the
   meanings of the words in \sref{instruction-eg} are intended to put
   on the record this characterization of movement and purpose.

\item
	The second component relates to the set of referents that the
   instruction describes and evokes: the elbow, the coupling nut
   adjacent to the elbow, the fuel line, and the sealing ring on the
   fuel line.  The actor is expected to be familiar with these
   referents; this familiarity might come from the actor's general
   experience with the aircraft, from a diagram accompanying a block
   of instructions, or just from the physical surroundings as the
   actor carries out the instructions.  In any case, the expectation
   of familiarity corresponds to a constraint on the idealized
   conversational record: the specified referents with the specified
   properties must be found there.  Indeed, in understanding
   \sref{instruction-eg}, the actor can and should use this constraint
   together with the shared information from the conversational record
   to identify the intended objects and landmarks.  Thus,
   communicative intent must represent this constraint on the
   conversational record and anticipate the actor's use of it to
   resolve the instructor's references.

\item
   	The third component accounts for the collection of
   constructions by which the instructor frames the instruction.  The
   instruction is an imperative; that choice shows (among other
   things) that the instructor's relationship with the actor empowers
   the instructor to impose obligations for action on the actor.  (In
   our domain, maintenance instructions are in fact military orders.)
   Meanwhile, the use of definite noun phrases that omit the article
   \emph{the} reflects the distinctive telegraphic style adopted in
   these instructions.  Of course, the relationship of instructor and
   actor and the distinctive linguistic style of the domain are both
   part of the conversational record, and the instructor anticipates
   that the actor will make connections with these shared
   representations in interpreting the constructions in
   \sref{instruction-eg}.  Thus communicative intent must also
   represent these connections.

\end{itemize}
   	To represent communicative intent, then, we will need to
   associate a formal representation of the utterance in
   \sref{instruction-eg} with a model of interpretation that describes
   these three components: how the utterance adds information that
   links up with the goals of communication; how it imposes
   constraints that link up with shared characterizations of objects;
   and how it establishes specific connections to the status of
   participants and referents in the discourse.

	Schematically, we can represent the form of the utterance
   using a dependency tree as shown in \sref{instruction-deps}.
\sentence{instruction-deps}{
\mbox{
\small
\leaf{coupling-nut (zero-def)}
\leaf{elbow (zero-def)}
\branch{1}{onto (\term{vp} modifier)}
\leaf{fuel-line (\term{n} modifier)}
\branch{1}{sealing-ring (zero-def)}
\branch{1}{uncover (infinitive)}
\branch{1}{$\langle$purpose$\rangle$ (bare infinitival adjunct)}
\branch{3}{slide (imperative)}
\tree
}}
	This tree analyzes the utterance as being made up of
   \textsc{elements} bearing specific content and realized in specific
   syntactic constructions; these elements form the nodes in the tree.
   Thus, the leftmost leaf, labeled \emph{coupling-nut (zero-def)},
   represents the fact that the noun \emph{coupling nut} is used here,
   in construction with the zero definite determiner characteristic of
   this genre, to contribute a noun phrase to the sentence.
   Generally, these elements include lexical items, as \ling{coupling
   nut} does; but in cases such as the
   $\langle$\ling{purpose}$\rangle$ element, we may simply find some
   distinctive syntax associated with meaning that could otherwise be
   realized by a construction with explicit lexical material (\ling{in
   order}, for a purpose relation).  Edges in the tree represent
   operations of syntactic combination; the child node may either
   supply a required \term{complement} to the parent node (as the node
   for \emph{coupling-nut} does for its parent \emph{slide}) or may
   provide an optional \term{modifier} that supplements the parent's
   interpretation (as the node for $\langle$\ling{purpose}$\rangle$
   does for its parent \emph{slide}).

   	We pair \sref{instruction-deps} with a record of
   interpretation by taking into account two sources of information:
   the \textsc{grammatical conventions} that associate meaningful
   conditions with an utterance across contexts, in a public
   representation accessible to speaker and hearer; and the
   \textsc{speaker's presumptions} which describe specific
   instantiations for these conditions in the current context, and
   determine the precise communicative effects of the utterance in
   context.

	We assume that grammatical conventions associate each of the
   elements in \sref{instruction-deps} with an \term{assertion} that
   contributes to the update intended for the utterance; a
   \term{presupposition} intended to ground the utterance in shared
   knowledge about the domain; and a \term{pragmatic} condition
   intended to reflect the status of participants and referents in the
   discourse.  There is a long history in computational linguistics
   for the assumption that utterance meaning is a conjunction of
   atomic contributions made by words (in constructions); see
   particularly \cite{hobbs:promiscuity}.  Our use of assertion and
   presupposition reflects the increasingly important role of this
   distinction in linguistic semantics, in such works as
   \cite{vandersandt:anaphora,kamp/rossdeutscher:tl94}; the particular
   assertions and presuppositions we use draw not only on linguistic
   theory but also on research in connecting linguistic meanings with
   independently-motivated domain representations, such as those
   required for animating human avatars
   \cite{badler:cacm99,badler:agents00}.  Our further specification of
   pragmatic conditions is inspired by accounts of constructions in
   discourse in terms of contextual requirements, such as
   \cite{hirschberg:thesis,ward:thesis,prince:open-prop,birner:thesis,gundel/hz:referring}.

	As an illustration of these threefold conventions, consider
   the item \emph{slide} as used in \sref{instruction-eg} and
   represented in \sref{instruction-deps}.  Here \emph{slide}
   introduces an event $a1$ in which $H$ (the hearer) will \emph{move}
   $N$ (the coupling nut) along a path $P$ (from its current location
   along the surface of the pipe to the elbow); this event is to occur
   \emph{next} in the maintenance procedure.

	At the same time, \emph{slide} provides a presupposed
   constraint that $P$ \emph{start at} the current location of the nut
   $N$ and that $P$ lie along the \emph{surface} of an object.  This
   constraint helps specify what it means for the event to be a
   sliding, but also helps identify both the nut $N$ and the elbow
   $E$.  As an imperative, \ling{slide} carries a presupposed
   constraint on who the \emph{participants} in the conversation are,
   which helps identify the agent $H$ as the hearer, and at the same
   time introduces a variable for the speaker $S$.  Moreover,
   \ling{slide} carries the pragmatic constraint that $S$ be capable
   of imposing \emph{obligations} for physical action on $H$.

	These conditions can be schematized as in
   \sref{slide-conditions}%
\dspftnt{
	From here on, we adopt the abbreviations \m{partic} for
   \m{participants}, \m{surf} for \m{surface}, and \m{obl} for
   \m{obligations}.
}:
\begin{examples}{slide-conditions}
\item
	Assertion: $\m{move}(a1,H,N,P) \wedge \m{next}(a1)$
\item
	Presupposition: $\m{partic}(S,H) \wedge \m{start-at}(P,N)
   \wedge \m{surf}(P)$
\item
	Pragmatics: $\m{obl}(S,H)$
\end{examples}
   	Note that these conditions take the form of constraints on the
   values of variables; this helps explain why we see
   \textsc{description} as central to the problem of sentence
   planning.  We call the variables that appear in such constraints
   the \textsc{discourse anaphors} of an element; we call the values
   those variables take, the element's \textsc{discourse referents}.
   Our terminology follows that of \cite{webber:tense}, where a
   discourse anaphor specifies an entity by relation (perhaps by an
   inferential relation) to a referent represented in an evolving
   model of the discourse.  (Throughout, we follow the Prolog
   convention with anaphors--variables in upper case and
   referents--constants in lower case.)
   
	When elements are combined by syntactic operations, the
   grammar describes both syntactic and semantic relationships among
   them.  Semantic relationships are represented by requiring
   coreference between discourse anaphors of combined elements.  We
   illustrate this by considering the element \emph{coupling-nut},
   which appears in combination with \emph{slide}.  The grammar
   determines that the element presupposes a coupling nut (\emph{cn})
   represented by some discourse anaphor $R$.  The pragmatics of the
   element is the condition that the genre supports the zero definite
   construction (\emph{zero-genre}) and that the referent for $R$ has
   definite status in the conversational record.  The element carries
   no assertion.  Thus, this use of \ling{coupling-nut} carries the
   conditions schematized in \sref{cn-conditions}.
\begin{examples}{cn-conditions}
\item
	Assertion: ---
\item
	Presupposition: $\m{cn}(R)$
\item
	Pragmatics: $\m{def}(R) \wedge \m{zero-genre}$
\end{examples}
	Now, when this element serves as the direct object of the
   element \emph{slide} as specified in \sref{instruction-deps}, the
   coreference constraints of the grammar kick in to specify that what
   is slid must be the coupling nut; formally, in this case, the $N$
   of \sref{slide-conditions} must be the same as the $R$ of
   \sref{cn-conditions}.  Applying this constraint, we would represent
   the conditions imposed jointly by \emph{slide} and
   \emph{coupling-nut} in combination as in
   \sref{slide-cn-conditions}.
\begin{examples}{slide-cn-conditions}
\item
	Assertion: $\m{move}(a1,H,N,P) \wedge \m{next}(a1)$
\item
	Presupposition: $\m{partic}(S,H) \wedge \m{start-at}(P,N)
   \wedge \m{surf}(P) \wedge \m{cn}(N)$
\item
	Pragmatics: $\m{obl}(S,H) \wedge \m{def}(N) \wedge \m{zero-genre}$
\end{examples}

	Let us now return to instruction \sref{instruction-eg}.
\rsref{\myinstruction}
	In all, our exposition in this paper represents the content of
   \sref{instruction-eg} with the three collections of constraints on
   discourse anaphors in \sref{all-conditions}; we associate
   \sref{all-conditions} with \sref{instruction-eg} through the
   derivation of \sref{instruction-eg} as tree \sref{instruction-deps}
   in our grammar for English.
\begin{examples}{all-conditions}
\item
	Assertion: $\m{move}(a1,H,N,P) \wedge \m{next}(a1) \wedge
   \m{purpose}(a1,a2) \wedge \m{uncover}(a2,H,R)$
\item
	Presupposition: $\m{partic}(S,H) \wedge \m{start-at}(P,N)
   \wedge \m{surf}(P) \wedge \m{cn}(N) \wedge \m{end-on}(P,E) \wedge
   \m{el}(E) \wedge \m{sr}(R) \wedge \m{fl}(F) \wedge \m{nn}(R,F,X)$
\item
	Pragmatics: $\m{obl}(S,H) \wedge \m{def}(N) \wedge \m{def}(E)
   \wedge \m{def}(R) \wedge \m{def}(F) \wedge \m{zero-genre}$
\end{examples}
	Spelling out the example in more detail, we see that in
   addition to the asserted constraints \emph{move} and \emph{next}
   contributed by the element \emph{slide}, we have a \emph{purpose}
   constraint contributed by the bare infinitival adjunct and an
   \emph{uncover} constraint contributed by the element
   \emph{uncover}; in addition to the presupposed constraints
   \emph{partic}, \m{start-at}, \emph{surf} and \emph{cn} contributed
   by \emph{slide} and \emph{coupling-nut}, we have an \emph{end-on}
   constraint contributed by \emph{onto}, an \emph{el} constraint
   contributed by \emph{elbow}, an \emph{sr} constraint contributed by
   \emph{sealing-ring} and \m{fl} and \m{nn} constraints contributed
   by the noun-noun modifier use of \emph{fuel-line}; \m{nn} uses a
   variable $X$ to abstract some close relationship between the fuel
   line $F$ and the sealing ring $R$ which grounds the noun-noun
   compound.

   	In any use of an utterance like \sref{instruction-eg}, the
   speaker intends the presupposition and the pragmatics of the
   utterance to link up in a specific way with particular individuals
   and propositions from the conversational record; the speaker
   likewise intends the assertion to settle particular open questions
   in the discourse in virtue of the information it presents about
   particular individuals.  These links constitute the
   \textsc{presumptions} the speaker makes with an utterance; these
   presumptions must be recorded in an interpretation over and above
   the shared conventions that we have already outlined.  We assume
   that these presumptions take the form of \textsc{inferences} that
   the speaker is committed to in generation and that the hearer must
   recover in understanding.  

	We return to the element \emph{slide} of
   \sref{instruction-deps} to illustrate this ingredient of
   interpretation.  We take the speaker of \sref{instruction-eg} to be
   a computer system (including an NLG component), which represents
   itself as a conversational participant $s0$ and represents its user
   as a conversational participant $h0$.  We suppose that the coupling
   nut to be moved here is identified as $n11$ in the system's model
   of the aircraft, the fuel-line joint is identified as $j2$ and the
   elbow is identified as $e2$.  In order to describe paths, we use a
   function $l$ whose arguments are a landmark and a spatial relation
   and whose result is the place so-related to the landmark.  For
   example, $l(\m{on},e2)$ is the place \emph{on the elbow}.  We also
   use a function $p$ whose arguments are two places and whose result
   is the direct path between them.  For example,
   $p(l(\m{on},j2),l(\m{on},e2))$ is the path that the coupling nut
   follows here.  (For a similar spatial ontology, see
   \cite{jackendoff:structures}.)  Then the system here intends the
   contribution that the next action, $a1$, is one where $h0$ moves
   $n11$ by path $p(l(\m{on},j2),l(\m{on},e2))$.  This contribution
   follows by inference from the meaning of \emph{slide} in general
   together with the speaker's commitments to pick out particular
   discourse referents from the conversational record and, where
   necessary, to rely on background knowledge about these referents
   and about aircraft maintenance in general.

	Let's adopt the notation that a boxed expression represents an
   update to be made to the conversational record, while an underlined
   expression represents a feature already present in the
   conversational record; boxed and underlined expressions are
   \term{domain} representations and can be specialized, when
   appropriate, to application-specific ontologies and models.  The
   other expressions we have seen are \term{linguistic}
   representations, since they are associated with lexical items and
   syntactic constructions in a general way.  An edge indicates an
   inferential connection between a linguistic representation and a
   domain representation.  Then we can provide representations of the
   presumption associated with the assertion of \emph{slide} in
   \sref{instruction-eg} by \sref{slide-assert-inference}.
\sentence{slide-assert-inference}{
	\mbox{\small
	\leaf{$\m{move}(a1,H,N,P)$}
	\branch{1}{\goalinst{\m{move}(a1,h0,n11,p(l(\m{on},j2),l(\m{on},e2)))}}
	\tree}
	\mbox{\small
	\leaf{$\m{next}(a1)$}
	\branch{1}{\goalinst{\m{next}(a1)}}
	\tree}
}

	Given what we have supposed, in uttering
   \sref{instruction-eg}, the system is also committed to inferences
   which establish instances of the presupposition and the pragmatics
   of \emph{slide} for appropriate referents.  Our conventions
   represent these further inferences as in
   \sref{slide-other-inferences}.
\sentence{slide-other-inferences}{
	\mbox{\small
	\leaf{\sharedinst{\m{partic}(s0,h0)}}
	\branch{1}{$\m{partic}(S,H)$}
	\tree}
	\mbox{\small
	\leaf{\sharedinst{\m{start-at}(p(l(\m{on},j2),l(\m{on},e2)),n11)}}
	\branch{1}{$\m{start-at}(P,N)$}
	\tree}
	\mbox{\small
	\leaf{\sharedinst{\m{surf}(p(l(\m{on},j2),l(\m{on},e2)))}}
	\branch{1}{$\m{surf}(P)$}
	\tree}
	\mbox{\small
	\leaf{\sharedinst{\m{obl}(s0,h0)}}
	\branch{1}{$\m{obl}(S,H)$}
	\tree}
}

	In \sref{slide-other-inferences}, we use the same predicates
   for domain and linguistic relationships, so the inferences required
   in all cases can be performed by simple unification.  But our
   framework will enable more complicated (and more substantive)
   connections.  For example, suppose we use a predicate
   $\m{loc}(L,O)$ to indicate that the place $L$ is the location of
   object $O$.  Then we would represent the fact that the nut is
   located on the joint as \sref{loc-assumption}.
\sentence{loc-assumption}{
	$\m{loc}(l(\m{on},j2),n11)$
}
	We know that if an object is in some place, then any path from
   that place begins at the object; \sref{loc-inference} formalizes
   this generalization.
\sentence{loc-inference}{
	$\forall loe(\m{loc}(l,o) \imp \m{start-at}(p(l,e),o))$
}
   	Since they provide common background about this equipment and
   about spatial action in general, both of these facts belong in the
   conversational record.

	From \sref{loc-assumption} and \sref{loc-inference} we can
   infer that the path on the joint starts at the nut; that leads to a
   record of inference as in \sref{loc-inference-tree}.
\sentence{loc-inference-tree}{
	\mbox{\small
	\leaf{\sharedinst{\m{loc}(l(\m{on},j2),n11)}}
	\branch{1}{$\m{start-at}(P,N)$}
	\tree}
}
	That is, the understanding behind \sref{loc-inference-tree} is
   that $\m{loc}(l(\m{on},j2),n11)$ is a fact from the conversational
   record intended to be linked with the linguistic presupposition
   $\m{start-at}(P,N)$ by appeal to the premise \sref{loc-inference}
   from the conversational record.

	Similarly, we propose to analyze the modifier \emph{fuel-line}
   in keeping with the inferential account of noun-noun compounds
   proposed in \cite{hsam:abduction,hobbs:abduction/journal}.  This
   item carries a very general linguistic presupposition.  There must
   be a fuel line $F$ and some close relationship $X$ between $F$ and
   the object $R$ that the modifier applies to.  In the context of
   this aircraft, this presupposition is met because of the fact that
   the particular ring intended here is designed \emph{for} the fuel
   line: $X = \m{for}$.  This link exploits a domain-specific
   inference rule to the effect that one thing's being designed for
   another counts as the right kind of close relationship for
   noun-noun modification.  Concretely, we might use this structure to
   abstract the inference:
\sentence{inference-tree}{
	\mbox{\small
	\leaf{\underline{$\m{for}(r11,f4)$}}
	\branch{1}{$\m{nn}(R,F,X)$}
	\tree}
}
	As with \sref{loc-inference-tree}, \sref{inference-tree}
   represents that $\m{for}(r11,f4)$ is a shared fact linked with the
   linguistic presupposition $\m{nn}(R,F,X)$ by appeal to a shared
   rule, here \sref{inference-rule}.
\sentence{inference-rule}{
	$\forall ab (\m{for}(a,b) \imp \m{nn}(a,b,\m{for}))$
}

\begin{figure}
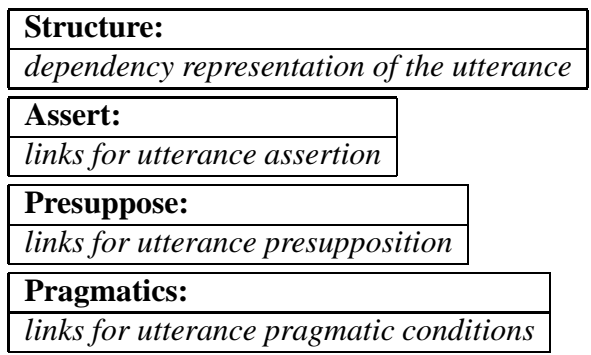

\begin{center}
\begin{tabular}{l}
\lxstructure{\emph{dependency representation of the utterance}} \\[2ex]
\alinks{\emph{links for utterance assertion}} \\[2ex]
\pslinks{\emph{links for utterance presupposition}} \\[2ex]
\pglinks{\emph{links for utterance pragmatic conditions}}
\end{tabular}
\end{center}
\caption{General form of communicative intent representation.}
\label{rep-format-eg}
\end{figure}

\begin{figure}
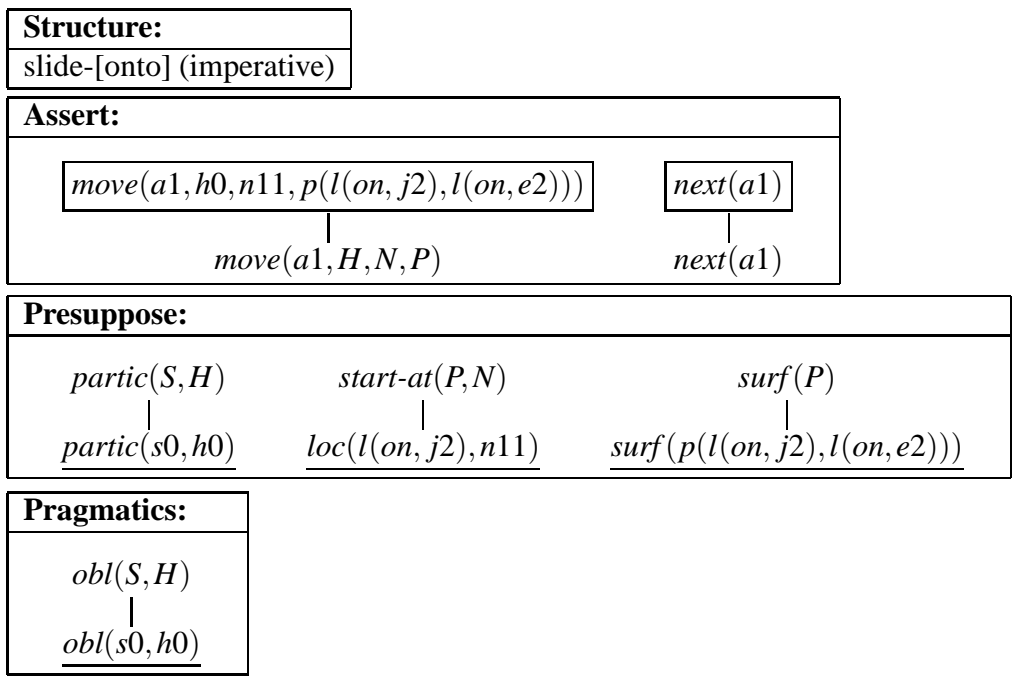

\begin{center}
\begin{tabular}{l}
\lxstructure{slide-[onto] (imperative)} \\[2ex]
\alinkst{
	\mbox{
	\leaf{$\m{move}(a1,H,N,P)$}
	\branch{1}{\goalinst{\m{move}(a1,h0,n11,p(l(\m{on},j2),l(\m{on},e2)))}}
	\tree}
	\mbox{
	\leaf{$\m{next}(a1)$}
	\branch{1}{\goalinst{\m{next}(a1)}}
	\tree}
} \\[6ex]
\pslinkst{
	\mbox{
	\leaf{\sharedinst{\m{partic}(s0,h0)}}
	\branch{1}{$\m{partic}(S,H)$}
	\tree}
	\mbox{
	\leaf{\sharedinst{\m{loc}(l(\m{on},j2),n11)}}
	\branch{1}{$\m{start-at}(P,N)$}
	\tree}
	\mbox{
	\leaf{\sharedinst{\m{surf}(p(l(\m{on},j2),l(\m{on},e2)))}}
	\branch{1}{$\m{surf}(P)$}
	\tree}
} \\[6ex]
\pglinkst{
	\mbox{
	\leaf{\sharedinst{\m{obl}(s0,h0)}}
	\branch{1}{$\m{obl}(S,H)$}
	\tree}
}
\end{tabular}
\end{center}
\caption[Interpretation of \emph{slide} in \sref{instruction-eg}.]{%
Interpretation of \emph{slide} in \sref{instruction-eg}.  The
   speaker's presumptions map out intended connections to discourse
   referents as follows: the speaker $S$, $s0$; the hearer $H$, $h0$;
   the nut $N$, $n11$; the path $P$, $p(l(\m{on},j2),l(\m{on},e2))$;
   the elbow $E$, $e2$.  The fuel-line joint is $j2$.
}
\label{slide-interp-fig}
\end{figure}

	In general, then, the communicative intent behind an utterance
   must include three inferential records.  The first collection of
   inferences links the assertions contributed by utterance elements
   to updates to the conversational record that the instruction is
   intended to achieve; in the case of \sref{slide-assert-inference},
   we add instances of the assertion identified by the speaker.  The
   second collection of inferences links the presuppositions
   contributed by the utterance elements to intended instances in the
   conversational record.  The final collection of inferences links
   the pragmatic constraints of the utterance elements to intended
   instances in the conversational record.  We will represent these
   inferences in the format of Figure~\ref{rep-format-eg}.  Reading
   Figure~\ref{rep-format-eg} from bottom to top, we find a version of
   Clark's ladder of intentions, with higher links dependent on lower
   ones: that is, the inference to pragmatics and presupposition are
   prerequisites for successful interpretation, while the inferences
   from the assertion contingently determine the contribution of
   interpretation.
	Such diagrams constitute a complete record of communicative
   intent, since they include the linguistic structure of the
   utterance and lay out the conventional meanings assigned to this
   structure as well as the presumed inferences linking these meanings
   to context.  For example, Figure~\ref{slide-interp-fig} displays
   the communicative intent associated with the utterance of
   \emph{slide}.

\begin{figure}
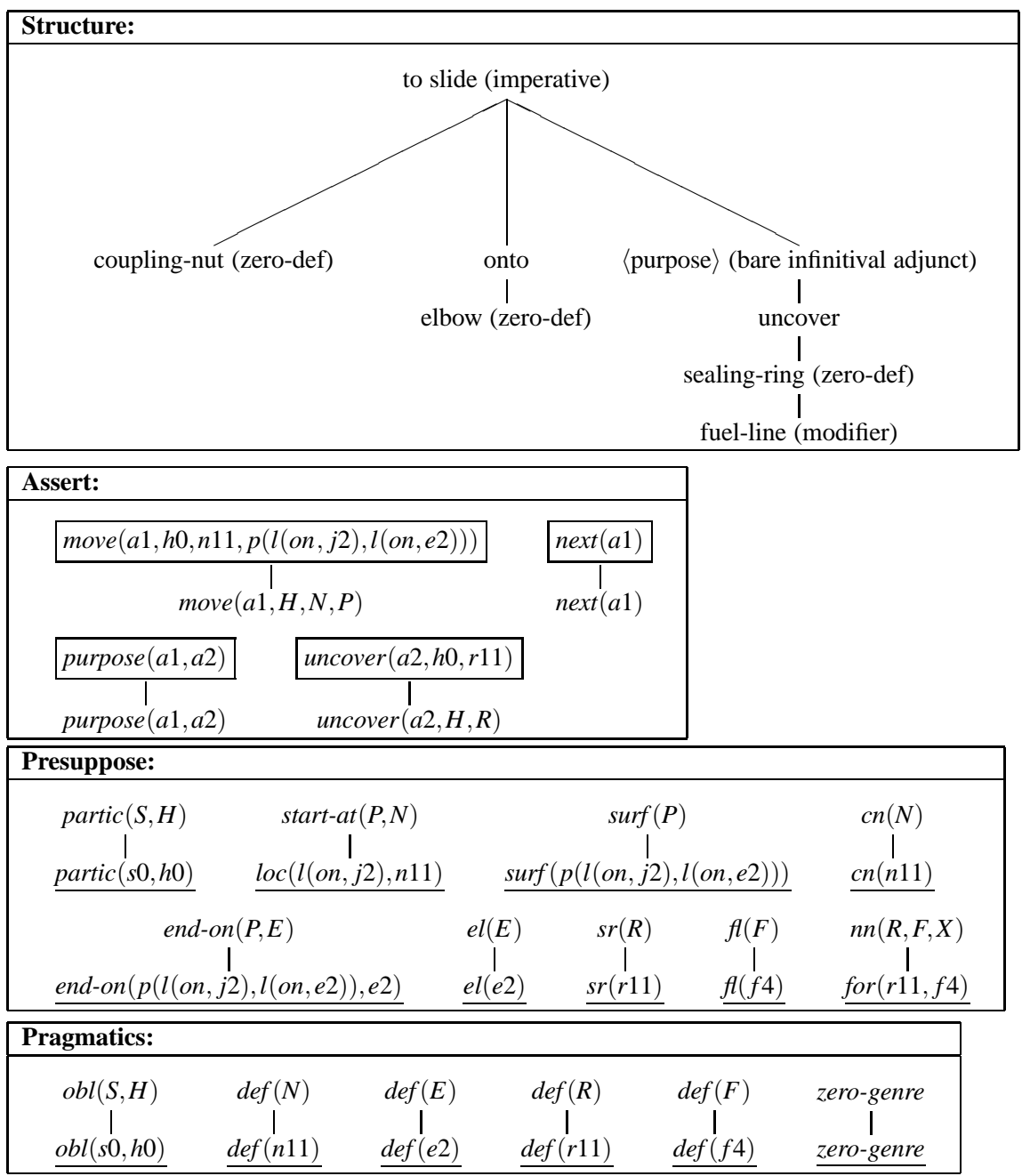

\small
\begin{center}
\begin{tabular}{l}
\lxstructuret{
\mbox{
\leaf{coupling-nut (zero-def)}
\leaf{elbow (zero-def)}
\branch{1}{onto}
\leaf{fuel-line (modifier)}
\branch{1}{sealing-ring (zero-def)}
\branch{1}{uncover}
\branch{1}{$\langle$purpose$\rangle$ (bare infinitival adjunct)}
\branch{3}{to slide (imperative)}
\tree}
} \\[19ex]
\alinkst{
\mbox{
	\leaf{$\m{move}(a1,H,N,P)$}
	\branch{1}{\goalinst{\m{move}(a1,h0,n11,p(l(\m{on},j2),l(\m{on},e2)))}}
	\tree
}
\mbox{
	\leaf{$\m{next}(a1)$}
	\branch{1}{\goalinst{\m{next}(a1)}}
	\tree
} \\[2em]
\mbox{
	\leaf{$\m{purpose}(a1,a2)$}
	\branch{1}{\goalinst{\m{purpose}(a1,a2)}}
	\tree
}
\mbox{
	\leaf{$\m{uncover}(a2,H,R)$}
	\branch{1}{\goalinst{\m{uncover}(a2,h0,r11)}}
	\tree
}} \\[11ex]
\pslinkst{
\mbox{
	\leaf{\sharedinst{\m{partic}(s0,h0)}}
	\branch{1}{$\m{partic}(S,H)$}
	\tree
}
\mbox{
	\leaf{\sharedinst{\m{loc}(l(\m{on},j2),n11)}}
	\branch{1}{$\m{start-at}(P,N)$}
	\tree
}
\mbox{
	\leaf{\sharedinst{\m{surf}(p(l(\m{on},j2),l(\m{on},e2)))}}
	\branch{1}{$\m{surf}(P)$}
	\tree
}
\mbox{
	\leaf{\sharedinst{\m{cn}(n11)}}
	\branch{1}{$\m{cn}(N)$}
	\tree
}
\\[2em]
\mbox{
	\leaf{\sharedinst{\m{end-on}(p(l(\m{on},j2),l(\m{on},e2)),e2)}}
	\branch{1}{$\m{end-on}(P,E)$}
	\tree
}
\mbox{
	\leaf{\sharedinst{\m{el}(e2)}}
	\branch{1}{$\m{el}(E)$}
	\tree
}
\mbox{
	\leaf{\sharedinst{\m{sr}(r11)}}
	\branch{1}{$\m{sr}(R)$}
	\tree
}
\mbox{
	\leaf{\sharedinst{\m{fl}(f4)}}
	\branch{1}{$\m{fl}(F)$}
	\tree
}
\mbox{
	\leaf{\underline{$\m{for}(r11,f4)$}}
	\branch{1}{$\m{nn}(R,F,X)$}
	\tree
}} \\[11ex]
\pglinkst{
\mbox{
	\leaf{\sharedinst{\m{obl}(s0,h0)}}
	\branch{1}{$\m{obl}(S,H)$}
	\tree
}
\mbox{
	\leaf{\sharedinst{\m{def}(n11)}}
	\branch{1}{$\m{def}(N)$}
	\tree
}
\mbox{
	\leaf{\sharedinst{\m{def}(e2)}}
	\branch{1}{$\m{def}(E)$}
	\tree
}
\mbox{
	\leaf{\sharedinst{\m{def}(r11)}}
	\branch{1}{$\m{def}(R)$}
	\tree
}
\mbox{
	\leaf{\sharedinst{\m{def}(f4)}}
	\branch{1}{$\m{def}(F)$}
	\tree
}
\mbox{
	\leaf{\sharedinst{\m{zero-genre}}}
	\branch{1}{$\m{zero-genre}$}
	\tree
}}
\end{tabular}
\end{center}
\caption[Communicative intent for \sref{instruction-eg}.]{%
Communicative intent for \sref{instruction-eg}.  The grammar specifies
   meanings as follows: For \emph{slide}, assertions \m{move} and
   \m{next}; for the bare infinitival adjunct, \m{purpose}; for
   \emph{uncover}, \m{uncover}.  For \emph{slide}, presuppositions
   \m{partic}, \m{start-at} and \m{surf}; for \emph{coupling-nut},
   \m{cn}; for \emph{onto}, \m{end-on}; for \emph{elbow}, \m{el}; for
   \emph{sealing-ring}, \m{cn}; for \emph{fuel-line}, \m{fl} and
   \m{nn}.  For \emph{slide}, pragmatics \m{obl}; for other nouns,
   pragmatics \m{def} and \m{zero-genre}.  The speaker's presumptions
   map out intended connections to discourse referents as follows: the
   speaker $S$, $s0$; the hearer $H$, $h0$; the nut $N$, $n11$; the
   path $P$, $p(l(\m{on},j2),l(\m{on},e2))$; the elbow $E$, $e2$; the
   ring $R$, $r11$; the fuel-line $F$, $f4$; the relation $X$,
   $\m{for}$.  The fuel-line joint is $j2$.
}
\label{instruct-intent-fig}
\end{figure}

	Figure~\ref{instruct-intent-fig} schematizes the full
   communicative intent for \sref{instruction-eg} using the notational
   conventions articulated thus far.
	As a whole, the utterance carries the syntactic structure of
   \sref{instruction-deps}; in Figure~\ref{instruct-intent-fig} this
   structure is paired with inferential representations that simply
   group together the inferences involved in interpreting the
   individual words in their specific syntactic constructions.  

\subsection{Reasoning with Communicative Intent in Conversation}
\label{rep-use-intro-subsec}   

\begin{figure}
\begin{center}
\mbox{\psfig{figure=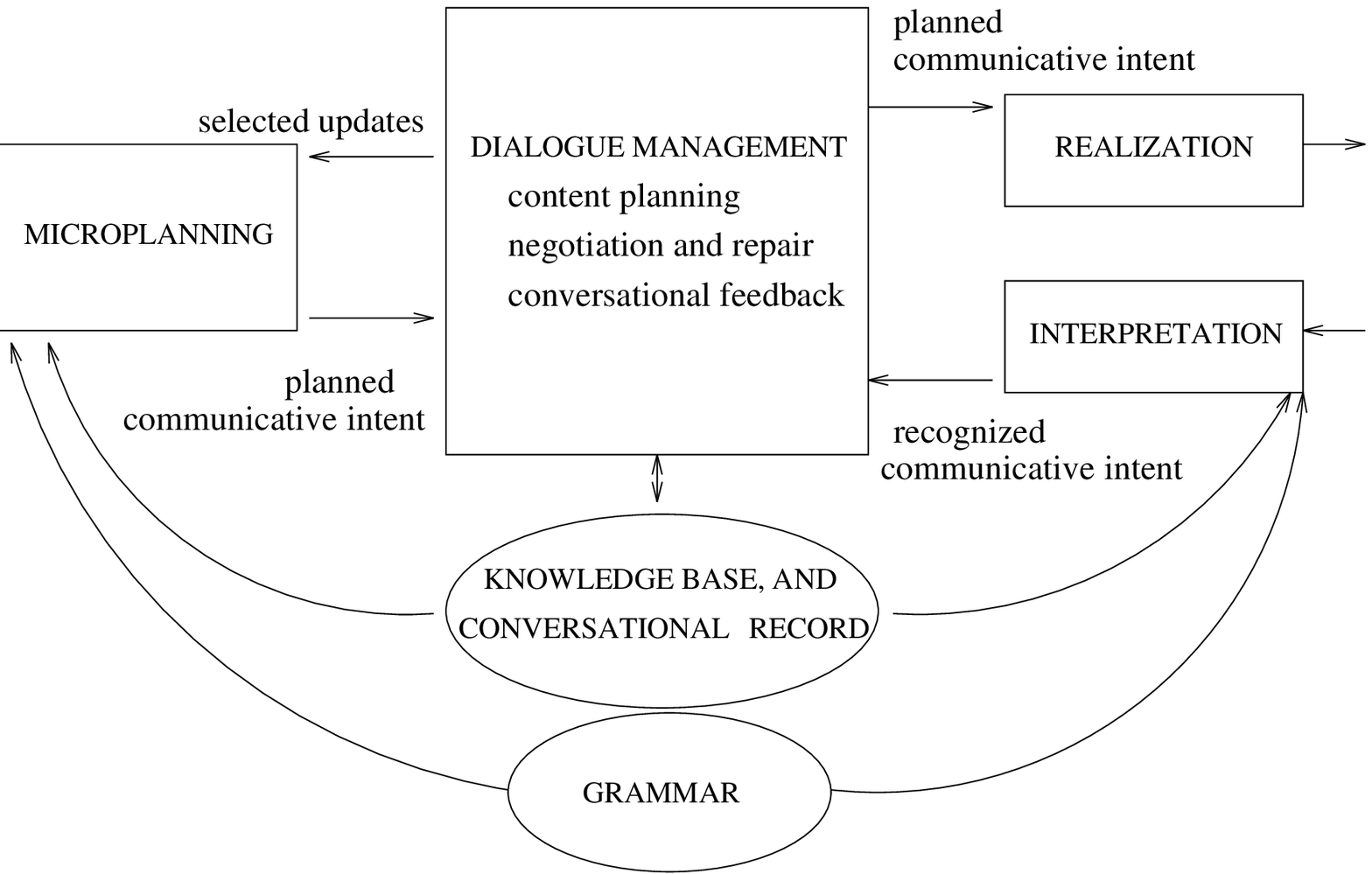,scale=.85,silent=}}
\end{center}
\caption{A conversational architecture for communicative-intent--based
   microplanning.}
\label{conversation-arch-fig}
\end{figure}

	We now return to our initial characterization of conversation
   as a complex collaborative and deliberative process, guided by
   representations of communicative intent such as that of
   Figure~\ref{instruct-intent-fig}.  This characterization locates
   microplanning within the architecture depicted in
   Figure~\ref{conversation-arch-fig}.

   	In Figure~\ref{conversation-arch-fig}, content planning is one
   of a number of subtasks carried out by a general dialogue manager.
   The dialogue manager tracks the content of conversation through
   successive turns, through such functions as following up on an
   utterance \cite{moore/paris:cl/planning,j.moore:phd}, repairing an
   utterance \cite{heeman/hirst:collab}, and updating a model of the
   ongoing collaboration \cite{rich:ai01}.  The dialogue manager also
   coordinates the interaction in the conversation, by managing
   turn-taking, acknowledgment and other conversational signals
   \cite{cassell:cacm}.
   
	Once content planning has derived some updates that need to be
   made to the conversational record, the dialogue manager passes
   these updates as input to the microplanning module.  In response,
   the microplanner derives a communicative-intent representation that
   spells out a way to achieve this update using an utterance of
   concrete linguistic forms.  To construct this representation, the
   microplanner consults both the grammar and a general
   \textsc{knowledge base}.  This knowledge base specifies the
   system's private domain knowledge, as well as background
   information about the domain that all participants in the
   conversation are presumed to share.  It maintains information
   conveyed in the conversation, thus including and extending the
   system's model of the conversational record.

	The output communicative intent constructed by the
   microplanner returns to the dialogue manager; the dialogue manager
   not only can forward this communicative intent for realization but
   also can use it as a general resource for collaboration.  Thus,
   Figure~\ref{conversation-arch-fig} reproduces and extends the NLG
   pipeline of Figure~\ref{arch-fig}.  \cite{cassell:inlg00} describes
   more fully the integration of dialogue management and
   communicative-intent--based microplanning in one implemented
   conversational agent.

	In a communicative-intent representation, as illustrated in
   the structure of Figure~\ref{instruct-intent-fig}, we find the
   resources required for a flexible dialogue manager to pursue
   instruction \sref{instruction-eg} with an engaged conversational
   partner.  To start with, the structure is a self-contained record
   of what the system is doing with this utterance and how it is doing
   it.  The structure maps out the contributions that the system wants
   on the record and the assertions that signal these contributions;
   it maps out the constraints presupposed by the utterance and the
   unique matches for these constraints that determine the referents
   the instruction has.  Because the structure combines grammatical
   knowledge and information from the conversational record in this
   unambiguous way, the dialogue manager can utter it with the
   expectation that the utterance will be understood (provided the
   model of the conversational record is correct and provided the
   interpretation process does not demand more effort than the user is
   willing or able to devote to it).    

	More generally, we expect that communicative-intent
   representations offer a resource for the dialogue manager to
   respond to future utterances.  Although we have yet to implement
   such deliberation, let us outline briefly how communicative intent
   may inform such responses; such considerations help to situate
   structures such as that of Figure~\ref{instruct-intent-fig} more
   tightly within our general characterization of conversation.

	As a first illustration, suppose the user asks a
   clarification question about the instructed action, such as
   \sref{clarification}.
\sentence{clarification}{
	So I want to get at the sealing-ring at the joint under the
   coupling-nut?
}
	By connecting the communicative intent from
   \sref{instruction-eg} with the communicative intent recognized for
   \sref{clarification}, the dialogue manager can infer that the actor
   is uncertain about which sealing-ring the system intended to
   identify with \ling{fuel-line sealing-ring}.  In carrying out this
   inference and in formulating an appropriate answer (\ling{that's
   right}, perhaps), the explicit links in communicative intent
   between presupposed content and the conversational record are
   central.  In other words, the dialogue manager can use
   communicative intent as a data structure for plan recognition and
   plan revision in negotiating referring expressions, as in
   \cite{heeman/hirst:collab}.

	As a second illustration, suppose the user asks a follow-up
   question about the instructed action, perhaps \sref{followup}.
\sentence{followup}{
	How does that uncover the sealing ring?
}
	\sref{followup} refers to the sliding and the uncovering
   introduced by \sref{instruction-eg}; in fact, \sref{followup}
   shares with \sref{instruction-eg} not only reference but also
   substantial vocabulary.  Accordingly, by connecting the intent
   behind \sref{followup} to that for \sref{instruction-eg}, the
   dialogue manager may infer that the intent for
   \sref{instruction-eg} was successfully recognized.  At the same
   time, by comparing the intent for \sref{followup} with that for
   \sref{instruction-eg}, the dialogue manager can discover that,
   because the actor needs to know how the sliding will achieve the
   current purpose, the actor has not fully accepted instruction
   \sref{instruction-eg}.  The information provided in
   \sref{instruction-eg} and \sref{followup} can serve as a starting
   point for repair: knowing what information the actor has narrows
   what information the user might need.  More generally, if
   structures for communicative intent also record the inferential
   relationships that link communicative goals to one another, the
   dialogue manager may attempt the more nuanced responses to
   expressions of doubt and disagreement described in
   \cite{moore/paris:cl/planning,carberry/lambert:cl99}.

	With this background, we can now present the key idea behind
   the \spud\ system: The structure of
   Figure~\ref{instruct-intent-fig} provides a resource for
   deliberation not just for the dialogue manager but also for the
   microplanner itself.  The microplanner starts with a task set by
   the dialogue manager: this utterance is to contribute, in a
   recognizable way, the updates that a \ling{move} is \ling{next} and
   its \ling{purpose} is to \ling{uncover}.  The microplanner can see
   to it that its utterance satisfies these requirements by adding
   interpreted elements, such as the structure for \emph{slide} of
   Figure~\ref{slide-interp-fig}, one at a time, to a provisional
   communicative-intent representation.  In each of these steps, the
   microplanner can use its assessment of the overall interpretation
   of the utterance to make progress on the interrelated problems of
   lexical choice, aggregation and referring expression generation.

\begin{figure}
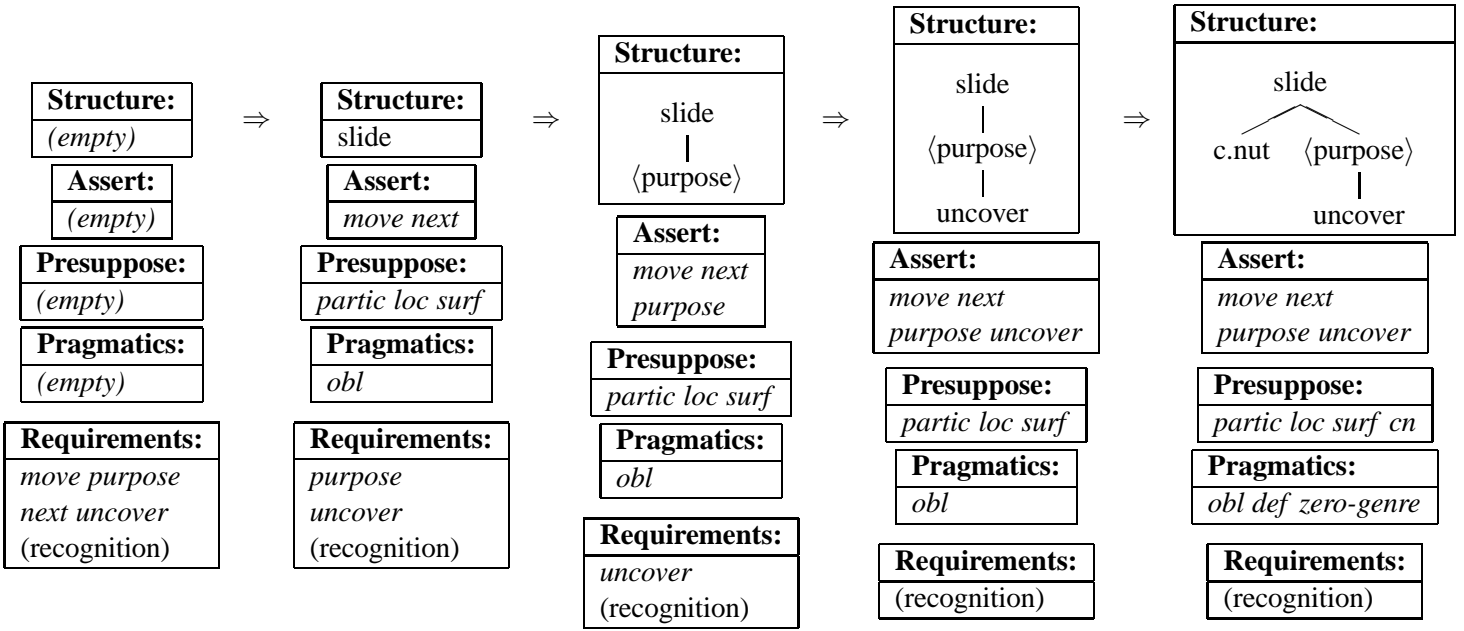

\hspace*{\fill}\makebox[0in]{
\small
\begin{tabular}[t]{c}
\lxstructure{\emph{(empty)}} \\[2ex]
\alinks{\emph{(empty)}} \\[2ex]
\pslinks{\emph{(empty)}} \\[2ex]
\pglinks{\emph{(empty)}} \\[3ex]
\thingstodo{\ling{move} \ling{purpose} \\ \ling{next} \ling{uncover}
   \\ (recognition)}
\end{tabular}
	$\Rightarrow$
\begin{tabular}[t]{c}
\lxstructure{slide} \\[2ex]
\alinks{\ling{move} \ling{next}} \\[2ex]
\pslinks{\ling{partic} \ling{loc} \ling{surf}} \\[2ex]
\pglinks{\ling{obl}} \\[3ex]
\thingstodo{\ling{purpose} \\ \ling{uncover} \\ (recognition)}
\end{tabular}
	$\Rightarrow$
\begin{tabular}[t]{c}
\lxstructuret{\mbox{\leaf{$\langle$purpose$\rangle$}\branch{1}{slide}\tree}} \\[6ex]
\alinks{\ling{move} \ling{next} \\ \ling{purpose}} \\[4ex]
\pslinks{\ling{partic} \ling{loc} \ling{surf}} \\[2ex]
\pglinks{\ling{obl}} \\[3ex]
\thingstodo{\ling{uncover} \\ (recognition)}
\end{tabular}
	$\Rightarrow$
\begin{tabular}[t]{c}
\lxstructuret{\mbox{\leaf{uncover}\branch{1}{$\langle$purpose$\rangle$}\branch{1}{slide}\tree}} \\[8ex]
\alinks{\ling{move} \ling{next} \\ \ling{purpose} \ling{uncover}} \\[4ex]
\pslinks{\ling{partic} \ling{loc} \ling{surf}} \\[2ex]
\pglinks{\ling{obl}} \\[3ex]
\thingstodo{(recognition)}
\end{tabular}
	$\Rightarrow$
\begin{tabular}[t]{c}
\lxstructuret{\mbox{\leaf{c.nut}\leaf{uncover}\branch{1}{$\langle$purpose$\rangle$}
	\branch{2}{slide}\tree}} \\[8ex]
\alinks{\ling{move} \ling{next} \\ \ling{purpose} \ling{uncover}} \\[4ex]
\pslinks{\ling{partic} \ling{loc} \ling{surf} \ling{cn}} \\[2ex]
\pglinks{\ling{obl} \ling{def} \ling{zero-genre}} \\[3ex]
\thingstodo{(recognition)}
\end{tabular}}
\hspace*{\fill}
\caption[Stages of Microplanning]{%
A schematic view of the initial stages of microplanning for
   \sref{instruction-eg}.  Each state includes a provisional
   communicative intent and an assessment of further work required,
   such as updates to achieve.  Each transition represents the
   addition of a new interpreted element.}
\label{nlg-steps-fig}
\end{figure}

	Figure~\ref{nlg-steps-fig} offers a schematic illustration of
   a few such steps: it tracks the addition first of \ling{slide},
   then of a purpose adjunct, then of \ling{uncover}, and finally of
   \ling{coupling-nut}, all to an initially empty structure.  (Note
   that in Figure~\ref{nlg-steps-fig} we abbreviate inference
   structures and specified updates to the predicates they establish;
   we use the tag \ling{recognition} as a mnemonic that the
   microplanner is responsible for making sure these structures can be
   recognized as intended.)

	To start, the first transition in Figure~\ref{nlg-steps-fig},
   which results in a structure that repeats
   Figure~\ref{slide-interp-fig}, can be viewed as a description of
   the use of the particular word \emph{slide} in a particular
   syntactic construction to achieve particular effects.  We will see
   that a generator can create such descriptions by an inferential
   matching process that checks a pattern of lexical meaning against
   the discourse context and against the specified updates.  In
   particular, to be applicable at a specific stage of generation, a
   lexical item must have an interpretation to contribute: the item's
   assertion must hold; the item's presupposition and pragmatics must
   find links in the conversational record.  Moreover, to be prefered
   over alternative options, use of the item should push the
   generation task forward: in general, the updates the item achieves
   should include as many as possible of those specified in the
   microplanning problem, and as few others as possible; in general,
   the links the item establishes to shared context should appeal to
   specific shared content that facilitates the hearer's
   plan-recognition interpretation process.

	Thus, in deriving structures like that of
   Figure~\ref{slide-interp-fig} from its grammatical inventory, the
   generator can implement a model of lexical and grammatical choice.
   The generator determines available options by inference and selects
   among alternatives by comparing interpretations.  

	Meanwhile, in extending provisional communicative intent as
   suggested in Figure~\ref{nlg-steps-fig}, the generator's further
   lexical and syntactic choices can simultaneously reflect the its
   strategies for aggregation and for referring expression generation.
   Take the addition of an element like the bare infinitival purpose
   clause, in step two of Figure~\ref{nlg-steps-fig}.  As with
   \ling{slide}, this entry represents a pattern of interpretation
   where linguistic meaning mediates between the current context and
   potential update to the context.  In particular, the entry for a
   bare infinitival purpose clause depends on an event $a1$ with an
   agent $h0$ already described by the main verb of the provisional
   instruction (in this case \emph{slide}).  The entry relates $a1$ to
   another event $a2$ which $a1$ should achieve and which also has
   $h0$ as the agent; here $a2$ is to be described as an
   \emph{uncovering} by a subsequent step of lexical choice.  Thus the
   syntax and semantics of the entry amount to a pattern for
   aggregation: the modifier provides a way of extending an utterance
   that the generator can use to include additional related
   information about referents already described in the ongoing
   utterance.

	As another illustration, take the addition of a complement
   like \emph{coupling nut}, as in step four of
   Figure~\ref{nlg-steps-fig}, or a modifier like \emph{fuel-line}.
   The contribution of these entries is to add constraints on the
   context that the hearer must match to interpret the utterance.
   With \emph{coupling nut}, for example, the hearer learns that the
   referent for $N$ must actually be a coupling nut; similarly, with
   \emph{fuel-line}, the hearer learns that the referent for $R$ must
   be for some fuel line $F$.  Here we find the usual means for
   ensuring reference in NLG: augmenting the content of an utterance
   by additional presupposed relationships.

\subsection{Communicative-Intent--Based Microplanning in \spud}
\label{spud-intro-subsec}

   	Sections~\ref{converse-intro-subsec}--\ref{rep-use-intro-subsec}
   have characterized microplanning as a problem of constructing
   representations of communicative intent to realize communicative
   goals.  Communicative intent is a detailed representation of an
   utterance that combines inferences from a declarative description
   of language, the grammar, and from a declarative description of
   context, the conversational record.  This representation supports
   the reasoning required for a dialogue manager to produce, support
   and defend the generated utterance as part of a broader
   conversational process.  At the same time, by setting up
   appropriate microplanning choices and providing the means to make
   them, this representation reconciles the decision-making required
   for microplanning tasks like lexical choice, referring expression
   generation and aggregation.

   	Our characterization of sentence planning is not so far from
   Appelt's \cite{appelt:planning}.  One difference is that Appelt
   takes a speech-act view of communicative action, so that
   communicative intent is not an abstract resource for conversational
   process but a veridical inference about the dynamics of agents'
   mental state; this complicates Appelt's representations and
   restricts the flexibility of his system.  Closer still is the work
   of Thomason and colleagues
   \cite{thomason/hobbs/moore,thomason/hobbs:abduction} in the
   interpretation-as-abduction framework
   \cite{hobbs:abduction/journal}; they construct abductive
   interpretations as an abstract representation of communicative
   intent, by reasoning from a grammar and from domain knowledge.

   	A key contribution of our research, over and above these
   antecedents, is the integration of a suite of assumptions and
   techniques for effective implementation and development of
   communicative-intent--based microplanners.
\begin{itemize}
\item
	We use the feature-based lexicalized tree-adjoining grammar
   formalism (LTAG) to describe microplanning derivations
   \cite{joshi:tag,schabes90}.  Each choice that arises in using this
   grammar for generation realizes a specified meaning by concrete
   material that could be added to an incomplete sentence, as
   advocated by \cite{joshi87-gen} and anticipated already in
   Section~\ref{rep-intro-subsec}.  In fact, LTAG offers this space of
   choices directly on the derivation of surface syntactic structures,
   eliminating any need for ``abstract'' linguistic structures or
   resources.

\item   
	We use a logic-programming strategy to link linguistic
   meanings with specifications of the conversational record and
   updates to it.  We base our specification language on modal logic
   in order to describe the different states of information in the
   context explicitly \cite{stone:modal/lp,jlac00}; however, the logic
   programming inference ensures that a designer can assess and
   improve the computational cost of the queries involved in
   constructing communicative intent.

\item
	By treating presuppositions as anaphors
   (cf.~\cite{vandersandt:anaphora}), we carry over efficient
   constraint-satisfaction techniques for managing ambiguity in
   referring expressions from prior generation research
   \cite{mellish:descriptions,haddock:thesis,dale/haddock:referring}.

\item
	We associate grammatical entries with pragmatic constraints on
   context that model the different discourse functions of different
   constructions \cite{ward:thesis,prince:open-prop}.  This provides
   both a principled model of syntactic choice and a declarative
   language for controlling the output of the system to match the
   choices observed in a given corpus or sublanguage.

\item
	We adopt a head-first, greedy search strategy.  Our other
   principles are compatible with searching among all partial
   representations of communicative intent, in any order.  But a
   head-first strategy allows for a particularly clean implementation
   of grammatical operations; and the modest effort required to design
   specifications for greedy search is easily repaid by improved
   system performance.

\end{itemize}
   	Although many of these techniques have seen success in recent
   generation systems, \spud's distinctive focus on communicative
   intent results in basic and important divergences from other
   systems; we return to a more thorough review of previous work in
   Section~\ref{related-work-section}.

	In the remainder of this paper, we first describe the grammar
   formalism we have developed and the model of interpretation that
   associates grammatical structures declaratively with possible
   communicative intent.  We then introduce the \spud\ sentence
   planner as a program that searches (greedily) through grammatical
   structures to derive a communicative intent representation that
   describes a desired update to the conversational record and that
   can be recognized by the hearer.  We go on to illustrate how
   \spud's declarative processing provides a natural framework for
   addressing sentence planning subtasks like referring expression
   generation, lexical and syntactic choice and aggregation, and how
   it supports a concrete methodology for building grammatical
   resources for specific generation problems.

\section{Grammar Organization}
\label{grammar-section}

   	In \spud, a grammar consists of a set of \term{syntactic
   constructions}, a set of \term{lexical entries}, and a database of
   \term{morphological rules}.
   
\subsection{Syntactic Constructions}
\label{const-subsec}

	Syntactic constructions are specified by four components in
   \spud:
\begin{examples}{const-general}
\item
	a \term{name}, an identifier under which other parts of the
   grammar refer to the construction;
\item
	a set of \term{parameters}, open variables for referential
   indices in the definition (which are instantiated to discourse
   referents in a particular use of the construction);
\item
	a \term{pragmatic condition}, which expresses a constraint
   that the construction imposes on the discourse context in terms of
   its parameters; and
\item
	a \term{syntactic structure}, which maps out the linguistic
   form of the construction.
\end{examples}
	The syntactic structure is represented as a tree of compound
   nodes.  Internal nodes in the tree bear the following attributes:
\begin{examples}{tree-internal-general}
\item
	a \term{category}, such as \term{np}, \term{v}, etc.;
\item
	\term{indices}, a list of the parameters that the node refers
   to and that additional syntactic material combined with this node
   may describe;
\item
	a \term{top feature structure}, a list of attribute-value
   pairs (including variable values shared with other feature
   structures elsewhere in the tree) which describes the syntactic
   constraints imposed on this node from above; and
\item
	a \term{bottom feature structure}, another such list of
   attribute-value pairs which describes the syntactic constraints
   imposed on this node from below.
\end{examples}
	Leaves in the tree fall into one of four classes:
   \term{substitution sites}, \term{foot nodes}, \term{given-word
   nodes} and lexically-dependent word nodes or \term{anchor nodes}.
   Like internal nodes, substitution sites and foot nodes are loci of
   syntactic operations and are associated with categories and
   indices.  Any tree may have at most one foot node, and that foot
   node must have the same category and indices as the root.  A
   given-word node includes a specific lexeme (typically a
   closed-class or function item) which appears explicitly in all uses
   of the construction.  An anchor node is associated with an
   instruction to include a word retrieved from a specific lexical
   entry; trees may have multiple anchors and lexical entries may
   contain multiple words.  In addition, all leaves are specified with
   a single feature structure which describes the constraints imposed
   on the node from above.  Note that, in the case of anchor nodes,
   these constraints must be satisfied by the lexical items retrieved
   for the node.
	
	\sref{ctree-long} shows the tree structure for the zero
   definite noun phrase required in \sref{instruction-eg} for
   \emph{coupling nut} and \emph{sealing ring}.
\sentence{ctree-long}{
	\mbox{
	\leaf{\fbox{$
	      \begin{array}{l}
		\term{anchor}:\#1 \\
		\term{features}: [\term{number}: \fbox{1}]
	      \end{array}
	      $}}
	\branch{1}{\fbox{$
	      \begin{array}{l}
		\term{cat}:\term n \\
		\term{indices}: U \\
		\term{top}:  [\term{number}: \fbox{1}] \\
		\term{bottom}:  [\term{number}: \fbox{1}] 
	      \end{array}
	      $}}
	\branch{1}{\fbox{$
	      \begin{array}{l}
		\term{cat}:\term{np} \\
		\term{indices}: U \\
		\term{top}:  [\term{number}: \fbox{1}] \\
		\term{bottom}:  [\term{number}: \fbox{1}] 
	      \end{array}
	      $}}
	\tree}
}
	Evidently such structures, and the full specifications
   associated with them, can be quite involved.  For exposition,
   henceforth we will generally suppress feature structures.  We will
   write internal nodes in the form $\term{cat}(\term{indices})$;
   anchors, in the form $\term{cat}\Diamond\term{n}$ (for the
   \term{n}th token of a lexical item, a word of category \term{cat});
   substitution nodes, in the form
   $\term{cat}(\term{indices})\downarrow$; foot nodes, in the form
   $\term{cat}(\term{indices})\footnodemark$; and given-word nodes
   just by the words associated with them.
   
	With these conventions, the syntactic entry for the zero
   definite construction (associated with \ling{sealing-ring} for
   example) is given in \sref{const-example}.
\begin{examples}{const-example}
\item
	\term{name}: zerodefnptree
\item
	\term{parameters}: $U$
\item
	\term{pragmatics}: $\m{zero-genre} \wedge \m{def}(U)$
\exitem{cetree}
	\term{tree}: 
	\mbox{
	\leaf{$\term{n}\Diamond$1}
	\branch{1}{$\term{n}'(U)$}
	\branch{1}{$\term{np}(U)$}
	\tree}
\end{examples}
	Observe that \sref{ctree-long} appears simply as
   \msref{const-example}{cetree}.

\subsection{Lexical Entries}
\label{lex-subsec}

	\spud\ lexical entries have the following structure.
\begin{examples}{lex-general}
\item
	a \term{name}, a list of the lexemes that anchor the entry
   (most entries have only one lexeme, but entries for idioms may
   have several);
\item
	a set of \term{parameters}, open variables for referential
   indices in the definition (which are instantiated to discourse
   referents in a particular use of the entry);
\item
	a \term{target}, an expression constraining the category and indices
   of the node in a syntactic structure at which this lexical entry could
   be incorporated, and indicating whether the entry is added as a
   complement or as a modifier;
\item
	a \term{content condition}, a formula specifying a constraint
   on the parameters of the entry that the entry will assert when the
   entry is used to update the conversational record;
\item
	\term{presupposition}, a formula specifying a constraint on the
   parameters of the entry that the entry must presuppose;
\item
	\term{pragmatics}, a formula specifying a constraint on the status
   in the discourse of parameters of the entry;
\item
	an \term{anchoring feature structure}, a list of attribute-value
   pairs that constrain the anchor nodes where lexical material from this
   entry is inserted into a syntactic construction; and
\item
	a \term{tree list}, specifying the trees that the lexical item
   can anchor by name and parameters (note that the tree list
   in fact determines what the target of the entry must be).
\end{examples}

	\sref{lex-example} gives an example of such a lexical item:
   the entry for \emph{sealing-ring} as used, among other ways, with
   the zero definite noun phrase illustrated in \sref{const-example}.
\begin{examples}{lex-example}
\item
	\term{name}: sealing-ring
\item
	\term{parameters}: $N$
\item
	\term{target}: $\term{np}(N)$ [complement]
\item
	\term{content}: $sr(N)$
\item
	\term{presupposition}: ---
\item
	\term{pragmatics}: ---
\item
	\term{anchor features}: $[\term{number}: \term{singular}]$ 
\item	
	\term{tree list}: $\mbox{zerodefnptree}(N), \ldots$
\end{examples}

\subsection{Lexico-grammar}
\label{lexgram-subsec}

	The basic elements of grammatical derivations are lexical entries
   used in specific syntactic constructions.  These elements are
   declarative combinations of the two kinds of specifications presented in
   Sections~\ref{const-subsec} and~\ref{lex-subsec}.  Abstractly, the
   combination of a lexical entry and a syntactic construction requires the
   following steps.
\begin{examples}{combo-steps}
\item
	The parameters of the lexical entry are instantiated to
   suitable discourse referents.
\item
	The parameters of the construction are instantiated to discourse
   referents as specified by the tree list of the lexical entry.
\item
	Anchor nodes in the tree are replaced by corresponding
   given-word nodes constructed from the name of the lexical entry;
   and the top feature structures of anchor nodes are unified with the
   anchor features of the lexical entry to give the top features of
   the new given-word nodes.
\item
	The assertion and the presupposition of the combined entry are
   determined, in one of two possible ways.  In one possible case, the
   content condition of the lexical entry provides the assertion while
   the presupposition of the lexical entry provides the presupposition
   of the combined element.  In the other, the content condition and
   any presupposition of the lexical entry are conjoined to give the
   presupposition of the combined element; in this case the element
   carries no assertion.
\item
	The pragmatics of the syntactic construction is conjoined with the
   pragmatics of the lexical entry.
\end{examples}
	Thus, abstractly, we can see the syntactic construction of
   \sref{const-example} coming together with the lexical entry of
   \sref{lex-example} to yield the particular lexico-grammatical option
   described in \sref{combo-example}.
\begin{examples}{combo-example}
\exitem{cet}
	\term{tree}: \mbox{
	\leaf{sealing-ring}
	\branch{1}{$\term{n}'(R)$}
	\branch{1}{$\term{np}(R)$}
	\tree}
\item
	\term{target}: $\term{np}(R)$ [complement] 
\item
	\term{assertion}: ---
\item
	\term{presupposition}: $\m{sr}(R)$
\item
	\term{pragmatics}: $\m{def}(R) \wedge \m{zero-genre}$
\end{examples}
	(Again, feature structures are suppressed here, but note that
   feature sharing ensures that each of the nodes in the tree is in
   fact marked with singular number.)  This is the entry for
   \ling{sealing-ring} which is used in deriving the communicative
   intent of Figure~\ref{instruct-intent-fig}.

\subsection{Morphological rules}

	We have seen that lexico-grammatical entries such as
   \sref{combo-example} contain not specific surface word-forms but
   merely lexemes labeled with features.  This allows feature-values
   to be propagated through grammatical derivations.  In this way, the
   derivation can select an appropriate realization for an underlying
   lexeme as a function of agreement processes in the language.

	A database of morphological rules accomplishes this selection.
   Each lexeme is paired with a list of feature-realization patterns.
   To determine the form to use in realizing a given lexeme at a node
   with given features $F$ in a grammatical derivation, \spud\ scans
   this list until the feature structure in a pattern subsumes $F$;
   \spud\ uses the realization associated with this pattern.

	For example, we might use \sref{realize-example} to determine the
   realization of \m{sealing-ring} in \sref{combo-example} as ``sealing
   ring''.
\begin{examples}{realize-example}
\item
	\term{lexeme}: sealing-ring; \term{patterns}:
\item
	$[\term{number}: \term{singular}] \rightarrow$ sealing ring
\item
	$[\term{number}: \term{plural}] \rightarrow$ sealing rings
\end{examples}

\section{Grammatical Derivation and Communicative Intent}
\label{inference-section}

   	To assemble communicative intent, \spud\ deploys
   lexico-grammatical entries like \sref{combo-example} one by one, as
   depicted in Figure~\ref{nlg-steps-fig}.  As
   Section~\ref{problem-sec} suggested, these steps involve both
   grammatical inference to link linguistic structures together and
   contextual inference to link linguistic meanings to domain-specific
   representations.  We now describe the specific form of these
   inferential processes in \spud.

\subsection{Grammatical Inference}
\label{grammar-inference-subsec}

	In \spud's grammar, the trees of entries like
   \sref{combo-example} describe a set of elementary structures for a
   feature-based lexicalized tree-adjoining grammar, or LTAG
   \cite{joshi:tag,vijay:diss,schabes90}.  In all TAG formalisms,
   entries can be combined into larger trees by two operations, called
   \term{substitution} and \term{adjoining}.  Elementary trees without
   foot nodes are called \term{initial} trees and can only substitute;
   trees with foot nodes are called \term{auxiliary} trees, and can
   only adjoin.  The trees that these operations yield are called
   \term{derived} trees; we regard the computation of derived trees as
   an inference about a complex structure that follows from a
   declarative specification of elementary structures.  In a grammar
   with features, derived trees are completed by unifying the top and
   bottom features on each node.

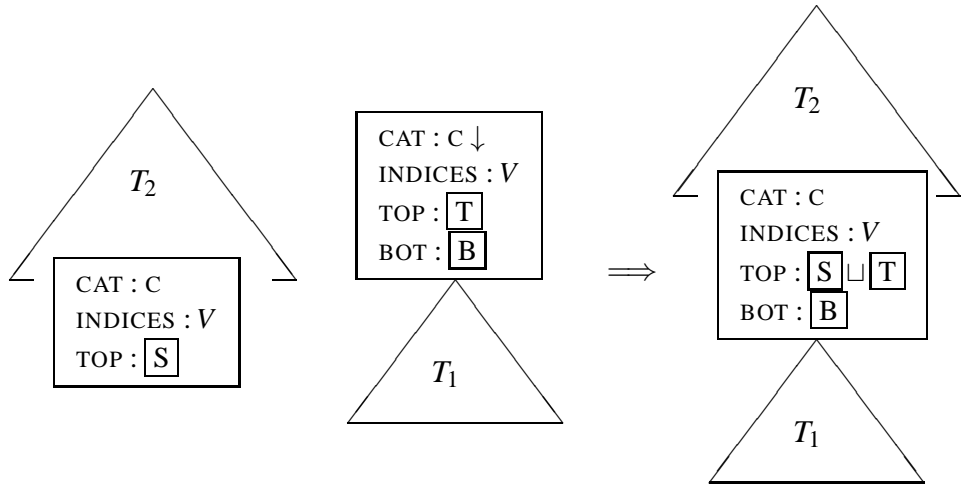
\begin{figure}
\begin{center}
\setlength{\unitlength}{0.012500in}%
\raisebox{-0.75in}{
\begin{picture}(120,120)(20,685)
\thinlines
\put( 80,825){\line(-3,-4){ 60}}
\put( 80,825){\line( 3,-4){ 60}}
\put( 20,745){\line( 1, 0){ 10}}
\put(140,745){\line( -1, 0){ 10}}
\put( 70,780){\makebox(0,0)[lb]{$T_2$}}
\put( 35,700){\makebox(0,0)[lb]{
	\fbox{\small$\begin{array}{l}
		\term{cat}:\term c \\
		\term{indices}: V \\
		\term{top}:  \fbox{S}
	\end{array}$}}
}
\end{picture}}
\raisebox{-0.75in}{
\begin{picture}(120,120)(20,765)
\thinlines
\put( 80,825){\line(-3,-4){ 45}}
\put( 80,825){\line( 3,-4){ 45}}
\put( 35,765){\line( 1, 0){ 90}}
\put( 70,780){\makebox(0,0)[lb]{$T_1$}}
\put( 35,825){\makebox(0,0)[lb]{
	\fbox{\small$\begin{array}{l}
		\term{cat}:\term c\downarrow \\
		\term{indices}: V \\
		\term{top}:  \fbox{T} \\
		\term{bot}:  \fbox{B} 
	\end{array}$}
}}
\end{picture}}
$\Longrightarrow$
\raisebox{-1.125in}{
\begin{picture}(120,180)(20,620)
\thinlines
\put( 80,825){\line(-3,-4){ 60}}
\put( 80,825){\line( 3,-4){ 60}}
\put( 20,745){\line( 1, 0){ 10}}
\put(140,745){\line( -1, 0){ 10}}
\put( 70,780){\makebox(0,0)[lb]{$T_2$}}
\put( 35,685){\makebox(0,0)[lb]{
	\fbox{\small$\begin{array}{l}
		\term{cat}:\term c \\
		\term{indices}: V \\
		\term{top}:  \fbox{S} \sqcup \fbox{T} \\
		\term{bot}:  \fbox{B} 
	\end{array}$}
}}
\put( 80,685){\line(-3,-4){ 45}}
\put( 80,685){\line( 3,-4){ 45}}
\put( 70,640){\makebox(0,0)[lb]{$T_1$}}
\put( 35,625){\line( 1, 0){ 90}}
\end{picture}}
\end{center}
\caption{Substitution of $T_1$ into $T_2$.}
\label{subst-fig}
\end{figure}

	In substitution, the root of an initial tree is identified
   with a leaf of another elementary or derived structure, called the
   \textsc{substitution} site.  The top feature structure of the
   substitution site is unified with the top feature structure of the
   root of the initial tree.  Figure~\ref{subst-fig} schematizes this
   operation.

\begin{figure}
\begin{center}
\setlength{\unitlength}{0.012500in}%
\raisebox{-1.125in}{
\begin{picture}(120,180)(20,630)
\thinlines
\put( 80,825){\line(-3,-4){ 60}}
\put( 80,825){\line( 3,-4){ 60}}
\put( 20,745){\line( 1, 0){ 10}}
\put(140,745){\line( -1, 0){ 10}}
\put( 70,780){\makebox(0,0)[lb]{$T_{2a}$}}
\put( 35,685){\makebox(0,0)[lb]{
	\fbox{\small$\begin{array}{l}
		\term{cat}:\term c \\
		\term{indices}: V \\
		\term{top}:  \fbox{T$_2$} \\
		\term{bot}:  \fbox{B$_2$} 
	\end{array}$}
}}
\put( 80,685){\line(-3,-4){ 45}}
\put( 80,685){\line( 3,-4){ 45}}
\put( 70,640){\makebox(0,0)[lb]{$T_{2b}$}}
\put( 35,625){\line( 1, 0){ 90}}
\end{picture}}
\raisebox{-1.125in}{
\begin{picture}(120,180)(20,685)
\thinlines
\put( 80,825){\line(-3,-4){ 60}}
\put( 80,825){\line( 3,-4){ 60}}
\put( 20,745){\line( 1, 0){ 10}}
\put(140,745){\line( -1, 0){ 10}}
\put( 70,780){\makebox(0,0)[lb]{$T_1$}}
\put( 35,825){\makebox(0,0)[lb]{
	\fbox{\small$\begin{array}{l}
		\term{cat}:\term c \\
		\term{indices}: V \\
		\term{top}:  \fbox{T$_{1r}$} \\
		\term{bot}:  \fbox{B$_{1r}$}
	\end{array}$}
}}
\put( 35,705){\makebox(0,0)[lb]{
	\fbox{\small$\begin{array}{l}
		\term{cat}:\term c_{*} \\
		\term{indices}: V \\
		\term{top}:  \fbox{T$_{1f}$}  
	\end{array}$}
}}
\end{picture}}
$\Longrightarrow$
\raisebox{-2.125in}{
\begin{picture}(120,340)(20,480)
\thinlines
\put( 80,825){\line(-3,-4){ 60}}
\put( 80,825){\line( 3,-4){ 60}}
\put( 20,745){\line( 1, 0){ 10}}
\put(140,745){\line( -1, 0){ 10}}
\put( 70,780){\makebox(0,0)[lb]{$T_{2a}$}}
\put( 30,685){\makebox(0,0)[lb]{
	\fbox{\small$\begin{array}{@{}l@{}}
		\term{cat}:\term c \\
		\term{indices}: V \\
		\term{top}:  \fbox{T$_2$} \sqcup \fbox{T$_{1r}$} \\
		\term{bot}:  \fbox{B$_{1r}$} 
	\end{array}$}
}}
\put( 80,685){\line(-3,-4){ 60}}
\put( 80,685){\line( 3,-4){ 60}}
\put( 20,605){\line( 1, 0){ 10}}
\put(140,605){\line( -1, 0){ 10}}
\put( 70,640){\makebox(0,0)[lb]{$T_{1}$}}
\put( 30,535){\makebox(0,0)[lb]{
	\fbox{\small$\begin{array}{@{}l@{}}
		\term{cat}:\term c \\
		\term{indices}: V \\
		\begin{array}{@{}c@{}}\term{top}\\\term{bot}\end{array}: \fbox{B$_2$} \sqcup \fbox{T$_{1f}$} \\
   		\term{(no adjoining)}
	\end{array}$}
}}
\put( 80,535){\line(-3,-4){ 45}}
\put( 80,535){\line( 3,-4){ 45}}
\put( 70,490){\makebox(0,0)[lb]{$T_{2b}$}}
\put( 35,475){\line( 1, 0){ 90}}
\end{picture}}
\end{center}
\caption{Adjunction of $T_1$ into $T_2$}
\label{adjoin-fig}
\end{figure}
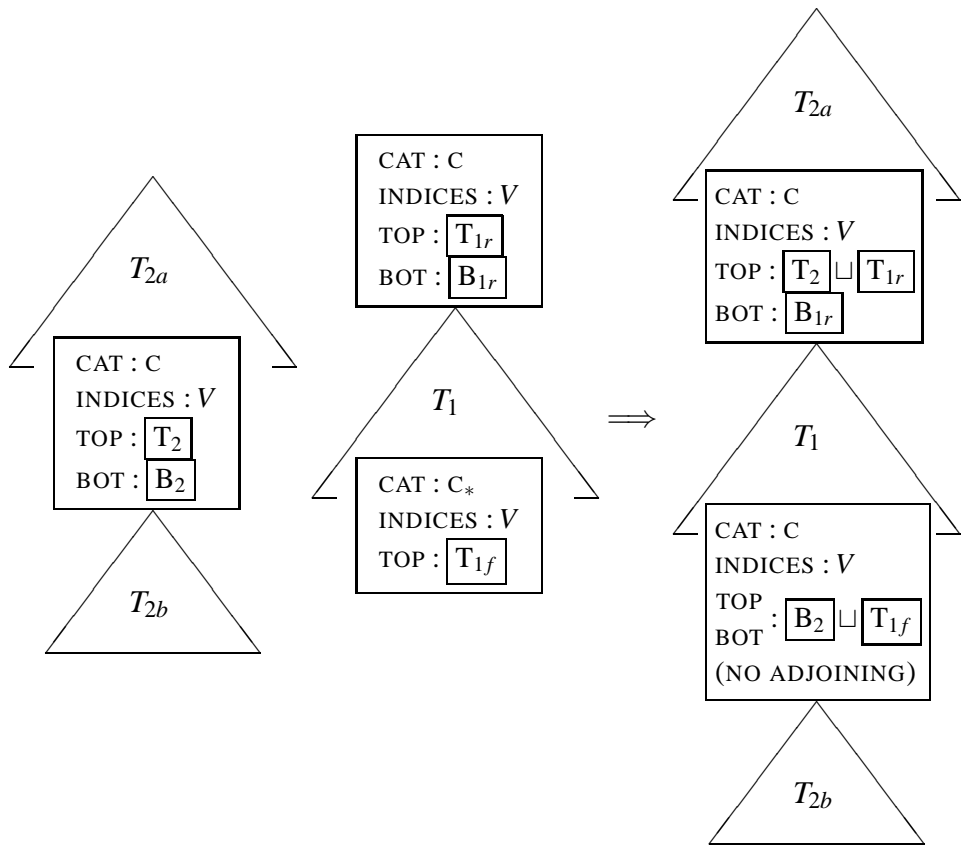

	Adjoining is a more complicated splicing operation, where an
   elementary structure \term{displaces} some subtree of another
   elementary or derived structure.  The node in this structure where
   the replacement applies is called the \term{adjunction site}; the
   excised subtree is then substituted back into the first tree at the
   distinguished \term{foot} node.  As part of an adjoining operation,
   the top feature structure of the adjunction site is unified with
   the top feature structure of the root node of the auxiliary tree;
   the bottom feature structure of the adjunction site is unified with
   the feature structure of the foot node.  After an adjoining
   operation, no further adjoining is possible at the foot node.
   This is schematized in Figure~\ref{adjoin-fig}.

	In substitution, the substitution site and the root node of
   the substituted tree must have the same category; likewise, in
   adjoining, the root node, the foot node and the adjunction site
   must all have the same category.  Moreover, as our trees
   incorporate indices labeling the nodes, there is the further
   requirement that any nodes that are identified through substitution
   or adjoining must carry identical indices.

   	The identification of indices in trees determines the
   interface between syntax and semantics in \spud.  \spud\ adopts an
   ontologically promiscuous semantics \cite{hobbs:promiscuity}, in
   the sense that each entry used in the derivation of an utterance
   contributes a constraint to its overall semantics.  Syntax
   determines when the constraints contributed by different
   grammatical entries describe the same variables or discourse
   anaphors.  For example, take the phrase \emph{slide the sleeve
   quickly}.  Its lexical elements contribute constraints describing
   an event $e$ in which agent $x$ \ling{slides} object $y$ along path
   $p$; describing an individual $z$ that is a \ling{sleeve}; and
   describing an event $e'$ that is \ling{quick}.  The
   syntax--semantics interface provides the guarantee that $y=z$ and
   $e=e'$ (i.e., that the sleeve is what is slid and that the sliding
   is what is quick).  It does so by requiring that the index $y$ of
   the object \term{np} substitution site of \ling{slide} unify with
   the index $z$ of the root \term{np} for \ling{sleeve}, and by
   requiring that the index $e$ of the \term{vp} adjunction site for
   \ling{slide} unify with the index $e'$ of the \term{vp} foot node
   for \ling{quickly}.  (See
   \cite{hobbs:promiscuity,hobbs:abduction/journal} for more details
   on ontologically promiscuous semantics.)

	Note that this strategy contrasts with other approaches to
   LTAG semantics, such as \cite{candito/kahane:tag+98a}, which
   describe meanings primarily in terms of function-argument
   relations.  (It is also possible to combine both function-argument
   and constraint semantics, as in
   \cite{joshi:1999,kallmeyer/joshi:ams99}.)  Like Hobbs, we use
   semantic representations as a springboard to explore the
   relationships between sentence meaning, background knowledge and
   inference---relationships which are easiest to state in terms of
   constraints.  In addition, the use of constraints harmonizes with
   our perspective that the essential microplanning task is to
   construct extended descriptions of individuals
   \cite{genw-paper,webber:acl99}.

	Let us illustrate the operations of grammatical inference by
   describing how the structure for \emph{fuel-line} can combine with
   the structure for \emph{sealing ring} by adjoining.
   \emph{Fuel-line} will be associated with a combined
   lexico-syntactic realization as in \sref{fuel-combo}.
\begin{examples}{fuel-combo}
\exitem{flt}
	\term{tree}: \mbox{
	\leaf{fuel-line}
	\branch{1}{$\term{n}'(F)$}
	\branch{1}{$\term{np}(F)$}
	\leaf{$\term{n}'(R)\footnodemark$}
	\branch{2}{$\term{n}'(R)$}
	\tree}
\item
	\term{target}: $\term{n}'(R)$ [modifier]
\item
	\term{assertion}: ---
\item
	\term{presupposition}: $\m{fl}(F) \wedge
   \m{nn}(R,F,X)$
\item
	\term{pragmatics}: $\m{def}(F)$
\end{examples}
	We can adjoin \msref{fuel-combo}{flt} into
   \msref{combo-example}{cet} using the $\term{n}'(R)$ node as the
   adjunction site, to obtain the structure in \sref{combo-combo}.
\sentence{combo-combo}{
\mbox{
	\leaf{fuel-line}
	\branch{1}{$\term{n}'(F)$}
	\branch{1}{$\term{np}(F)$}
	\leaf{sealing-ring}
	\branch{1}{$\term{n}'(R)$}
	\branch{2}{$\term{n}'(R)$}
	\branch{1}{$\term{np}(R)$}
	\tree}
}

	When we put together entries by TAG operations, we can
   represent the meaning of the combined structure as the
   component-wise conjunction of the meanings of its constituents.  In
   the case of \sref{combo-example} and \sref{fuel-combo} this would
   yield:
\begin{examples}{ccc}
\item
	\term{assertion}: ---
\exitem{cccp}
	\term{presupposition}: $\m{fl}(F) \wedge
   \m{nn}(R,F,X) \wedge \m{sr}(R)$
\item
	\term{pragmatics}: $\m{def}(F) \wedge \m{def}(R) \wedge
   \m{zero-genre}$
\end{examples}
   (As explained in the next section, we can also directly describe
   the joint interpretation of combined elements, in terms of intended
   links to the conversational record and intended updates to it.)

   	In addition to explicitly setting out the structure of a TAG
   derived tree as in \sref{combo-combo}, we can also describe a
   derived tree implicitly in terms of operations of substitution and
   adjoining which generate the derived tree.  Such a description is
   called a TAG \term{derivation tree} (see \cite{vijay:diss} for a
   formal definition and discussion of TAG derivation trees).  Each
   node in a derivation tree represents an elementary tree that
   contributes to the derived tree.  Each edge in a derivation tree
   specifies a mode of combination: the child node is combined to the
   parent node by a specified TAG operation at a specified node in the
   structure.  For example, \sref{derivation-tree} shows the
   derivation tree corresponding to \sref{combo-combo}.
\sentence{derivation-tree}{
\begin{singlespace}
\hspace*{1in}
\makebox[0in]{
\leaf{\begin{tabular}{c}
      	Tree \msref{fuel-combo}{flt}:\emph{fuel-line}  \\
	by adjoining at node $\term{n}'$	
      \end{tabular}
      }
\branch{1}{Tree \msref{combo-example}{cet}:\emph{sealing-ring}}
\tree
}
\end{singlespace}}
	Derivation trees indicate the decisions required to produce a
   sentence and outline the search space for the generation system
   more perspicuously than do derived trees.  This makes derivation
   trees particularly attractive structures for describing an NLG
   system; for example, we can represent a TAG derivation tree for
   utterance \sref{instruction-eg} with a structure isomorphic to the
   the dependency tree \sref{instruction-deps}.

\subsection{Contextual Inference}
\label{context-inference-subsec}

   	\spud\ assembles structures and meanings such as
   \sref{combo-combo}--\sref{derivation-tree} to exploit connections
   between linguistic meanings and domain-specific representations.
   For example, the presupposition \msref{ccc}{cccp} connects the
   meaning of the constituent \emph{fuel-line coupling nut} with
   shared referents $f4$ and $r11$ in the aircraft domain; \spud\
   might use the connection to identify these referents to the user.

	\spud's module for contextual inference determines the
   availability of such connections.  The main resource for this
   module is a domain-specific knowledge base, specified as logical
   formulas.  This knowledge base describes both the private
   information available to the system and the shared information that
   characterizes the state of the conversation.  Tasks for contextual
   inference consult this knowledge base: \spud\ first translates a
   potential connection between meaning and context into a
   theorem-proving query, and then confirms or rejects the connection
   by using a logic programming search strategy to evaluate the query
   against the contextual knowledge base.  When the inferential
   connection is established, \spud\ can record the inference as a
   constituent of its communicative intent.%
\dspftnt{
	Note that this strategy is strongly monotonic: \spud's
   inference tasks are deductive and the links \spud\ adds to
   communicative intent cannot be threatened by the addition of
   further information.  Previous researchers have pointed out that
   much inference in interpretation is nonmonotonic
   \cite{lascarides:relations,hobbs:abduction/journal}.  We take it as
   future work to extend \spud's contextual inference,
   communicative-intent representations, and search strategy to this
   more general case.
}

	We now describe \spud's knowledge base, \spud's queries, and
   the inference procedure that evaluates them in more detail.
   \spud's knowledge base is specified in first-order modal logic.
   First-order modal logic extends first-order classical logic by the
   addition of \term{modal operators}; these operators can be used to
   relativize the truth of a sentence to a particular time, context or
   information-state.  We will use modal operators to refer to a
   particular body of knowledge.  Thus, if $p$ is a formula and $\Box$
   is a modal operator, then $\Box p$ is a formula; $\Box p$ means
   that $p$ follows from the body of knowledge associated with $\Box$.

	For specifications in NLG, we use four such operators:
   \speaker\ represents the private knowledge of the generation
   system; \user\ represents the private knowledge of the other party
   to the conversation, the user; \shared\ represents the content of
   the conversational record; and finally \meanings\ (for
   \term{meaning postulates}) represents a body of semantic
   information that follows just from the meanings of words.  We
   regard the four sources of information as subject to the eleven
   axiom schemes presented in \sref{conversation.ops}:
\begin{examples}{conversation.ops}
\exitem{c.accurate}
	$\speaker p \imp p. \;\;\; \user p \imp p. 
	\;\;\; \shared p \imp p. \;\;\; \meanings p \imp p.$

\exitem{c.intro}
	$\speaker p \imp \speaker\speaker p.
   \;\;\; \user p \imp \user\user p.
   \;\;\; \shared p \imp \shared\shared p.
   \;\;\; \meanings p \imp \meanings\meanings p.$

\exitem{c.shared}
	$\meanings p \imp \shared p. \;\;\;
	\shared p \imp \speaker p. \;\;\;
	\shared p \imp \user p$
\end{examples}
	The system's information, the user's information, the
   conversational record and the background semantic information are
   all accurate, according to the idealization of
   \msref{conversation.ops}{c.accurate}.  The effect of
   \msref{conversation.ops}{c.intro} is that hypothetical reasoning
   with respect to a body of knowledge retains access to all the
   information in it.  Finally, \msref{conversation.ops}{c.shared}
   ensures that semantic knowledge and the contents of the
   conversational record are in fact shared.  \cite{stone:phdthesis}
   explores the relationship between this idealization of conversation
   implicit in these inference schemes and proposals for reasoning
   about dialogue context by Clark and Marshall
   (1981)\nocite{clark/marshall:mutual} and others.  For current
   purposes, note that inferences using the schemes in
   \sref{conversation.ops} are not intended to characterize the
   explicit beliefs of participants in conversation veridically.
   Instead, the inferences contribute to a data structure,
   communicative intent, whose principal role is to support
   conversational processes such as plan recognition, coordination and
   negotiation.

	In this paper, we consider specifications of domain knowledge
   and queries of domain knowledge that can be restricted to the
   logical fragment involving \ling{definitions} of category $D$ and
   \ling{queries} of category $Q$ defined by the following,
   mutually-recursive rules:
\sentence{logic.fragment}{
$\begin{array}{l}
D ::= Q \;\;\; | \;\;\; Q \imp D \;\;\; | \;\;\; \forall x D \\
Q ::= \shared D \;\;\; | \;\;\; \speaker D \;\;\; | \;\;\; \user D
   \;\;\; | \;\;\; Q \wedge Q    \;\;\; | \;\;\; A \\
\end{array}$
}
	$A$ schematizes over any atomic formula; $x$ schematizes over
   any bound variable.  We use the notation $?K \proves q$ to denote
   the task of proving a $Q$-formula $q$ as a query from a knowledge
   base $K$ consisting of a set of $D$-formulas; we indicate by
   writing $K \proves q$ that this task results in the construction of
   a proof, and thus that the query succeeds.

	This fragment allows for the kind of clauses and facts that
   form the core of a logic programming language like Prolog.  In
   addition, these clauses and facts may make free use of modal
   operators; they may have nested implications and nested quantifiers
   in the body of rules, provided they are immediately embedded under
   modal operators.  There have been a number of proposals for logic
   programming languages along these lines, such as
   \cite{cerro:molog,debart:modal/lp,baldoni:jlc98}.  Our
   implementation follows \cite{stone:modal/lp}, which also allows for
   more general specifications including disjunction and existential
   quantifiers.  For a discussion of NLG inference using the more
   general modal specifications, see \cite{jlac00}.

   	\spud's knowledge base is a set of $D$ formulas.  These
   formulas provide all the information about the world and the
   conversation that \spud\ can draw on to construct and to evaluate
   possible communicative intent.  Concretely, for \spud\ to construct
   communicative intent, the knowledge base must support any
   assertions, presuppositions and pragmatics that \spud\ decides to
   appeal to in its utterance.  Thus, the knowledge base should
   explicitly set up as system knowledge any information that \spud\
   may assert; if some intended update relates by inference to an
   assertion, the knowledge base must provide, as part of the
   conversational record, rules sufficient to infer the update from
   the assertion.  Moreover, the knowledge base must provide, as part
   of the conversational record, formulas which entail the
   presuppositions and pragmatic conditions that \spud\ may impose.
   Meanwhile, for \spud\ to assess whether the hearer will interpret
   an utterance correctly, the knowledge base must describe the
   context richly enough to characterize not just the intended
   communicative intent for a provisional utterance, but also any
   potential alternatives to it.

   	For the communicative intent of
   Figure~\ref{instruct-intent-fig}, then, the knowledge base must
   include the specific private facts that underlie the assertion in
   the instruction, as in \sref{instruct-kb-private}:
\sentence{instruct-kb-private}{
$\begin{array}[t]{l}
\speaker \m{move}(a1,h0,n11,p(l(\m{on},j2),l(\m{on},e2))). \\
\speaker \m{next}(a1). \\
\speaker \m{purpose}(a1,a2). \\
\speaker \m{uncover}(a2,h0,r11).
\end{array}$
}
	(Recall that, in words, \sref{instruct-kb-private} describes
   the \ling{next} action, a \ling{move} event which takes the nut
   along a specified path and whose \ling{purpose} is to
   \ling{uncover} the sealing-ring.)  For this communicative intent,
   no further specification is required for the links between
   assertions and updates.  Updates are expressed in the same terms as
   meanings here, so the connection will follow as a matter of logic.

	At the same time, the knowledge base must include the specific
   facts and rules that permit the presuppositions and pragmatics of
   the instruction to be recognized as part of the conversational
   record.  \msref{instruct-kb-public}{ikp-instances} spells out the
   instances that are simply listed in the conversational record;
   \msref{instruct-kb-public}{ikp-inferences} describes the rules and
   premises that allow the noun-noun compound and the spatial
   presuppositions to be interpreted by inference as in
   \sref{loc-inference-tree} and \sref{inference-tree}.
\begin{examples}{instruct-kb-public}
\exitem{ikp-instances}
$\begin{array}[t]{ll}
\shared \m{partic}(s0,h0).  & \shared \m{obl}(s0,h0). \\
\shared \m{surf}(p(l(\m{on},j2),l(\m{on},e2))). & \shared \m{zero-genre}. \\
\shared \m{cn}(n11). & \shared \m{def}(n11). \\
\shared \m{el}(e2). & \shared \m{def}(e2). \\
\shared \m{sr}(r11). & \shared \m{def}(r11). \\
\shared \m{fl}(f4). & \shared \m{def}(f4). \\
\end{array}$

\exitem{ikp-inferences}
$\begin{array}[t]{ll}
\shared \m{for}(r11,f4). & 
	\shared \forall ab (\m{for}(a,b) \imp \m{nn}(a,b,\m{for})). \\
\shared \m{loc}(l(\m{on},j2),n11) & \\
\shared \forall loe (\m{loc}(l,o) \imp \m{start-at}(p(l,e),o)). &
	\shared \forall se (\m{end-on}(p(s,l(\m{on},e)),e))
\end{array}$
\end{examples}
	(Again, with our conventions,
   \msref{instruct-kb-public}{ikp-instances} spells out such facts as
   that $s0$ and $h0$ are the speaker and hearer \ling{participating}
   in the current conversation, and that $s0$ is empowered to impose
   \ling{obligations} on $h0$.  Likewise,
   \msref{instruct-kb-public}{ikp-inferences} indicates that the ring
   is \ling{for} the fuel-line, and that \ling{for} is the right kind
   of relationship to interpret a noun-noun compound; that a path that
   starts where an object is \ling{located} \ling{starts at} the
   object; and that any path whose endpoint is \ling{on} an object
   \ling{ends on} the object.)

	Of course, the knowledge base cannot be limited to just the
   facts that figure in this particular communicative intent.  \spud\
   is designed to be supplied with a number of other facts, both
   private and shared, about the discourse referents evoked by the
   instruction.  This way \spud\ has substantive lexical choices that
   arise in achieving specified updates to the state of the
   conversation.  \spud\ also expects to be supplied with additional
   facts describing other discourse referents from the context.  This
   way \spud\ can consult the specification of the context to arrive
   at meaningful assessments of ambiguities in interpretation.  For
   instance, the knowledge base must describe any other fuel lines and
   other sealing rings to settle whether there are is any referential
   ambiguity in the phrase \emph{fuel-line sealing ring}.  For
   exposition, we note only the bare-bones alternatives required for
   \spud\ to generate \sref{instruction-eg} given the task of
   describing the upcoming uncovering motion:
\begin{examples}{instruct-kb-alts}
\item
	$\shared \m{sr}(a_{r11})$
\item
	$\shared \m{surf}(p(l(\m{on},j2),l(a_{on},a_{e2})))$
\end{examples}
   	There must be another sealing ring $a_{r11}$ for \spud\ to
   explicitly indicate $r11$ as the \ling{fuel-line} sealing ring; and
   there must be another path to slide $n11$ along, for \spud\ to
   explicitly describe the intended path as \ling{onto elbow}.%
\dspftnt{
	\spud's greedy search also requires that this alternative path
   not end \ling{on} anything, but instead end perhaps \ling{around}
   or \ling{over} its endpoint.  The explanation for this depends on
   the results of Section~\ref{hearer-inference-subsec} and
   Section~\ref{algorithm-section}, but briefly, \spud\ will adjoin
   the modifier \ling{onto} only if \ling{onto} by itself rules out
   some path referents (and thus by itself helps the hearer to
   interpret the instruction).
}

   	Now we consider the steps involved in linking grammatical
   structures such as \sref{combo-example} or
   \sref{combo-combo}--\sref{ccc} to domain-specific representations.
   As described in Section~\ref{grammar-inference-subsec}, the grammar
   delivers an assertion $A$, a presupposition $P$ and pragmatics $Q$
   for each derivation tree.  Links to domain-specific representations
   come as \spud\ constructs a communicative intent for this
   derivation tree by reasoning from the context.

	In doing this, \spud\ must link up $P$ and $Q$ in a specific
   way with particular referents and propositions from the
   conversational record.  We introduce an assignment $\sigma$ taking
   variables to terms to indicate the correspondence between anaphors
   and intended referents.  (We write out assignments as lists of the
   form $\{ \ldots V_i \leftarrow t_i \ldots \}$ where each variable
   $V_i$ is assigned term $t_i$ as its value; for any structure $E$
   containing variables, and any assignment $\sigma$ of values to
   those variables, we use $E\sigma$ to indicate the result of
   replacing the occurrences of variables in $E$ by the terms assigned
   by $\sigma$.)  In addition, \spud\ must link up the assertion $A$
   with particular open questions in the discourse in virtue of the
   information it presents about particular individuals.  We
   schematize any such update as a condition $U$.

	These links between $A$, $P$ and $Q$ and the context
   constitute the presumptions that \spud\ makes with its utterance;
   \spud\ explicitly records them in its representation of
   communicative intent.  Since these links are inferences,
   constructing them is a matter of proof.  In \spud, these proof
   tasks are carried out using logic programming inference and a modal
   specification of context.
\begin{itemize}
\item
	Checking that the intended instance of the assertion $A$ is
   true corresponds to the proof task:
\[
	?K \proves \speaker A\sigma
\]
	That is, does some instance of $A\sigma$ follow from the
   information available to the speaker?  As usual in logic
   programming, if $\sigma$ leaves open the values of some variables,
   then the proof actually describes a more specific instance
   $\speaker A\sigma'$ where the substitution $\sigma'$ possibly
   supplies values for these additional variables.

\item
	Checking that the intended instance of the assertion $A$ leads
   to the update $U$ corresponds to the proof task:
\[
	?K \proves \shared (\shared A\sigma \imp \shared U)
\]
	That is, considering only the content of the conversational
   record, can we show that when $A\sigma$ is added to the
   conversational record, $U$ also becomes part of the conversational
   record?  Note that $\shared (\shared p \imp \shared p)$ is a valid
   formula of modal logic, for any $p$.  Such a query always succeeds,
   regardless of the specification $K$.

\item
	Checking that a presupposition $P$ is met for an intended
   instance corresponds to the proof task:
\[
	?K \proves \shared P\sigma
\]
	That is, does $P\sigma$ follow from the conversational record?
   More generally, determining the potential instances under which the
   presupposition $P$ is met corresponds to the proof task:
\[
	? K \proves \shared P
\]
	Each proof shows how the context supports a specific
   resolution $\sigma'$ of underspecified elements in the meaning of
   the utterance, by deriving an instance $P\sigma'$.  Such instances
   need not be just the one that the system intends.  Checking that
   pragmatic conditions $Q$ are met for an intended instance also
   corresponds to a query $?K \proves \shared Q\sigma$.
\end{itemize}
	Our logic programming inference framework allows queries and
   knowledge bases to be understood operationally as instructions for
   search, much as in Prolog; see \cite{mnps:uniform}.  For example, a
   query $\Box p$ is an instruction to move to a new possible world
   and consider the query $p$ there; a query $\forall x \, p$ is an
   instruction to consider a new arbitrary individual in place of $x$
   in proving $p$.  A query $p \imp q$ is an instruction to assume
   $p$ temporarily while considering the query $q$; a query $p \wedge
   q$ is an instruction to set up two subproblems for search: a query
   of $p$ and a query of $q$.  Logical connectives in knowledge-base
   clauses, meanwhile, are interpreted as describing matches for
   predicates, first-order terms, and possible worlds in atomic
   queries, and as setting up subproblems with additional queries of
   their own.  Overall then, each theorem-proving problem initiates a
   recursive process where the inference engine breaks down complex
   queries into a collection of search problems for atomic queries,
   backward-chains against applicable clauses in the knowledge base to
   search for matches for atomic queries, and takes on any further
   queries that result from the matches.  

	As in Prolog, the course and complexity of the proof process
   can be determined from the form of the queries and the
   knowledge-base.  Thus, when necessary, performance can be improved
   by astute changes in the representation and formalization of domain
   relationships.  Proof search is no issue with
   \sref{instruction-eg}, for example; inspection of the clauses in
   \sref{instruct-kb-private}, \sref{instruct-kb-public} and
   \sref{instruct-kb-alts} will confirm that logic programming search
   explores the full search space for generation queries for this
   instruction without having to reason recursively through
   implications.

\subsection{Concrete Representations of Communicative Intent}
\label{data-structure-subsec}

\begin{figure}
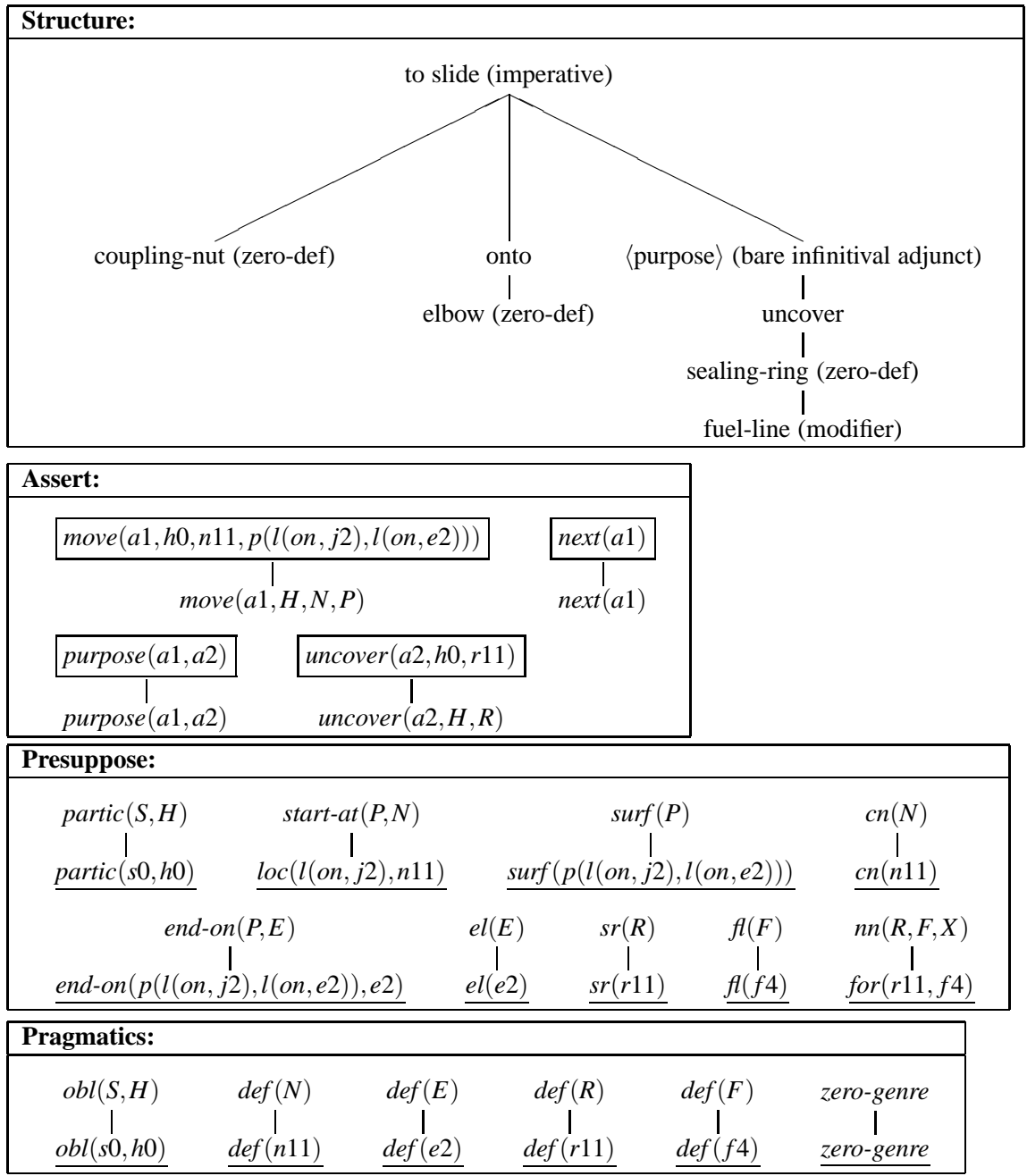

\small
\begin{center}
\begin{tabular}{l}
\lxstructuret{
\mbox{
\leaf{coupling-nut (zero-def)}
\leaf{elbow (zero-def)}
\branch{1}{onto}
\leaf{fuel-line (modifier)}
\branch{1}{sealing-ring (zero-def)}
\branch{1}{uncover}
\branch{1}{$\langle$purpose$\rangle$ (bare infinitival adjunct)}
\branch{3}{to slide (imperative)}
\tree}
} \\[19ex]
\alinkst{
\mbox{
	\leaf{$\m{move}(a1,H,N,P)$}
	\branch{1}{\goalinst{\m{move}(a1,h0,n11,p(l(\m{on},j2),l(\m{on},e2)))}}
	\tree
}
\mbox{
	\leaf{$\m{next}(a1)$}
	\branch{1}{\goalinst{\m{next}(a1)}}
	\tree
} \\[2em]
\mbox{
	\leaf{$\m{purpose}(a1,a2)$}
	\branch{1}{\goalinst{\m{purpose}(a1,a2)}}
	\tree
}
\mbox{
	\leaf{$\m{uncover}(a2,H,R)$}
	\branch{1}{\goalinst{\m{uncover}(a2,h0,r11)}}
	\tree
}} \\[11ex]
\pslinkst{
\mbox{
	\leaf{\sharedinst{\m{partic}(s0,h0)}}
	\branch{1}{$\m{partic}(S,H)$}
	\tree
}
\mbox{
	\leaf{\sharedinst{\m{loc}(l(\m{on},j2),n11)}}
	\branch{1}{$\m{start-at}(P,N)$}
	\tree
}
\mbox{
	\leaf{\sharedinst{\m{surf}(p(l(\m{on},j2),l(\m{on},e2)))}}
	\branch{1}{$\m{surf}(P)$}
	\tree
}
\mbox{
	\leaf{\sharedinst{\m{cn}(n11)}}
	\branch{1}{$\m{cn}(N)$}
	\tree
}
\\[2em]
\mbox{
	\leaf{\sharedinst{\m{end-on}(p(l(\m{on},j2),l(\m{on},e2)),e2)}}
	\branch{1}{$\m{end-on}(P,E)$}
	\tree
}
\mbox{
	\leaf{\sharedinst{\m{el}(e2)}}
	\branch{1}{$\m{el}(E)$}
	\tree
}
\mbox{
	\leaf{\sharedinst{\m{sr}(r11)}}
	\branch{1}{$\m{sr}(R)$}
	\tree
}
\mbox{
	\leaf{\sharedinst{\m{fl}(f4)}}
	\branch{1}{$\m{fl}(F)$}
	\tree
}
\mbox{
	\leaf{\underline{$\m{for}(r11,f4)$}}
	\branch{1}{$\m{nn}(R,F,X)$}
	\tree
}} \\[11ex]
\pglinkst{
\mbox{
	\leaf{\sharedinst{\m{obl}(s0,h0)}}
	\branch{1}{$\m{obl}(S,H)$}
	\tree
}
\mbox{
	\leaf{\sharedinst{\m{def}(n11)}}
	\branch{1}{$\m{def}(N)$}
	\tree
}
\mbox{
	\leaf{\sharedinst{\m{def}(e2)}}
	\branch{1}{$\m{def}(E)$}
	\tree
}
\mbox{
	\leaf{\sharedinst{\m{def}(r11)}}
	\branch{1}{$\m{def}(R)$}
	\tree
}
\mbox{
	\leaf{\sharedinst{\m{def}(f4)}}
	\branch{1}{$\m{def}(F)$}
	\tree
}
\mbox{
	\leaf{\sharedinst{\m{zero-genre}}}
	\branch{1}{$\m{zero-genre}$}
	\tree
}}
\end{tabular}
\end{center}
\caption[Communicative intent for \sref{instruction-eg}.]{%
Communicative intent for \sref{instruction-eg}.  The grammar specifies
   meanings as follows: For \emph{slide}, assertions \m{move} and
   \m{next}; for the bare infinitival adjunct, \m{purpose}; for
   \emph{uncover}, \m{uncover}.  For \emph{slide}, presuppositions
   \m{partic}, \m{start-at} and \m{surf}; for \emph{coupling-nut},
   \m{cn}; for \emph{onto}, \m{end-on}; for \emph{elbow}, \m{el}; for
   \emph{sealing-ring}, \m{cn}; for \emph{fuel-line}, \m{fl} and
   \m{nn}.  For \emph{slide}, pragmatics \m{obl}; for other nouns,
   pragmatics \m{def} and \m{zero-genre}.  The speaker's presumptions
   map out intended connections to discourse referents as follows: the
   speaker $S$, $s0$; the hearer $H$, $h0$; the nut $N$, $n11$; the
   path $P$, $p(l(\m{on},j2),l(\m{on},e2))$; the elbow $E$, $e2$; the
   ring $R$, $r11$; the fuel-line $F$, $f4$; the relation $X$,
   $\m{for}$.  The fuel-line joint is $j2$.
}
\label{instruct-intent-fig-2}
\end{figure}

	We can now return to the communicative intent of
   Figure~\ref{instruct-intent-fig} to describe the concrete
   representations by which \spud\ implements it.  For reference, we
   repeat Figure~\ref{instruct-intent-fig} as
   Figure~\ref{instruct-intent-fig-2} here.

   	The grammar delivers a TAG derivation whose structure is
   isomorphic to the tree-structure of
   Figure~\ref{instruct-intent-fig-2}.  That derivation is associated
   with a meaning that we represent as the triple of conditions of
   \msref{instruct-meaning}{instruct-assert}--\msref{instruct-meaning}{instruct-prags};
   \msref{instruct-meaning}{instruct-instance} spells out the
   instantiation $\sigma$ under which this meaning is to be linked to
   the communicative context:
\begin{examples}{instruct-meaning}
\exitem{instruct-assert}
	Assertion: 
	$\m{move}(a1,H,N,P) \wedge 
	 \m{next}(a1) \wedge
	\m{purpose}(a1,a2) \wedge
	\m{uncover}(a2,H,R)$
\exitem{instruct-presup}
	Presupposition: 
	$\m{partic}(S,H) \wedge
	\m{start-at}(P,N) \wedge
	\m{surf}(P) \wedge
	\m{cn}(N) \wedge
	\m{end-on}(P,E) \wedge
	\m{el}(E) \wedge
	\m{sr}(R) \wedge
	\m{fl}(F) \wedge
	\m{nn}(R,F,X)$
\exitem{instruct-prags}
	Pragmatics: 
	$\m{obl}(S,H) \wedge
	\m{def}(N) \wedge
	\m{def}(E) \wedge
	\m{def}(R) \wedge
	\m{def}(F) \wedge
	\m{zero-genre}$
\exitem{instruct-instance}
	Instance: 
	$\{ H \leftarrow h0, 
	S \leftarrow s0,
	N \leftarrow n11,
	P \leftarrow p(l(\m{on},j2),l(\m{on},e2)),
	R \leftarrow r11,
	E \leftarrow e2,
	F \leftarrow f4,
	X \leftarrow \m{for}
	\}$
\end{examples}
	(We abbreviate the assertion
   \msref{instruct-meaning}{instruct-assert} by $M$; and abbreviate
   the instance \msref{instruct-meaning}{instruct-instance} by
   $\sigma$.)
   
	\spud\ connects these meanings with domain-specific
   representations as schematized by the inference notation of
   Section~\ref{rep-intro-subsec} and as formalized by the modal logic
   queries described in Section~\ref{context-inference-subsec}.  For
   example, an inference schematized in \sref{slide-assert-inference},
   repeated as \sref{move-inference-tree}, is required to justify the
   assertion-instance
   $\m{move}(a1,h0,n11,p(l(\m{on},j2),l(\m{on},e2)))=\m{move}(a1,H,N,P)\sigma$
   and to link it with one of the system's goals for the instruction.
\sentence{move-inference-tree}{
	\mbox{
	\leaf{$\m{move}(a1,H,N,P)$}
	\branch{1}{\goalinst{\m{move}(a1,h0,n11,p(l(\m{on},j2),l(\m{on},e2)))}}
	\tree}
}
	Concretely this corresponds to two proofs which we obtain from
   the knowledge base $K$:
\begin{examples}{move-inference-q}
\exitem{miq-true}
	$K \proves \speaker \m{move}(a1,H,N,P)\sigma$
\exitem{miq-used}
	$K \proves \shared (\shared (M\sigma \imp
		    \shared \m{move}(a1,h0,n11,p(l(\m{on},j2),l(\m{on},e2)))$
\end{examples}
	The proof \msref{move-inference-q}{miq-true} shows that the
   speaker knows about this motion; the proof
   \msref{move-inference-q}{miq-used} shows that the overall assertion
   of the sentence will add the description of this motion to the
   conversational record.  Note that
   \msref{move-inference-q}{miq-used} relates the overall assertion of
   the utterance to the update achieved by a particular word.  In
   general, we anticipate the possibility that a single
   domain-specific fact may be placed on the conversational record by
   combining the information expressed by multiple words.  For
   example, one word may both provide an inference on its own and
   complete a complex inference in combination with words already in a
   sentence.  We return to this possibility in
   Section~\ref{economy-subsec}.

   	Each conjunct of the assertion in
   \msref{instruct-meaning}{instruct-assert} contributes its inference
   to the system's communicative intent.  In each case, \spud\
   represents the inference portrayed informally as a tree in
   Figure~\ref{instruct-intent-fig-2} as a pair of successful queries
   from $K$, as in \sref{move-inference-q}.

	Next, consider a presupposition, such as the general form
   $\m{nn}(R,F,X)$ and its concrete instance $\m{nn}(r11,f4,\m{for}) =
   \m{nn}(R,F,X)\sigma$.  Corresponding to the informal inference of
   \sref{inference-tree-2} we have the proof indicated in
   \sref{inference-q}.
\sentence{inference-tree-2}{
	\mbox{
	\leaf{\underline{$\m{for}(r11,f4)$}}
	\branch{1}{$\m{nn}(R,F,X)$}
	\tree}
}
\sentence{inference-q}{
	$K \proves \shared \m{nn}(R,F,X)\sigma$
}
	The proof of \sref{inference-q} proceeds by backward chaining
   using the axiom $\shared \forall ab (\m{for}(a,b) \imp
   \m{nn}(a,b,\m{for}))$ and grounds out in the axiom $\shared
   \m{for}(r11,f4)$; hence the correspondence with
   \sref{inference-tree-2}.

	Each conjunct of the presupposition and each conjunct of the
   pragmatics requires a link to the shared context---an inference as
   in \sref{inference-tree-2}---and in each case \spud\ represents
   this link by a successful query as in \sref{inference-q}.

	Appendix~\ref{grammar-sample} gives a grammar fragment
   sufficient to generate \sref{instruction-eg} in \spud.  By
   reference to the trees of this grammar, \spud's complete
   representation of communicative intent for \sref{instruction-eg} is
   given in Figure~\ref{instruct-intent-spud-fig}.
\begin{figure}
\small
\begin{center}
\begin{tabular}{l}
\lxstructuret{\hspace*{-.44in}
\mbox{
\leaf{\begin{tabular}{c}
	Tree \sref{syntax-4}:coupling-nut \\
	by subst. at node $\term{np}$
      \end{tabular}}
\leaf{\begin{tabular}{c}
	Tree \sref{syntax-4}:elbow \\
	by substituting at node $\term{np}$	
      \end{tabular}}
\branch{1}{\begin{tabular}{c}
	Tree \sref{syntax-5a}:onto  \\
	by adjoing at node $\term{vp}_{\ling{path}}$	
      \end{tabular}}
\leaf{\begin{tabular}{c}
	Tree \sref{syntax-6}:fuel-line  \\
	by adjoining at node $\term{n}'$
      \end{tabular}}
\branch{1}{\begin{tabular}{c}
	Tree \sref{syntax-4}: sealing-ring \\
	by subst. at node $\term{np}$
       \end{tabular}}
\branch{1}{\begin{tabular}{c}
	Tree \sref{syntax-3}: uncover \\
	by subst. at node $\term{s}_i$
       \end{tabular}}
\branch{1}{\begin{tabular}{c}
	Tree \sref{syntax-2}: $\langle$purpose$\rangle$  \\
	by adjoining at node $\term{vp}_{\ling{purp}}$
	\end{tabular}}
\branch{3}{\begin{tabular}{c}
	Tree \sref{syntax-1}: slide 
	initial tree
	\end{tabular}}
\tree}} \\[27ex]
\alinks{$\begin{array}{l}
K \proves \speaker \m{move}(a1,H,N,P)\sigma \;\;\;\;\; 
K \proves \speaker \m{next}(a1)\sigma \\
K \proves \shared(\shared M\sigma \imp \shared
   \m{move}(a1,h0,n11,p(l(\m{on},j2),l(\m{on},e2)))) \\
K \proves \shared(\shared M\sigma \imp \shared \m{next}(a1))
\end{array}$ \\[2em]
$\begin{array}{l}
K \proves \speaker \m{purpose}(a1,a2)\sigma \\
K \proves \shared(\shared M\sigma \imp \shared \m{purpose}(a1,a2))
\end{array}$ ~~~~
$\begin{array}{l}
K \proves \speaker \m{uncover}(a2,H,R)\sigma \\
K \proves \shared(\shared M\sigma \imp \shared \m{uncover}(a2,h0,r11))
\end{array}$} \\[9ex]
\pslinks{$K \proves \shared \m{partic}(S,H)\sigma$ ~~~~
$K \proves \shared \m{start-at}(P,N)\sigma$ ~~~~
$K \proves \shared \m{surf}(P)\sigma$ \\
$K \proves \shared \m{cn}(N)\sigma$ ~~~~
$K \proves \shared \m{end-on}(P,E)\sigma$ ~~~~
$K \proves \shared \m{el}(E)\sigma$ \\
$K \proves \shared \m{sr}(R)\sigma$ ~~~~
$K \proves \shared \m{fl}(F)\sigma$ ~~~~
$K \proves \shared \m{nn}(R,F,X)\sigma$}  \\[6ex]
\pglinks{$K \proves \shared \m{obl}(S,H)\sigma$ ~~~~
$K \proves \shared \m{def}(N)\sigma$ ~~~~
$K \proves \shared \m{def}(E)\sigma$ \\
$K \proves \shared \m{def}(R)\sigma$ ~~~~
$K \proves \shared \m{def}(F)\sigma$ ~~~~
$K \proves \shared \m{zero-genre}\sigma$}
\end{tabular}
\end{center}
\caption[\spud's representation of Figure~\ref{instruct-intent-fig-2}.]{%
\spud's representation of the communicative intent in
   Figure~\ref{instruct-intent-fig-2}.  Note two abbreviations for the
   figure: \\[1em]
\hspace*{\fill}
$\begin{array}{c}	
	M := \m{move}(a1,H,N,E) \wedge 
	 \m{next}(a1) \wedge
	\m{purpose}(a1,a2) \wedge
	\m{uncover}(a2,H,R) \\
 	\sigma :=  
	\{ H \leftarrow h0, 
	S \leftarrow s0,
	N \leftarrow n11,
	P \leftarrow p(l(\m{on},j2),l(\m{on},e2)),
	R \leftarrow r11,
	E \leftarrow e2,
	F \leftarrow f4,
	X \leftarrow \m{for}
	\}
\end{array}$
\hspace*{\fill} \\[1em]
	Note also that $K$ refers to the knowledge base specified in
   \sref{instruct-kb-private} and \sref{instruct-kb-public}.
}
\label{instruct-intent-spud-fig}
\end{figure}

\subsection{Recognition of Communicative Intent}
\label{hearer-inference-subsec}

   	Recall from Section~\ref{converse-intro-subsec} that
   structures such as that of Figure~\ref{instruct-intent-spud-fig}
   represent not only the interpretations that speakers intend for
   utterances but also interpretations that hearers can recognize for
   them; in the ideal case, an utterance achieves the updates to the
   conversation that the speaker intends because the hearer
   successfully recognizes the speaker's communicative intent.  In
   generating an utterance, \spud\ anticipates the hearer's
   recognition of its intent by consulting a final, inferential model.

	This model incorporates some simplifications that reflect the
   constrained domains and the constrained communicative settings in
   which NLG systems are appropriate.  Each of these assumptions
   represents a starting point for further work to derive a more
   systematic and more general model of interpretation.
\begin{itemize}
\item
	We assume that the hearer can identify the intended lexical
   elements as contributing to the utterance, and can reconstruct the
   intended structural relationships among the elements.  That is, we
   assume successful parsing and word-sense disambiguation.  On this
   assumption, the hearer always has the correct syntactic structure
   for an utterance and a correct representation of its assertion,
   presupposition and pragmatics.  For example, for utterance
   \sref{instruction-eg} as in Figure~\ref{instruct-intent-spud-fig},
   the hearer gets the syntactic structure of the figure and the three
   conditions of meaning from
   \msref{instruct-meaning}{instruct-assert}--\msref{instruct-meaning}{instruct-prags}.

\item
	We assume that each update that the utterance is intended to
   achieve must either be an instance of an open question that has
   been explicitly raised by preceding discourse, or correspond to an
   assertion that is explicitly contributed by one of the lexical
   elements in the utterance itself.  Once the hearer identifies the
   intended instance of the assertion $M\sigma$, the hearer can arrive
   at the intended update-inferences by carrying out a set of queries
   of the form $\shared(\shared M\sigma \imp \shared Q)$.  Our
   assumption dictates that the set of possible formulas for $Q$ is
   finite and is determined by the hearer's information; we make the
   further assumption that the domain inferences are sufficiently
   short and constrained that the search for each query is bounded (of
   course, the generator requires this to design its
   utterances---whether or not it assesses the hearer's
   interpretation).  The two assumptions justify counting all updates
   as successfully recognized as long as the hearer can recognize the
   intended instance $\sigma$ of the assertion.

\item
	We assume that the hearer attempts to resolve the
   presupposition according to a shared ranking of \textsc{salience}.
   This ranking is formalized using the notion of a \textsc{context
   set}.  Each \textsc{referent}, $\entity$, comes with a context set
   $\distractors(\entity)$ including it and its distractors; the
   context set for $\entity$ determines all the referents that a
   hearer will consider as possible alternatives in resolving a variable
   $X$ that the speaker intends to refer to $\entity$.  This can
   represent a ranking because we can have $\entityA \in
   \distractors(\entityB)$ without $\entityB \in
   \distractors(\entityA)$; in this case $\entityA$ is more salient
   than $\entityB$.  During the reference resolution process, then,
   the hearer might have to run through the context set for $\entityA$
   before expanding the search to include the context set for
   $\entityB$.  In practice, we simply assume that the hearer must
   recognize the context set successfully.  That means that the hearer
   will consider a set of potential resolutions where variables are
   instantiated to elements of appropriate context sets; we represent
   this set of potential resolutions as a set of substitutions
   $D(\sigma)$ defined as follows:
\sentence{d-sigma}{
	$\sigma' \in D(\sigma)$ if and only if for each variable $X$
   that occurs in the presupposition of the utterance, $\sigma'(X) \in
   D(\sigma(X))$
}
	To make this assumption reasonable we have made limited use of
   gradations in salience.

\item
	We assume that the hearer does not use the pragmatic
   conditions in order to determine the speaker's intended
   substitution $\sigma$.  The hearer simply checks, once the hearer
   has resolved $\sigma$ using the presupposition, that there is a
   unique inference that justifies the corresponding instance of the
   pragmatics.
\end{itemize}
	It follows from these assumptions that interpretation is a
   constraint-satisfaction problem, as in
   \cite{mellish:descriptions,haddock:thesis,dale/haddock:referring}.
   In particular, the key task that the hearer is charged with is to
   recognize the inferences associated with the presupposition of the
   utterance.  That presupposition is an open formula $P$ composed of
   the conjunction of the individual presupposition formulas $P_i$
   contributed by lexical elements.  The resolutions compatible with
   the hearer's information about the utterances are the instances of
   $P$ that fit the conversational record and the attentional state of
   the discourse.  Formally, we can represent this as $\Sigma'$ defined
   in \sref{constraint-correct}.
\sentence{constraint-correct}{
	$\Sigma' := \{ \sigma' \in D(\sigma) : K \proves \shared
   P\sigma' \}$
}

	Each of the formulas $P_i$ determines a relation $R_i$ on
   discourse referents that characterizes instances that the speaker
   may have intended; \spud\ computes this relation by querying the
   knowledge base as in \sref{constraint-query}, and represents it
   compactly in terms of the free variables that occur in $P_i$.
\sentence{constraint-query}{
	$R_i = \{ \sigma' \in D(\sigma) : K \proves \shared P_i\sigma' \}$
}
   	\spud\ then uses an arc-consistency constraint-satisfaction
   heuristic on these relations to solve for $\Sigma'$
   \cite{mackworth:constraint}.  (This is a conservative but efficient
   strategy for eliminating assignments that are inconsistent with the
   constraints.)  \spud\ counts the inferences for the presupposition
   as successfully recognized when the arc-consistency computation
   leaves only a single possibility, namely the intended resolution
   $\sigma$.

\section{Microplanning as a Search Task}
\label{algorithm-section}

   	The preceding sections have been leading up to a 
   characterization of microplanning as a formal search task
   \cite{nilsson:ai71}.  
	We argued in Section~\ref{problem-sec} that a generator must
   represent the interpretation of an utterance as a data structure
   which records inferences that connect the structure of an utterance
   with its meaning, ground the meaning of an utterance in the current
   context, and draw on the meaning of the utterance to register
   specified information in the conversational record.
	In Section~\ref{grammar-section}, we described the grammatical
   knowledge which defines the structure and meaning of utterances; in
   Section~\ref{context-inference-subsec}, we described the
   inferential mechanisms which encode the relationships between
   utterance meaning and an evolving conversational record.  With
   these results, we obtain the specific data structure that \spud\
   uses to represent communicative intent, in the kinds of records
   schematized in Figure~\ref{instruct-intent-spud-fig}; and the
   concrete operations that \spud\ uses to derive representations of
   communicative intent, by the steps of grammatical composition and
   contextual inference described in
   Sections~\ref{grammar-inference-subsec},
   \ref{context-inference-subsec}, and \ref{hearer-inference-subsec}.
   	Thus, we obtain a characterization of the microplanning
   problem as a \textsc{search}, whose \textsc{result} is an
   appropriate communicative-intent data structure, and which
   \textsc{proceeds} by steps of grammatical derivation and contextual
   inference.

\subsection{A Formal Search Problem}
\label{formal-search-subsec}

   	In \spud, the specification of a microplanning search problem
   consists of the following components:
\begin{examples}{problem-eg}
\exitem{problem-space-eg}
	a background specification of a \textsc{grammar} $G$
   describing the system's model of language (as outlined in
   Section~\ref{grammar-section}) and a \textsc{knowledge base} $K$
   describing the system's model of its domain, its user and the
   conversational record (as outlined in
   Section~\ref{context-inference-subsec});
\exitem{problem-goal-eg}
	a set of formulas, \textsc{updates}, describing the
   specified facts that the utterance must add to the conversational
   record;
\exitem{problem-start-eg}
	a specification of the \textsc{root node} of the syntactic
   tree corresponding to the utterance.  This specification involves a
   syntactic category; variables specifying the indices of the root
   node; a substitution $\sigma_0$ describing the intended values that
   those variables must have; and a top feature structure, indicating
   syntactic constraints imposed on the utterance from the external
   context; cf.~\sref{tree-internal-general}.
\end{examples}
	For instance, we might specify the task of describing the
   sliding action $a1$ by an instruction such as \sref{instruction-eg}
   as follows.
\begin{examples}{problem-eg-eg}
\exitem{problem-space-eg-eg}
	The \textsc{grammar} $G$ outlined in
   Appendices~\ref{grammar-sample} and \ref{motion-sample}; the
   knowledge base outlined in \sref{instruct-kb-private},
   \sref{instruct-kb-public}, and \sref{instruct-kb-alts}.
\exitem{problem-goal-eg-eg}
	Four \textsc{updates}:
   $\m{move}(a1,h0,n11,p(l(\m{on},j2),l(\m{on},e2)))$; $\m{next}(a1)$;
   $\m{purpose}(a1,a2)$; $\m{uncover}(a2,h0,r11)$.
\exitem{problem-start-eg-eg}
	A root node $\term{s} \downarrow (E)$ with intended instance
   $\{ E \leftarrow a1 \}$.
\end{examples}

	The grammar and knowledge base of
   \msref{problem-eg}{problem-space-eg} determine the search space
   for the NLG task.  States in the search space are data structures
   for communicative intent, as argued for in
   Section~\ref{problem-sec} and as illustrated in
   Section~\ref{data-structure-subsec}.  In particular, each state
   involves:
\begin{examples}{search-state}
\item
	a syntactic structure $T$ derived according to $G$ and paired
   with a meaning $\langle A, P, Q \rangle$ giving the assertion,
   presupposition and pragmatics of $T$ (respectively);
\item
	a substitution $\sigma$ determining the discourse referents
   intended for the variables in $A$, $P$, and $Q$;
\item
	inferences $K \proves \speaker A\sigma$, $K \proves \shared
   P\sigma$, and $K \proves \shared Q\sigma$---such inferences show
   that the context supports use of this utterance to describe
   $\sigma$;
\item
	inferences of the form $K \proves \shared (\shared A\sigma
   \proves \shared F)$ where $F$ is an update---such inferences
   witness that the utterance supplies needed information;
\item
	a constraint network approximating $\Sigma' := \{\sigma' \in
   D(\sigma) : K \proves \shared P\sigma' \}$---this network
   represents the hearer's interpretation of reference resolution.
\end{examples}

	The \term{initial state} for search is given in
   \sref{initial-search-state}.
\begin{examples}{initial-search-state}
\item
	a syntactic structure consisting of a single substitution site
   matching the root node of the problem specification
   \msref{problem-eg}{problem-start-eg} and paired with an empty
   meaning;
\item
	the specified intended resolution $\sigma_0$ of variables in
   this syntactic structure;
\item
	no inferences---a record that suffices to justify the empty
   meaning of the initial state but which shows that this state
   supplies no needed information;
\item
	an unconstrained network realizing $\Sigma' := \{ \sigma' \in
   D(\sigma_0) \}$.
\end{examples}

	A \term{goal state} for search is one where the three conditions
   of \sref{goal-eg} are met.
\begin{examples}{goal-eg}
\exitem{goal-syntax}
   	The syntactic structure of the utterance must be complete: top
   and bottom features of all syntactic nodes must agree, and all
   substitution sites must be filled.

\exitem{goal-comm}
	For each update formula $F$, the communicative intent must
   include an update inference that establishes a substitution
   instance of $F$.  More formally, on the assumption that $M$ is the
   assertion of the utterance and that $\sigma$ is the intended
   instance of $M$, the requirement is that the communicative intent
   include an inferential record of the form $K \proves
   \shared(\shared M\sigma \imp \shared F\sigma')$.

\exitem{goal-ident}
	The arc-consistency approximation to the key
   presupposition-recognition problem the hearer faces for the
   communicative intent, as defined in
   Section~\ref{hearer-inference-subsec}, identifies uniquely the
   intended substitution of knowledge-base discourse referents for
   discourse-anaphor variables in the utterance.
\end{examples}
	The requirements of \sref{goal-eg} boil down simply to this:
   the generator's communicative intent must provide a complete
   sentence \msref{goal-eg}{goal-syntax} that says what is needed
   \msref{goal-eg}{goal-comm} in a way the hearer will understand
   \msref{goal-eg}{goal-ident}.  Observe that the communicative intent
   of Figure~\ref{instruct-intent-spud-fig} fulfills the conditions in
   \sref{goal-eg} for the microplanning problem of
   \sref{problem-eg-eg}.

   	To derive a new state from an existing state as in
   \sref{search-state} involves the steps outlined in
   \sref{neighbor-eg}.
\begin{examples}{neighbor-eg}
\item
	Construct a lexico-grammatical element $L$, according to the
   steps of \sref{combo-steps}.
\item
	Apply a syntactic operation combining $L$ with the existing
   syntactic structure $T$
   (cf.~Section~\ref{grammar-inference-subsec}); the result is a new
   structure $T'$ and a new meaning $\langle A\wedge A', P\wedge P',
   Q\wedge Q'\rangle$ that takes into account the contribution
   $\langle A',P',Q'\rangle$ of $L$.
\item
	Ensure that the use of this element is supported in context,
   by proving $K \proves \speaker A'\sigma$, $K \proves \shared
   P'\sigma$ and $K \proves \shared Q'\sigma$; the result is a refined
   substitution $\sigma'$ describing the intended instantiation not
   just of $T$ but also of $L$.
\item
	Record the communicative effects of the new structure in any
   inferences $K \proves \shared(\shared(A\wedge A')\sigma' \proves
   \shared F)$ for outstanding updates $F$.
\item
	Refine the constraint network to take into account the new
   constraint $P'$.
\end{examples}
	Any state so derived from a given state is called a
   \term{neighbor} of that state.

   	Because such searches begin at an initial substitution site
   and derive neighbors by incorporating single elements into the
   ongoing structure, this characterization of microplanning in terms
   of search builds in \spud's head-first derivation strategy.  On the
   other hand, it is compatible with any search algorithm, including
   brute-force exhaustive search, a traditional heuristic search
   method such as \term{a$^*$} \cite{hart:astar}, or a stochastic
   optimization search \cite{ilex:inlg98a}.

\subsection{A Greedy Search Algorithm}
\label{greedy-search-subsec}

   	We chose to implement a greedy search algorithm in \spud.
   Greedy search applies iteratively to update a single state in the
   search space, the \term{current state}.  In each iteration, greedy
   search first obtains all the neighbors of the current state.
   Greedy search then ranks the neighbors by a heuristic evaluation
   intended to assess progress towards reaching a goal state.  The
   neighbor with the best heuristic evaluation is selected.  If this
   state is a goal state, search terminates; otherwise this state
   becomes the current state for the following iteration.

   	In developing \spud, we have identified a number of factors
   that give evidence of progress towards obtaining a complete,
   concise, natural utterance that conveys needed information
   unambiguously. 
\begin{enumerate}
\item
	How many update formulas the utterance has conveyed.  Other
   things being equal, if fewer updates remain unrealized, then fewer
   steps of lexical derivation will be required to convey this further
   required information.

\item
	How many alternative values the hearer could consider for each
   free variable which the system must resolve.  Other things being
   equal, the fewer values remain for each variable, the fewer steps
   of lexical derivation will be required to supply content that
   eliminates the ambiguity for the hearer.  The concrete measure for
   this factor in \spud\ is a sorted list containing the number of
   possible values for each ambiguous variable in the constraint
   network; lists are compared by the lexicographic ordering.

\item
	How \textsc{salient} the intended values for each free
   variable are.  Other things being equal, an utterance referring to
   salient referents may prove more coherent and easier for the hearer
   to resolve (irrespective of its length).  Again, the concrete
   measure for this factor in \spud\ is a sorted list of counts,
   compared lexicographically; the counts here are the sizes of
   context sets for each intended referent.

\item
	How many \textsc{flaws} remain in the syntactic structure of
   the utterance.  Flaws are open substitution sites and internal
   nodes whose top and bottom features do not unify.  Each flaw can
   only be fixed by a separate step of grammatical derivation.  Other
   things being equal, the fewer flaws remain, the fewer further
   syntactic operations will be required to obtain a complete
   grammatical utterance.  We also prefer states in which an existing
   flaw has been corrected but new flaws have been introduced, over a
   structure with the same overall number of flaws but where the last
   step of derivation has not resolved any existing flaws.

\item
	How \textsc{specific} the meanings for elements in the
   utterance are.  In general, an element with a more specific
   assertion offers a more precise description for the hearer; an
   element with a more specific presupposition offers more precise
   constraints for identifying objects; an element with a more
   specific pragmatic conditions fits the context more precisely.  We
   assess specificity off-line using the semantic information
   associated with the operator $\meanings$ .  If the query $?K
   \proves \meanings (M \imp N)$ succeeds, we count formula $M$ as at
   least as specific as $N$.  We prefer words with more specific
   pragmatics; then (other things being equal) words with more
   specific presuppositions; then (other things being equal) words
   with more specific content; then (other things being equal) words
   in constructions with more specific pragmatics.
\end{enumerate}	
   In our implementation of \spud, we use all these criteria,
   prioritized as listed, to rank alternative options.  That is,
   \spud\ ranks option $S$ ahead of option $S'$ if one of these
   factors favors $S$ over $S'$ and all factors of higher priority are
   indifferent between $S$ and $S'$.%
\dspftnt{
	It happens that this is also the treatment of ranked
   constraints in optimality theory \cite{prince/smolensky:science97}!
}

   	In designing \spud\ with greedy search, we drew on the
   influential example of \cite{dale/haddock:referring}, which used
   greedy search in referring expression generation; and on our own
   experience using greedy algorithms to design preliminary plans to
   achieve multiple goals \cite{webber:traumaid/aij98}.  As described
   in Sections~\ref{example-section} and~\ref{build-sec}, we believe
   that our experience with \spud\ supports our decision to use a
   sharply constrained search strategy; consistent search behavior
   makes it easier to understand the behavior of the system and to
   design appropriate specifications for it.  However, we do
   \term{not} claim that our experience offers a justification for the
   specific ranking we used beyond two very general preferences---a
   primary preference for adding lexical elements that make some
   progress on the generation task over those that make none (on
   syntactic, informational or referential grounds); and a secondary
   preference based on pragmatic specificity.  In general, the
   relationships between search algorithms, specification development
   and output quality for microplanning based on communicative intent,
   remains an important matter for future research.

\section{Solving \term{nlg} tasks with \spud}
\label{example-section}

   	In this section, we support our claims that decision-making
   based on communicative intent provides a uniform framework by which
   which \spud\ can simultaneously address all the subtasks of
   microplanning.  We further argue that such a framework is essential
   for generating utterances that are \textsc{efficient}, in that they
   exploit the contribution of a single lexico-grammatical element to
   multiple goals and indeed to multiple microplanning subtasks.
   Throughout the section, we illustrate how \spud's grammatical
   resources, inference processes, and search strategy combine to
   solve these problems together for instruction
   \sref{instruction-eg}.  Additional examples of using \spud\ in
   generation can be found in \cite{bourne:phd,cassell:inlg00}; we
   also investigate these issues from the perspective of designing
   specifications for \spud\ in Section~\ref{build-sec}.

\subsection{Referring Expressions}
\label{ref-subsec}

   	The problem of generating a referring expression for a simple
   (i.e., non-event) discourse referent $a$ is to devise a description
   that can be realized as a noun phrase by grammar $G$ and that
   uniquely identifies $a$ in context $K$.  Such a problem can be
   posed to \spud\ by the problem specification of
   \sref{ref-np-subtask}.
\begin{examples}{ref-np-subtask}
\item
	the grammar $G$ and context $K$ 
\item
	no updates to achieve
\item
	an initial node $\term{np}\downarrow(X)$ and an initial
   substitution $\sigma_0 = \{ X \leftarrow a \}$
\end{examples}
	By the criteria of \sref{goal-eg}, a solution to this task is
   a record of communicative intent which specifies a complete grammatical
   noun phrase and which determines a constraint-satisfaction network
   that identifies a unique intended substitution, including the
   assignment $X \leftarrow a$.

\begin{figure}
\begin{center}
\mbox{
\psfig{figure=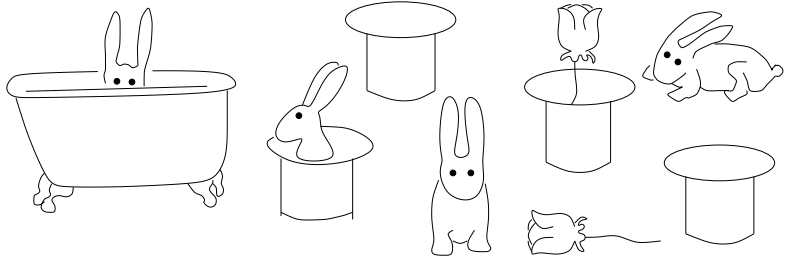,scale=.30,silent=}
}
\end{center}
\vspace*{-2ex}
\caption{A representation of the context for a referring expression
   generation task}
\label{rabbit:fig}
\vspace*{-1ex}
\end{figure}

   	The following example demonstrates the close affinity between
   \spud's strategy and the algorithm of
   \cite{dale/haddock:referring}.  In Figure~\ref{rabbit:fig}, we
   portray a context $K$ which supplies a number of salient
   individuals, including a rabbit $r1$ located in a hat $h1$; $K$
   records each individual with visual properties such as kind, size,
   and location.
	We consider the problem of generating a referring expression
   to identify $r1$.

\begin{figure}
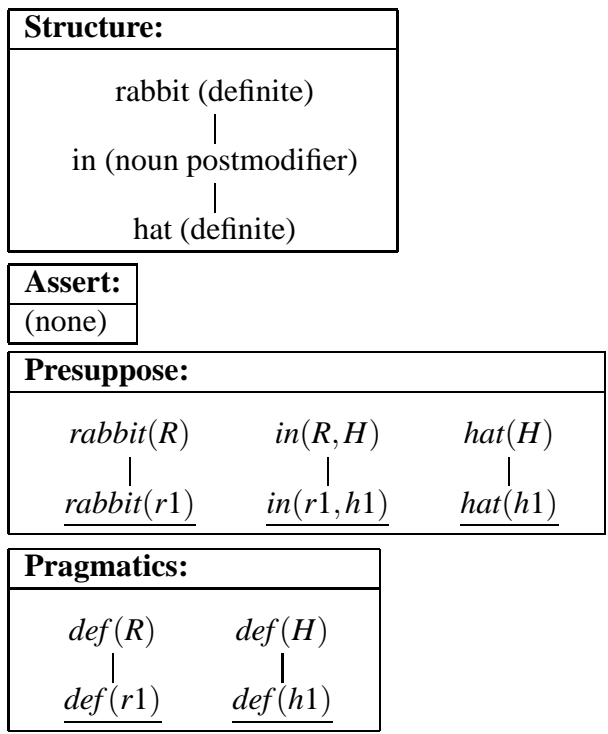

\begin{center}
\begin{tabular}{l}
\lxstructuret{\mbox{
\leaf{hat (definite)}
\branch{1}{in (noun postmodifier)}
\branch{1}{rabbit (definite)}
\tree}}
\\[8ex]
\alinks{(none)} \\[2ex]
\pslinkst{
	\mbox{
	\leaf{\sharedinst{\m{rabbit}(r1)}}
	\branch{1}{$\m{rabbit}(R)$}
	\tree}
	\mbox{
	\leaf{\sharedinst{\m{in}(r1,h1)}}
	\branch{1}{$\m{in}(R,H)$}
	\tree}
	\mbox{
	\leaf{\sharedinst{\m{hat}(h1)}}
	\branch{1}{$\m{hat}(H)$}
	\tree}
} \\[6ex]
\pglinkst{
	\mbox{
	\leaf{\sharedinst{\m{def}(r1)}}
	\branch{1}{$\m{def}(R)$}
	\tree}
	\mbox{
	\leaf{\sharedinst{\m{def}(h1)}}
	\branch{1}{$\m{def}(H)$}
	\tree}
}
\end{tabular}
\end{center}
\caption{Communicative intent for \emph{the rabbit in the hat}.}
\label{rabbit:ci}
\end{figure}

	With a suitable grammar, $K$ allows us to construct the
   communicative intent schematized in Figure~\ref{rabbit:ci} for
   \sref{rabbit-np}.
\sentence{rabbit-np}{
	the rabbit in the hat
}
	\spud's model of interpretation, like Dale and Haddock's,
   predicts that the hearer successfully recognizes this communicative
   intent, because the context supplies a unique pair of values for
   variables $R$ and $H$ such that $R$ is a rabbit, $H$ is a hat, and
   $R$ is in $H$.  Thus, \sref{rabbit-np} represents a potential
   solution to the reference task both for \spud\ and for Dale and
   Haddock.

	In fact, in deriving \emph{the rabbit in the hat}, the two
   algorithms would use parallel considerations to take comparable
   steps.  \spud's derivation, like Dale and Haddock's, consists of
   three steps in which specific content enriches a description: first
   \emph{rabbit}, then \emph{in} and finally \emph{hat}.  For both
   algorithms, the primary consideration to use these steps of
   derivation is that each narrows the domain of values for variables
   more than the available alternative steps.

   	We note three important contrasts between \spud's approach and
   Dale and Haddock's, however.  First, \spud\ typically formulates
   referring expressions not in isolated subtasks as suggested in
   \sref{ref-np-subtask} but rather as part of a single, overall
   process of sentence formulation.  \spud's broader view is in fact
   necessary to generate instructions such as
   \sref{instruction-eg}---a point we return to in detail in
   Section~\ref{economy-subsec}.

	Second, \spud's options at each step are determined by
   grammatical syntax, whereas Dale and Haddock's must be determined
   by a separate specification of possible conceptual combinations.
   For example, \spud\ directly encodes the syntactic requirement that
   a description should have a head noun using the \term{np}
   substitution site; for Dale and Haddock this requires an \emph{ad
   hoc} restriction on what concepts may be included at certain stages
   of description.

	Third, Dale and Haddock adopt a fixed, depth-first strategy
   for adding content to a description.  Particularly since
   \cite{dale/reiter:cogsci}, such fixed (and even domain-specific)
   strategies have become common for referring expressions made up of
   properties of a single individual.  It is difficult to generalize a
   fixed strategy to relational descriptions, however.  Indeed,
   Horacek \cite{horacek:referring} challenges fixed strategies with
   examples that show the need for modification at multiple points in
   an \term{np}, such as \sref{Horacek:eg}.
\sentence{Horacek:eg}{
        the table with the apple and with the banana
}
	In \spud, the order of adding content is flexible.  An LTAG
   derivation allows modifiers to adjoin at any node at any step of
   the derivation.  This places descriptions such as \sref{Horacek:eg}
   within \spud's search space.  (\spud's flexibility also contrasts
   with a top-down derivation in a context-free grammar, where
   modifiers must be chosen before heads and there is a resulting
   tension between providing what the syntax requires and going beyond
   what the syntax requires.  See \cite{elhadad/robin:fuf} for
   discussion of the resulting difficulties in search.)

\subsection{Syntactic Choice}
\label{syntax-subsec}

   	The problem of syntactic choice is to select an appropriate
   grammatical construction in which to realize a given lexical item.
   For example, for English noun phrases, the problem is to select an
   appropriate determiner from among options including the indefinite
   marker \ling{a}, the definite marker \ling{the} and the
   demonstrative markers \ling{this} and \ling{that}.  With main verbs
   in English sentences, the problem involves such decisions as the
   appropriate use of active or passive voice, and the appropriate
   fronting or preposing of marked argument constituents.

	For \spud, alternatives for such syntactic choices are
   represented as alternative states which \spud's greedy search must
   consider at some stage of generation.  All alternative syntactic
   entries whose pragmatic conditions are supported in the context
   will be available.  Since these syntactic alternatives share a
   common lexical specification, their interpretations differ only by
   the contribution of the distinct pragmatic conditions.  Recall that
   the pragmatics contributes neither to the updates that an utterance
   achieves nor to the resolution of referential ambiguity, in \spud's
   model of interpretation.  Accordingly, \spud's ranking of these
   alternatives is based only on the specificity of the pragmatic
   conditions.  \spud's strategy for syntactic choice is to select a
   licensed form whose pragmatic condition is maximally specific.

	As an illustration of this strategy, consider the syntactic
   frame for the verb \ling{slide} in instruction
   \sref{instruction-eg}.  The instruction exhibits the imperative
   frame \ling{slide \textsc{np}}.  Recall that we associate this
   frame semantically with the condition that a sliding is the next
   action that the hearer should perform; we associate it with the
   pragmatic condition that the speaker is empowered to impose
   obligations for action on the hearer.  This pragmatic condition
   distinguishes \ling{slide \textsc{np}} from other possible
   descriptions of this action.  One such possibility is \ling{you
   should slide \textsc{np}}; we would represent this as a neutral
   alternative with an always true pragmatic condition.  Thus, when
   \spud\ considers both alternatives, it favors \ling{slide
   \textsc{np}} because of its specific pragmatics.  (In
   \cite{gen-paper}, we consider choice of a topicalized frame,
   represented with the pragmatic conditions proposed for
   topicalization in \cite{ward:thesis}, over an unmarked frame; we
   describe how the generation of \ling{the syntax book, we have}
   follows from this specification under \spud's preference for
   specificity.)

	Syntactic frames for the noun phrases provide a similar
   illustration.  Noun phrases in our aircraft maintenance manuals are
   realized in one of two frames: a zero definite realization for a
   unique referent, as in \ling{coupling nut}, and a realization with
   an explicit numeral, used in the other cases (plural referents,
   such as \ling{two coupling nuts}, and indefinite singular
   referents, such as \ling{one coupling nut}).  We associate the zero
   definite realization with a pragmatic condition, as in
   \sref{const-example}, requiring a definite referent and an
   appropriate linguistic genre; the realization with the explicit
   numeral is a default whose pragmatic conditions are always
   satisfied for this genre.  The zero definite is chosen whenever
   applicable, by specificity.  More generally, whichever of the two
   entries, the zero-definite noun phrase or the numerical noun
   phrase, best applies to a referent in the maintenance domain,
   \spud\ will prefer that entry to the corresponding ordinary
   definite (\ling{the}) or indefinite (\ling{a}) noun-phrase entry.
   The genre-restricted entry carries a pragmatic condition on genre
   which the ordinary entry lacks; thus the genre-restricted entry is
   selected as more specific.

        We credit to systemic linguistics the idea that choices in
   syntactic realization should be made incrementally, by consulting a
   model of the discourse and a specification of the functional
   consequences of grammatical choices.  \cite{mathiessen:nigel} is a
   classic implementation for generation, while
   \cite{yang:systemic/tag} explores the close connection between
   systemic linguistics and TAG.  However, \spud\ departs from the
   systemic approach in that pragmatic conditions are associated with
   individual constructions rather than linguistic systems; this
   departure also necessitates \spud's criterion of specificity.
   Inspiration for both of these moves can be found in such recent
   research on the discourse function of syntactic constructions as
   \cite{prince:open-prop,hirschberg:thesis,ward:thesis,gundel/hz:referring,birner:thesis}.
   More generally, as hinted in our contrast of zero-definite noun
   phrases versus \ling{the} noun phrases, we hypothesize that
   pragmatically-conditioned constructions, selected in context by
   specificity, make for grammars that can incorporate general
   defaults in realization while also modeling the tendency of
   specific genres or sublanguages to adopt characteristic styles of
   communication \cite{kittredge:domain}.  This hypothesis merits
   further detailed investigation.

\subsection{Lexical Choice}
\label{lex-choice-subsec}

   	Problems of lexical choice arise whenever a microplanner must
   apportion abstract content onto specific lexical items that carry
   this content (in context).  Our model of this problem follows
   \cite{elhadad:constraints}.  According to this approach, in lexical
   choice, the microplanner must select words to contribute several
   independently-specified conditions to the conversational record.
   Some of these conditions characteristically ``float'', in that they
   tend to be realized across a range of syntactic constituents at
   different linguistic levels, and tend to be realized by lexical
   items that put other needed information on the record.  We agree
   with the argument of Elhadad et al. that a solution to such
   problems depends on declarative conceptual and linguistic
   descriptions of lexical items and accurate assessments of the
   contribution of lexical items to interpretation.  (We agree further
   that this lexical choice cannot be solved as an isolated
   microplanning subproblem, and must be solved concurrently with such
   other tasks as syntactic choice.)

   	Elhadad et al.'s example is \sref{elhadad-ai}; the sentence
   adopts an informal and concise style to describe an AI class for an
   academic help domain.
\sentence{elhadad-ai}{
	AI requires six assignments.
}
	The choice of verb \ling{requires} here responds to two
   generation goals.  First, it conveys simply that the AI class
   involves a given set of assignments.  The generator has other
   lexical alternatives, such as \ling{y has x} or \ling{there are x
   in y}, that do the same.  In addition, \ling{requires} conveys that
   the assignments represent a significant demand that the class
   places on its students.  This second feature distinguishes
   \ling{requires} from alternative lexical items and accounts for the
   generator's selection of it.

	Both for Elhadad et al. and for \spud, the selection of
   \ling{requires} for \sref{elhadad-ai} depends on its lexical
   representation, which must spell out the two contributions the verb
   can make.  In \spud, these contributions can be represented as
   assertions made when using \ling{require} to describe a state $S$
   associating a class $C$ with assignments $A$, as in
   \sref{require-assert}.
\sentence{require-assert}{
	Assertion: $\m{involve}(S,C,A) \wedge \m{demand}(A)$
}
	Meanwhile, a microplanning task might begin with goals to
   convey two specific instances about the AI class, \m{c1}; its
   assignments, \m{a1}; and an eventuality, \m{s1}, as in
   \sref{require-goals}.
\begin{examples}{require-goals}
\item
	$\m{involve}(s1,c1,a1)$
\item
	$\m{demand}(a1)$
\end{examples}
	In a context which supplies the information in
   \sref{require-goals}, \spud\ can add an instance of \ling{require}
   as in \sref{require-assert} to augment a sentence about $s1$; the
   instantiation $\sigma$ has $\{ S \leftarrow s1, C \leftarrow c1, A
   \leftarrow a1 \}$.  Using $M$ to abbreviate the \ling{require}
   assertion from \sref{require-assert}, \spud's assessment of
   interpretation now records the completed inferences in
   \sref{require-inf}.
\begin{examples}{require-inf}
\item
	$\shared (M\sigma \imp \m{involve}(s1,c1,a1))$
\item
	$\shared (M\sigma \imp \m{demand}(a1))$
\end{examples}
	Thus, \spud\ recognizes the opportunistic dual contribution of
   \ling{require}, and will therefore prefer \ling{require} to other
   lexical alternatives that do not make a similar contribution.
   
	Despite the high-level similarity, \spud's mechanisms for
   grammatical and contextual inference are quite different to those
   of \cite{elhadad:constraints}.  Elhadad et al. achieve flexibility
   of search by logic-programming constructs that allow programmers to
   state meaningful dependencies and alternatives in the generator's
   decisions in constructing a context-free phrase structure by
   top-down traversal.  For \spud, dependencies and alternatives are
   represented using the extended domain of locality of LTAG; \spud's
   strategy for updating decisions about the linguistic realization of
   floating constraints thus depend on its LTAG derivation and
   incremental interpretation.

	Moreover, because \spud's model of interpretation is broader,
   we account for more diverse interactions in microplanning; we
   explore this in more detail in Section~\ref{economy-subsec} and
   explore its consequences for the design of \spud\ specifications
   for lexical choice in Section~\ref{semantics-section}.

\subsection{Aggregation}
\label{aggregation-subsec}
   
	The microplanning process of aggregation constructs complex
   sentences in which assemblies of lexical items achieve multiple
   simultaneous updates to the conversational record.  Instruction
   \sref{instruction-eg} represents a case of aggregation because the
   combination of \ling{slide}, a bare infinitival purpose clause, and
   \ling{uncover} conveys four updates to the conversational record
   with a single sentence: the \ling{next} event is a \ling{sliding}
   whose \ling{purpose} is an \ling{uncovering}.

	Aggregation is so named because many microplanners produce
   complex sentences through syntactic operations that combine
   together, or aggregate, specifications of simple linguistic
   structures \cite{reiter/dale:bnlg}.  For example, such a system
   might derive instruction \sref{instruction-eg} by stitching
   together specifications for these simple sentences: \ling{slide the
   coupling nut to the elbow}; \ling{the sliding has a purpose};
   \ling{the purpose is uncovering the sealing ring}.  Each of these
   sentence specifications directly corresponds to a single given
   update.  The specifications can be combined by describing
   transformations that create embedded syntactic structures under
   appropriate syntactic, semantic and pragmatic conditions.

	In \spud, aggregation is not a distinct stage of microplanning
   that draws on idiosyncratic linguistic resources; instead,
   aggregation arises as a natural consequence of the incremental
   elaboration of communicative intent using a grammar.  Initial
   phases of lexicalization leave some updates unexpressed; for
   example, after \spud's selection in \sref{instruction-eg} of the
   imperative transitive verb \ling{slide}, \spud\ still has the goals
   of updating the conversational record to the event's
   \ling{purpose}, of \ling{uncovering}.  These lexical and syntactic
   decisions also trigger new grammatical entries that adjoin into
   \spud's provisional linguistic structure and augment the
   provisional communicative intent.  Such entries provide the
   grammatical resources by which \spud's subsequent lexicalization
   decisions can directly contribute to complex sentences that achieve
   multiple communicative goals.

	For example, in \sref{instruction-eg}, \ling{slide} introduces
   a $\textsc{vp}_{\m{purp}}$ node indexed by the sliding event $a1$
   and its agent $h0$.  This is a site where the lexico-syntactic
   entry in \sref{to-entry} could adjoin.
\begin{examples}{to-entry}
\exitem{to-entry-tree}
\term{tree:}
\mbox{\small
	\leaf{$\term{vp}_{\m{purp}}(A1,H)*$}
	\leaf{$\term{s}_i(A2,H)\downarrow$}
	\branch{2}{$\term{vp}_{\m{purp}}(A1,H)$}
\tree
}
\exitem{to-entry-target}
	\term{target}:
	$\term{vp}_{\m{purp}}(A1,H)$
\exitem{to-entry-assertion}
	\term{assertion}:
	$\m{purpose}(A1,A2)$
\exitem{to-entry-context}
	\term{presupposition} and \term{pragmatics}: ---
\end{examples}
	\sref{to-entry} is a declarative description of the form and
   meaning of an English bare infinitival purpose construction,
   expressed in the general terms required for reasoning about the
   interpretation of assemblies of linguistic constructions in
   context.  Specifically, \msref{to-entry}{to-entry-tree} assumes
   that the purpose clause modifies a specific \textsc{vp} node and
   subcategorizes for an infinitive \textsc{s}.%
\dspftnt{
	Lexicalization purists could add a covert subordinating
   conjunction to head the tree in \msref{to-entry}{to-entry-tree},
   but \spud\ does not require it.
}

	At the same time, \sref{to-entry} also has an operational
   interpretation for generation, as a pattern of possible
   aggregation: \sref{to-entry} describes when and how a description
   of an event can be extended to include a characterization of the
   purpose of the event.  This operational interpretation provides a
   complementary motivation for each of the constituents of
   \sref{to-entry}.  An aggregation pattern must indicate how new
   material can be incorporated into an existing sentence; this is the
   role of the target in \msref{to-entry}{to-entry-target}.  And it
   must indicate what updates are realized by the addition; this is
   the role of the assertion in \msref{to-entry}{to-entry-assertion}.

	More generally, an aggregation pattern must indicate how the
   syntactic realization of aggregated material depends on its
   subordination to or coordination with other linguistic structure.
   Languages generally offer lightweight constructs, such as
   participles and prepositional phrases, which augment a sentence
   with less than another full clause.  Syntactic trees such as that
   in \msref{to-entry}{to-entry-tree} provide a natural specification
   of these constructs.  Finally, the pattern must characterize the
   idiosyncratic interpretive constraints that favor one aggregated
   realization over another.  Not all realizations are equally good;
   alternatives may require specific informational or discourse
   relationships, such as the inferrability between events that some
   adjuncts demand \cite{cheng/mellish:inlg00}.  As an aggregation
   pattern, \sref{to-entry} represents such characterizations of
   requirements on context by appropriate pragmatic conditions or
   presuppositions.
   
	Selecting entry \sref{to-entry} is \spud's analogue of an
   aggregation process; by using it, \spud\ derives a provisional
   sentence including \ling{slide} and requiring a further infinitive
   clause.  \spud\ substitutes \ling{to uncover} for the infinitive
   sentence in the purpose clause in a subsequent step of
   lexicalization.  This grammatical derivation results in a single
   complex sentence that achieves four updates to the conversational
   record.

\subsection{Interactions in Microplanning}
\label{economy-subsec}

   	\spud\ is capable of achieving specified behavior on isolated
   microplanning tasks, but a key strength of \spud\ is its ability to
   model \term{interactions} among the requirements of microplanning.
   Different requirements can usually be satisfied in isolation by
   assembling appropriate syntactic constituents---for example, by
   identifying an individual using a noun phrase that refers to it or
   by communicating a desired property of an action using a verb
   phrase that asserts it.  However, many sentences exhibit an
   alternative, more efficient strategy which we have called
   \term{textual economy}: the sentences satisfy some microplanning
   objectives implicitly, by exploiting the hearer's (or reader's)
   recognition of inferential links to material elsewhere in the
   sentence that is there for independent reasons \cite{genw-paper}.
   Such material is therefore {\em overloaded} in the sense of
   \cite{pollack:overloading}.%
\dspftnt{
	Pollack used the term {\em overloading} to refer to cases
   where a single intention to act is used to wholly or partially
   satisfy several of an agent's goals simultaneously.
}

	The main clause of \sref{instruction-eg}, repeated as
   \sref{instruction-rep-eg}, is in fact illustrative of textual
   economy that exploits interactions among problems of referring
   expression generation and lexical choice within a single clause.
\sentence{instruction-rep-eg}{
	Slide coupling nut onto elbow.
}
	Consider the broader context in which
   \sref{instruction-rep-eg} will be used to instruct the action the
   depicted in Figure~\ref{instruct:fig1}.  Given the frequent use of
   coupling nuts and sealing rings to join vents together in aircraft,
   we cannot expect this context to supply a single, unique coupling
   nut.  Indeed, diagrams associated with instructions in our aircraft
   manuals sometimes explicitly labeled multiple similar parts.
   Allocating tasks of verb choice and referring expression generation
   to independent constituents in such circumstances would therefore
   lead to unnecessarily verbose utterances like
   \sref{instruction-redundant-eg}.
\sentence{instruction-redundant-eg}{
	Slide coupling nut that is over fuel-line sealing ring onto
   elbow.
}

	Instead, it is common to find instructions such as
   \sref{instruction-rep-eg}, in which these parts are identified by
   abbreviated descriptions; and such instructions seem to pose no
   difficulty in interpretation.  Intuitively, the hearer can identify
   the intended nut from \sref{instruction-rep-eg} because of the
   choice of verb: one of the semantic features of the verb
   \ling{slide} is the constraint that its object (here, the coupling
   nut) moves in contact along a surface to reach its destination
   (here, the elbow).  Identifying the elbow directs the hearer to the
   coupling nut on the fuel line, since that coupling nut alone lies
   along a common surface with the elbow.

   	The formal representation of communicative intent in
   Figure~\ref{instruct-intent-spud-fig} implements this explanation.
   It associates the verb \ling{slide} with proofs $K \proves \shared
   \m{surf}(P)$, $K \proves \shared \m{start-at}(P,N)$ and $K \proves
   \shared \m{end-on}(P,E)$ which together require the context to
   establish that the nut lie on a common surface with the elbow.
   Accordingly, the constraint-network model of communicative-intent
   recognition described in Section~\ref{hearer-inference-subsec} uses
   this requirement in determining candidate values for $N$ and $E$.
   The network will heuristically identify coupling nuts that lie on a
   common surface with an elbow.  In this case, the constraints
   suffice jointly to determine the arguments in the action.  Thus,
   when \spud\ constructs the communicative intent in
   Figure~\ref{instruct-intent-spud-fig}, it models and exploits an
   interaction between the microplanning tasks of referring expression
   generation and lexical choice.

	In \cite{genw-paper}, we make a similar point by analyzing the
   instruction \sref{rabbit:eg} in the context depicted in
   Figure~\ref{rabbit:fig}.
\sentence{rabbit:eg}{
	Remove the rabbit from the hat.
}
	From \sref{rabbit:eg}, the hearer should be able to identify
   the intended rabbit and the intended hat---even though the context
   supplies several rabbits, several hats, and even a rabbit in a
   bathtub and a flower in a hat.  The verb \ling{remove} presupposes
   that its object (here, the rabbit) starts out in the source (here,
   the hat), and this distinguishes the intended rabbit and hat in
   Figure~\ref{rabbit:fig} from the other ones.

	Where instructions such as \sref{instruction-rep-eg} exploit
   interactions between referring expression generation and lexical
   choice, instructions exhibiting \term{pragmatic overloading}
   exploit interactions between aggregation and lexical choice
   \cite{dieugenio/webber:overloading}.  DiEugenio and Webber
   characterize the interpretation of instructions with multiple
   clauses that describe complex actions, such as \sref{hold}.
\begin{examples}{hold}
\exitem{hold-alpha}
	Hold the cup under the spigot---
\exitem{hold-beta}
	---to fill it with coffee.
\end{examples}
	Here, the two clauses \msref{hold}{hold-alpha} and
   \msref{hold}{hold-beta} are related by enablement, a kind of
   purpose relation.  Because of this relation, the description in
   \msref{hold}{hold-beta} forms the basis of a constrained inference
   that provides additional information about the action described in
   \msref{hold}{hold-alpha}.  That is, while \msref{hold}{hold-alpha}
   itself does not specify the orientation of the cup under the
   spigot, its purpose \msref{hold}{hold-beta} can lead the hearer to
   an appropriate choice.  To fill a cup with coffee, the cup must be
   held vertically, with its concavity pointing upwards.  As noted in
   \cite{dieugenio/webber:overloading}, this inference depends on the
   information available about the action in \msref{hold}{hold-alpha}
   and its purpose in \msref{hold}{hold-beta}.
   The purpose specified in \sref{hold2} does not constrain cup
   orientation in the same way:
\sentence{hold2}{
	Hold the cup under the faucet to wash it.
}

   	In a representation of communicative intent, the pragmatic
   overloading of \sref{hold} manifests itself in an update to the
   conversational record that is achieved by inference.  Suppose that
   we represent the cup as $c1$, the action of holding it under the
   spigot as $a1$, and the needed spatial location and orientation as
   $o1$; at the same time, we may represent the filling as action
   $a2$, and the coffee as liquid $l1$.  We contribute by inference
   that the orientation is upright---$\m{upright}(o1)$---because we
   assert that $a1$ is an action where the hearer $h1$ holds $c1$ in
   $o1$---$\m{hold}(a1,h1,c1,o1)$---whose purpose is the action $a2$
   of filling $c1$ with $l1$---$\m{purpose}(a1,a2) \wedge
   \m{fill}(a2,h1,c1,l1)$; and because we count on the hearer to
   recognize that an event in which something is held to be filled
   must involve an upright orientation---in symbols:
\sentence{inference}{
	$\shared \forall ee'xcol [
	\m{hold}(e,x,c,o) \wedge
	\m{purpose}(e,e') \wedge
	\m{fill}(e',x,c,l) \imp
	\m{upright}(o)
	]$
}
	The notation of Section~\ref{rep-intro-subsec} records this
   inference as in \sref{upright-inference-tree}, a constituent of the
   communicative intent for \sref{hold}.
\sentence{upright-inference-tree}{
	\mbox{
	\leaf{$\m{hold}(a1,H,C,O)$}
	\leaf{$\m{purpose}(a1,a2)$}
	\leaf{$\m{fill}(a2,H,C,L)$}
	\branch{3}{\goalinst{\m{upright}(o1)}}
	\tree
	}
}

	Because \spud\ assesses the interpretations of utterances by
   looking for inferential possibilities such as
   \sref{upright-inference-tree}, it can recognize the textual economy
   in utterances such as \sref{hold}.  Moreover, because \spud\
   interleaves reasoning for aggregation and lexical choice (and
   referring expression generation), \spud\ can orchestrate the
   lexical content of clauses in order to take advantage of
   inferential links like that of \sref{upright-inference-tree}.

	Thus, suppose that \spud\ starts with the goal of describing
   the holding action in the main clause, describing the filling
   action, and indicating the purpose relation between them.  For the
   holding action, \spud's goals include making sure that the sentence
   communicates where the cup will be held and how it will be held
   (i.e., \m{upright}).  \spud\ first selects an appropriate
   lexico-syntactic tree for imperative \ling{hold}; \spud\ can choose
   to adjoin in the purpose clause next, in an aggregation move, and
   then to make the appropriate lexico-syntactic choice of
   \ling{fill}.  After this substitution, the semantic contributions
   of the sentence describe an action of \ling{holding an object}
   which \ling{can bring about} an action of \ling{filling that
   object}.  As shown in \cite{dieugenio/webber:overloading}, and as
   formalized in \sref{inference}, these are the premises of an
   inference that the object is held upright during the filling.  When
   \spud\ assesses the interpretation of this utterance, using logical
   queries about the updates it could achieve, it finds that the
   utterance has in fact conveyed how the cup is to be held.  \spud\
   has no reason to describe the orientation of the cup with
   additional content.

\section{Building specifications}
\label{build-sec}

   	We have seen how \spud\ plans sentences not by a modular
   pipeline of subtasks, but by general reasoning that draws on
   detailed linguistic models and a rich characterization of
   interpretation.  While this generality makes for an elegant
   uniformity in microplanning, it also poses substantial obstacles to
   the development of \spud\ specifications.  Because of \spud's
   general reasoning, changes to any lexical and syntactic entry have
   far-reaching and indirect consequences on generation results.

	In response to this challenge, we have developed a methodology
   for constructing lexicalized grammatical resources for generation
   systems such as \spud.  Our methodology involves guidelines for the
   construction of syntactic structures, for semantic representations
   and for the interface between them.  In this section, we describe
   this methodology in detail, and show, by reference to a case study
   in a specific instruction-generation domain, how this methodology
   helps ensure that \spud\ deploys its lexical and syntactic options
   as observed in a corpus of desired output.  In the future, we hope
   that this methodology can serve as a starting point for automatic
   techniques of specification development and validation from
   possibly paired corpora of syntactic and semantic
   representations------a problem that has begun to draw attention
   from the perspective of interpretation as well
   \cite{hockenmaier:jlac01}.

	The basic principle behind all of our guidelines is this:
   \textsc{The representation of a grammatical entry must make it as
   easy as possible for the generator to exploit its contribution in
   carrying out further planning.}  This principle responds to two
   concerns.  First, \spud\ is currently constrained to greedy or
   incremental search for reasons of efficiency.  At each step, \spud\
   picks the entry whose interpretation goes furthest towards
   achieving its communicative goals.  As the generator uses its
   grammar to build on these greedy choices, our principle facilitates
   the generator in arriving at a satisfactory overall utterance.
   More generally, we saw in Section~\ref{example-section} many
   characteristic uses of language in which separate lexico-syntactic
   elements jointly ensure needed features of communicative intent.
   This is an important way in which any generator needs to be able
   exploit the contribution of an entry it has already used, in line
   with our principle.
   
\subsection{Syntax}   
\label{syntax-section}
	
	Our first set of guidelines describes the elementary trees
   that we specify as syntactic structures for lexical items
   (including lexical items that involve a semantically-opaque
   combination of words).
\begin{enumerate}
\item
	The grammar must associate each item with its observed range
   of complements and modifiers, in the observed orders.  This
   constraint is common to any effort in grammar development; it is
   sufficiently well-understood to allow induction of LTAGs from
   treebanks \cite{chen/vijay:pt00,sarkar:naacl01}.
\item
	All syntactically optional elements, regardless of
   interpretation, must be represented in the syntax as modifiers,
   using the LTAG operation of adjunction.  This allows the generator
   to select an optional element when it is needed to achieve updates
   not otherwise conveyed by its provisional utterance.  Recall that,
   in LTAG, a substitution site indicates a constituent that must be
   supplied syntactically to obtain a grammatical sentence; we call a
   constituent so provided a \textsc{syntactic argument}.  The
   alternative is to rewrite a node so as to include additional
   material (generally optional) specified by an auxiliary tree; we
   call material so provided a \textsc{syntactic adjunct}.  If
   optional elements are represented as syntactic adjuncts, it is
   straightforward to select one whenever its potential benefit is
   recognized.  With other representations---for example, having a set
   of syntactic entries, each of which has a different number of
   syntactic arguments---the representation can result in artificial
   dependencies in the search space in generation, or even dead-end
   states in which the grammar does not offer a way to more precisely
   specify an ambiguous reference.  To use this representation
   successfully, a greedy generator such as \spud\ would have to
   anticipate how the sentence would be fleshed out later in order to
   select the right entry early on.
\item
	The desired linear surface order of complements and modifiers
   for an entry must be represented using hierarchies of nodes in its
   elementary tree.  In constructions with fixed word-order (the
   typical case for English), the nodes we add reflect different
   semantic classes which tend to be realized in a particular order.
   In constructions with free word-order (the typical case in many
   other languages), node-ordering would instead reflect the
   information-structure status of constituents.  Introducing
   hierarchies of nodes to encode linear surface order decouples the
   generator's search space of derivations from the overt output
   word-order.  It allows the generator to select complements and
   modifiers in any search order, while still realizing the
   complements and modifiers with their correct surface order.  This
   is important for \spud's greedy search; alternative
   designs---representing word-order in the derivation itself or in
   features that clash when elements appear in the wrong
   order---introduce dependencies into the search space for generation
   that make it more difficult for the generator to build on its
   earlier choices successfully.  However, for a generator which
   explores multiple search paths, the more flexible search space will
   offer more than one path to the same final structure, and
   additional checks will be required to avoid duplicate results.
\end{enumerate}
	Because of strong parallels in natural language syntax across
   categories (see for example \cite{jackendoff:xbar}), we anticipate
   that these guidelines apply for all constructions in a similar way.
   Here we will illustrate them with verbs, a challenging first case
   that we have investigated in detail; other categories, particularly
   complex adjectives, adverbials and discourse connectives, merit
   further investigation.

	We collected occurrences of the verbs \emph{slide},
   \emph{rotate}, \emph{push}, \emph{pull}, \emph{lift},
   \emph{connect}, \emph{disconnect}, \emph{remove}, \emph{position}
   and \emph{place} in the maintenance manual for the fuel system of
   the American F16 aircraft.  In this manual, each operation is
   described consistently and precisely.  Syntactic analysis of
   instructions in the corpus and the application of standard tests
   allowed us to cluster the uses of these verbs into four syntactic
   classes; these classes are consistent with each verb's membership
   in a distinct Levin class \cite{levin:verbs}.  Differences among
   these classes include whether the verb lexicalizes a path of motion
   (\emph{rotate}), a resulting location (\emph{position}), or a
   change of state (\emph{disconnect}); and whether a spatial
   complement is optional (as with the verbs just given) or obligatory
   (\emph{place}).  The sentences from our corpus in
   \sref{syntax-examples} illustrate these alternatives.
\begin{examples}{syntax-examples}
\item
	Rotate valve \underline{one-fourth turn clockwise}.  [Path]
\item
	Rotate halon tube to provide access.  [Path, unspecified]
\item
	Position one fire extinguisher \underline{near aircraft
   servicing connection point}.  [Resulting location]
\item
	Position drain tube.  [Resulting location, unspecified]
\item
	Disconnect generator set cable \underline{from ground power
   receptacle}.  [Change of state, specified source]
\item
	Disconnect coupling.  [Change of state, unspecified source]
\item 
	Place grommet \underline{on test set vacuum adapter}.
   [Resulting location, required]
\end{examples}

	We used our guidelines to craft \spud\ syntactic entries for
   these verbs.  For example, we associate \emph{slide} with the tree
   in \sref{tree}.  The structure reflects the optionality of the
   \emph{path} constituent and makes explicit the observed
   characteristic order of three kinds of modifiers: those specifying
   path, such as \ling{onto elbow}, which adjoin at
   $\term{vp}_{\ling{path}}$; those specifying duration, such as
   \ling{until it is released}, which adjoin at
   $\term{vp}_{\ling{dur}}$; and those specifying purpose, such as
   \ling{to uncover sealing ring}, which adjoin at
   $\term{vp}_{\ling{purp}}$.
\sentence{tree}{
\mbox{
\leaf{{\sc np}}
\leaf{{\sc v}$\Diamond$1}
\leaf{$\term{np}\downarrow$}
\branch{2}{$\term{vp}_{\m{path}}$}
\branch{1}{$\term{vp}_{\m{dur}}$}
\branch{1}{$\term{vp}_{\m{purp}}$}
\branch{2}{$\term{s}$}
\tree}	
}
	The requirements of generation in \spud\ induce certain
   differences between our trees and other LTAG grammars for English,
   such as the XTAG grammar \cite{doran:xtag,xtag}, even in cases when
   the XTAG trees do describe our corpus.  For example, the XTAG
   grammar represents \ling{slide} simply as in \sref{xtag-tree}.
\sentence{xtag-tree}{
\mbox{
\leaf{{\sc np}}
\leaf{{\sc v}$\Diamond$1}
\leaf{$\term{np}\downarrow$}
\branch{2}{$\term{vp}$}
\branch{2}{$\term{s}$}
\tree}	
}
	The XTAG grammar does not attempt to encode the different
   orders of modifiers, nor to assign any special status to path
   \textsc{pp}s with motion verbs.

\subsection{Semantic Arguments and Compositional Semantics}
\label{interface-section}

	Recall that, to express the semantic links between multiple
   entries in a derivation, we associate each node in a syntactic tree
   with indices representing individuals.  When one tree combines with
   another, and a node in one tree is identified with a node in the
   other tree, the corresponding indices are unified.  Thus, the
   central problem of designing the compositional semantics for a
   given entry is to decide which referents to explicitly represent in
   the tree and how to distribute those referents as indices across
   the different nodes in the tree.  (Of course, these decisions also
   inform subsequent specification of lexical semantics.)

	We refer to the collection of all indices that label nodes in
   an entry as the \textsc{semantic arguments} of the entry.  This
   notion of semantic argument is clearly distinguished from the
   notion of syntactic argument that we used in Section
   \ref{syntax-section} to characterize the syntactic structure of
   entries.  Each syntactic argument position corresponds to one
   semantic argument (or more), since the syntactic argument position
   is a node in the tree which is associated with some indices:
   semantic arguments.  However, semantic arguments need not be
   associated with syntactic argument positions.  For example, in a
   verb entry, we do not have a substitution site that realizes the
   eventuality that the verb describes.  But we treat this eventuality
   as a semantic argument to implement a Davidsonian account of event
   modifiers, cf.~\cite{davidson:lf/action}.  Because we count these
   implicit and unexpressed referents as semantic arguments, our
   notion is broader than that of \cite{candito/kahane:tag+98a} and is
   more similar to Palmer's essential arguments \cite{palmer90}.

	Our strategy for specifying semantic arguments is as follows.
   We always include at least one implicit argument that the structure
   as a whole describes; these are the \textsc{major arguments} of the
   structure.  (This is common in linguistics,
   e.g. \cite{jackendoff:structures}, and in computational
   linguistics, e.g. \cite{joshi:1999}.)  Moreover, since complements
   require semantic arguments, we have found the treatment of
   complements relatively straightforward---we simply introduce
   appropriate arguments.
   
	The treatment of optional constituents, however, is more
   problematic, and requires special guidelines.  Often, it seems that
   we might express the semantic relationship between a head $h$ and a
   modifier $m$ in two ways, as schematized in \sref{sem-arg-alts}.
\begin{examples}{sem-arg-alts}
\exitem{sem-arg-yes}
	$h(R,A) \wedge m(A)$
\exitem{sem-arg-no}
	$h(R) \wedge m(R,A)$
\end{examples}
	In \msref{sem-arg-alts}{sem-arg-yes}, we represent the head as
   relating its major argument $R$ to another semantic argument $A$;
   we interpret the modifier $m$ as specifying $A$ further.  In this
   case, we must provide $A$ as an index at the node where $m$
   adjoins.  In contrast, in \msref{sem-arg-alts}{sem-arg-no}, we
   interpret the modifier $m$ as relating the major argument $R$ of
   the head directly to $A$.  In this case, $A$ need not be a semantic
   argument of $h$, and we need only provide $R$ as an index at the
   node where $m$ adjoins.

	We treat the case \msref{sem-arg-alts}{sem-arg-no} as a
   default, and we require specific distributional evidence before we
   adopt a representation such as \msref{sem-arg-alts}{sem-arg-yes}.
   If a class of modifiers such as $m$ passes any of the three tests
   below, we represent the key entity $A$ as a semantic argument of
   the associated head $h$, and include $A$ as an index of the node to
   which $m$ adjoins.
\begin{enumerate}
\item
	The \textsc{presupposition test} requires us to compare the
   interpretation of a sentence with a modifier $m$, in which the head
   $h$ contributes an update, to the interpretation of a corresponding
   sentence without the modifier.  If the referent $A$ specified by
   the modifier can be identified implicitly as discourse-bound---so
   that the sentence without the modifier can have the same
   interpretation as the sentence with the modifier---then the
   modifier must specify $A$ as a semantic argument of the head $A$.
   In fact, $A$ must figure in the presupposition of $h$.  This is
   only a partial diagnostic, because semantic arguments need not
   always be presupposed.

 	\sref{disconnect} illustrates an application of the
   presupposition test for the locative modifier of the verb
   \ling{disconnect}.
\begin{examples}{disconnect}
\exitem{disconnect-explicit}
	(Find the power cable.)  Disconnect it from the power adaptor.
\exitem{disconnect-implicit}
	(The power cable is attached to the power adaptor.)  Disconnect it.
\end{examples}
	In \msref{disconnect}{disconnect-implicit}, it is understood
   that the power cable is to be disconnected \ling{from the power
   adaptor}; the modifier in \msref{disconnect}{disconnect-explicit}
   makes this explicit.  Thus \ling{disconnect} and \ling{from the
   power adaptor} pass the presupposition test.

	The motivation for the presupposition test is as follows.  In
   \spud, implicit discourse-bound references can occur in an entry
   $h$ used for an update, only when the presupposition of $h$ evokes
   a salient referent from the conversational record, as suggested by
   \cite{saeboe:presup}.  In \msref{disconnect}{disconnect-implicit},
   for example, this referent is the power adaptor and the
   presupposition is that the power cable is connected to it.  The
   representation of such presuppositions must feature a variable for
   the referent---we might have a variable $A$ for the adaptor of
   \msref{disconnect}{disconnect-implicit}.  Accordingly, in \spud's
   model of interpretation, the speaker and hearer coordinate on the
   value for this variable (that $A$ is the power adaptor, say) by
   reasoning from the presupposed constraints on the value of this
   variable.  To guarantee successful interpretation (again using
   greedy search), \spud\ needs to be able to carry out further steps
   of grammatical derivation that add additional constraints on these
   variables.  (For example, \spud\ might derive
   \msref{disconnect}{disconnect-explicit} from
   \msref{disconnect}{disconnect-implicit} by adjoining \ling{from the
   power adaptor} to describe $A$.)  But this is possible only if the
   variable is represented as a semantic argument.

\item
   	The \textsc{constituent ellipsis test} looks at the
   interpretation of cases of constituent ellipsis---certain anaphoric
   constructions that go proxy for a major argument of the head $h$.
   If modifiers in the same class as $m$ cannot be varied across
   constituent ellipsis, then these modifiers must characterize
   semantic arguments other than the major argument of $h$.

	For verbs, \ling{do so} is one case of constituent ellipsis.
   The locative PPs in \msref{do-so}{do-so-fail} pass the constituent
   ellipsis test for \ling{do so}, as they cannot be taken to describe
   Kim and Chris's separate destinations; the infinitivals in
   \msref{do-so}{do-so-pass}, which provide different reasons for Kim
   and Sandy, fail the constituent ellipsis test for \ling{do so}:
\begin{examples}{do-so}
\exitem{do-so-fail}
	\starred Kim ran quickly to the counter.  Chris did so to the
   kiosk.
\exitem{do-so-pass}
	Kim left early to avoid the crowd.  Sandy did so to find one.
\end{examples}
	A suceesful test with \ling{do so} suggests that $m$
   contributes a description of a referent that is independently
   related to the event---in other words, that $m$ specifies some
   semantic argument.  Its meaning should therefore be represented in
   the form $m(A)$.  For \msref{do-so}{do-so-fail}, for example, we
   can use a constraint $\m{to}(P,O)$ indicating that the path $P$ (a
   semantic argument of the verb) goes to the object $O$.
   
	A failed test with \ling{do so} suggests that $m$ directly
   describes a complete event.  Its meaning should therefore be
   represented in the form $m(R,A)$, where $m$ is some relational
   constraint and $R$ is an event variable.  For
   \msref{do-so}{do-so-pass}, for example, we can use the constraint
   $\m{purpose}(E,E')$, which we have already adopted to describe bare
   infinitival purpose clauses.

	A theoretical justification for the constituent ellipsis test
   depends on the assumption that material recovered from context in
   constituent ellipsis is invisible to operations of syntactic
   combination.  (For example, the material might be supplied
   atomically as discourse referent, as in \cite{hardt:dynamic/vpe},
   where \emph{do so} recovers a property or action discourse referent
   that has been introduced by an earlier predicate on events.)  Then
   a phrase that describes the major argument $R$ can combine with the
   ellipsis, but phrases that describe any another implicit referent
   $A$ cannot; these implicit referents are syntactically invisible.

\item
	The \textsc{transformation} test looks at how modifiers are
   realized across different syntactic frames for $h$; it is
   particularly useful when $m$ is headed by a closed-class item.  If
   some frames for $h$ permit $m$ to be realized as a discontinous
   constituent with an apparent ``long-distance'' dependency, then the
   modifier $m$ specifies a semantic argument.  (Note that failure of
   the transformation test would be inconclusive in cases where syntax
   independently ruled out the alternative realization.)

	For verbs, \ling{wh}-extraction constructions illustrate the
   transformation test:
\begin{examples}{extract}
\exitem{extract-pass}
	What did you remove the rabbit from?  (A: the hat)
\exitem{extract-fail}
	\starred What did you remove the rabbit at?  (A: the magic show)
\end{examples}
 	In these cases, a modifier is realized effectively in two
   parts: \ling{what...from} in \msref{extract}{extract-pass} and
   \ling{what...at} in \msref{extract}{extract-fail}.  Intuitively, we
   have a case of extraction of the \textsc{np} describing $A$ from
   within $m$.  When this is grammatical, as in
   \msref{extract}{extract-pass}, it suggests that $m$ specifies $A$
   as a semantic argument of the head; when it is not, as in
   \msref{extract}{extract-fail}, the test fails.

	In LTAG, a transformation is interpreted as a relation among
   trees in a tree family that have essentially the same meaning and
   differ only in syntax.  (In one formalization \cite{xia:tag+4},
   these relationships between trees are realized as descriptions of
   structure to add to elementary trees.)  A transformation that
   introduces the referent $A$ in the syntax--semantics interface and
   relates $A$ to the available referent $R$ in the semantics cannot
   be represented this way.  However, if some semantic argument $A$ is
   referenced in the original tree, the transformed analogue to this
   tree can easily realize $A$ differently.  If we describe the source
   location as the semantic argument $A$ in
   \msref{extract}{extract-pass} for example, the new realization
   involves an initial \emph{wh}-\textsc{np} substitution site
   describing the source $A$, and the corresponding stranded structure
   of the \textsc{pp} \emph{from t}.
\end{enumerate}
	Of course, these tests are not perfect and have on occasion
   revealed difficult or ambiguous cases; here too, further research
   remains in adapting these tests to categories of constituents that
   did not require intensive investigation in our corpus.

	We have combined these tests to designing the
   syntax--semantics interface for verbs in our generation grammar.
   In the case of \ling{slide}, these tests show that the path of
   motion is a semantic argument but a syntactic modifier.
   \sref{basic.motion.diagnostic} presents our diagnostics: extraction
   is good, \emph{do so} substitution is degraded, and \ling{slide}
   can make a presupposition about the path of motion that helps to
   identify both the object and the path.
\begin{examples}{basic.motion.diagnostic}
\item
	What did you slide the sleeve onto?
\item
	\starred Mary slid a sleeve onto the elbow and John did so
   onto the pressure sense tube.
\item
	Slide sleeve onto elbow [acceptable in a context with many
   sleeves, but only one connected on a surface with the elbow].
\end{examples}

	Suppose we describe an event $A$ in which $H$ slides object
   $O$ along path $P$.  We label the nodes of \sref{tree} with these
   indices as in \sref{tree-labels}.
\begin{examples}{tree-labels}
\item
	subject \term{np}: $H$
\item
	object \term{np}: $O$
\item
	\term{s}, $\term{vp}_{\ling{dur}}$: $A$
\exitem{purp-tree-label}
	$\term{vp}_{\ling{purp}}$: $A$, $H$
\exitem{path-tree-label}
	$\term{vp}_{\ling{path}}$: $A$, $O$, $P$
\end{examples}
	This labeling is motivated by patterns of modification we
   observed in maintenance instructions.  In particular, the index $H$
   for \msref{tree-labels}{purp-tree-label} allows us to represent the
   control requirement that the subject of the purpose clause is
   understood as the subject of the main sentence; meanwhile, the
   indices $O$ and $P$ for \msref{tree-labels}{path-tree-label} allows
   us to represent the semantics of path particles such as
   \ling{back}; \ling{back} presupposes an event or state preceding
   $A$ in time in which object $O$ was located at the endpoint of path
   $P$.

\subsection{Lexical Semantics}
\label{semantics-section}

	To complete a \spud\ specification, after following the
   methods outlined in Sections~\ref{syntax-section}
   and~\ref{interface-section}, we have only to specify the meanings
   of individual lexical items.  This task always brings potential
   difficulties.  However, the preceding decisions and the independent
   effects of \spud's specifications of content, presupposition and
   pragmatics greatly constrain what needs to be specified.

   	By specifying syntax and compositional semantics already, we
   have determined what lexicalized derivation trees the generator
   will consider; this maps out the search space for generation.
   Moreover, our strategy for doing so keeps open as many options as
   possible for extending a description of an entity we have
   introduced; it allows entries to be added incrementally to an
   incomplete sentence in any order, subject only to the constraint
   that a head must be present before we propose to modify it.
   Syntactic specifications guarantee correct word order in the
   result, while the syntax--semantics interface ensures correct
   connections among the interpretations of combined elements.  Thus,
   all that remains is to describe the communicative intent that we
   associate with the utterances in this search space.

   	The communicative intent of an utterance is made up of records
   for assertion, presupposition and pragmatics that depend on
   independent specifications from lexical items.  The content
   condition determines the generator's strategy for contributing
   needed information to the hearer; the presupposition determines,
   inter alia, reference resolution; the pragmatics determines other
   contextual links.  Thus we can consider these specifications
   separately and base each specification on clearly delineated
   evidence.  In what follows we will describe this process for the
   motion verbs we studied.

   	We begin with the content condition.  We know the kind of
   relationship that this condition must express from the verb's
   syntactic distribution (i.e., for \emph{slide}, the frames of
   \sref{syntax-examples} that lexicalize an optional path of motion),
   and from the participants in the event identified as semantic
   arguments of the verb (i.e., \emph{slide}, the event itself and its
   agent, object and path).  To identify the particular relationship,
   we consider what basic information we learn from discovering that
   an event of this type occurred in a situation where the possibility
   of this event was known.  For verbs in our domain, we found just
   four contrasts:
\begin{examples}{contrasts}
\item
	Whether the event merely involves a pure change of state,
   perhaps involving the spatial location of an object but with no
   specified path; e.g., \ling{remove} but not \ling{move}.
\item
	Whether the event must involve an agent moving an object from
   one place to another along a specified path; e.g., \ling{move} but
   not \ling{remove}.
\item
	Whether the event must involve the application of force by the
   agent; e.g., \ling{push} but not \ling{move}.
\item
	Whether the event must brought about directly through the
   agent's bodily action (and not through mechanical assistance or
   other indirect agency); e.g., \ling{place} but not \ling{position}.
\end{examples}
	Obviously, such contrasts are quite familiar from such
   research in lexical semantics as
   \cite{talmy:dynamics,jackendoff:structures}; they have also been
   explored successfully in action representation for animation
   \cite{badler:cacm99,badler:agents00}

	Many sets of verbs are identical in content by these features.
   One such set contains the verbs \ling{move}, \ling{slide},
   \ling{rotate} and \ling{turn}; these verbs contribute just that the
   event involves an agent moving an object along a given path.  Note
   that when \spud\ assesses the contribution of an utterance
   containing these verbs, it will treat the agent, object and path as
   particular discourse referents that it must and will identify.
   This is why we simply assume that the path is given in specifying
   the content condition for these verbs.  Of course, the verbs do
   provide different path information; we represent this separately,
   as a presupposition.  

	To specify the presupposition and pragmatics of a verb, we
   must characterize the links that the verbs impose between the
   action and what is known in the context about the environment in
   which the action is to be performed.  In some cases, these links
   are common across verb classes.  For instance, all motion verbs
   presuppose a current location for the object, which they assert to
   be the beginning of the path traveled.  In other cases, these links
   accompany particular lexical items; an example is the
   presupposition of \ling{slide}, that the path of motion maintains
   contact with some surface.

	In specifying these links, important evidence comes from the
   uses of lexical items observed in a corpus.  The following
   illustration is representative.  In the aircraft vent system, pipes
   may be sealed together using a sleeve, which fits snugly over the
   ends of adjacent pipes, and a coupling, which snaps shut around the
   sleeve and holds it in place.  At the start of maintenance, one
   \ling{removes} the coupling and \ling{slides} the sleeve away from
   the junction between the pipes.  Afterwards, one
   \mbox{\ling{(re-)positions}} the sleeve at the junction and
   \mbox{\ling{(re-)installs}} the coupling around it.  In the F16
   corpus, these actions are always described using these verbs.

	This use of verbs reflects not only the motions themselves but
   also the general design and function of the equipment.  For
   example, the verb \ling{position} is used to describe a motion that
   leaves its object in some definite location in which the object
   will be able to perform some intended function.  In the case of the
   sleeve, it would only be \textsc{in position} when straddling the
   pipes whose junction it seals.  Identifying such distinctions in a
   corpus thus points to the specification required for correct
   lexical choice.  In this case, we represent \emph{position} as
   presupposing some ``position'' where the object carries out its
   intended function.

	These specifications now directly control how \spud\ realizes
   the alternation.  To start, \spud's strategy of linking the
   presupposition and pragmatics to a knowledge base of shared
   information restricts what verbs are applicable in any
   microplanning task.  For example, when the sleeve is moved away
   from the junction, we can only describe it by \ling{slide} and not
   by \ling{position}, because the presupposition of \ling{position}
   is not met.

	At the same time, in contexts which support the presupposition
   and pragmatics of several alternatives, \spud\ selects among them
   based on the contribution to communicative intent of presupposition
   and pragmatics.  We can illustrate this with \ling{slide} and
   \ling{position}.  We can settle on a syntactic tree for each verb
   that best fits the context; and we have designed these trees so
   that either choice can be fleshed out by further constituents into
   a satisfactory utterance.  Similarly, these items are alike in that
   their assertions both specify the motion that the instruction must
   convey to the hearer.%
\dspftnt{
	Note that if the assertions were different in some relevant
   respect, the difference would provide a decisive reason for \spud\
   to prefer one entry over another.  \spud's top priority is to
   achieve its updates.  For example, \spud\ would prefer an entry if
   its assertion achieved a specified update by describing manner of
   motion and alternative entries did not.
}	
	The syntax, the syntax--semantics interface, and the assertion
   put \emph{slide} and \emph{position} on an equal footing, and only
   the presupposition and pragmatics could distinguish the two.

   	With differences in presuppositions come differences in
   possible resolutions of discourse anaphors to discourse referents;
   the differences depend on the properties of salient objects in the
   common ground.  The fewer resolutions that there are after
   selecting a verb, the more the verb assists the hearer in
   identifying the needed action.  This gives a reason to prefer one
   verb over another.  In general, we elect to specify a constraint on
   context as a presupposition exactly when we must model its effects
   on reference resolution.  

	In our example, general background indicates that each sleeve
   only has a single place where it belongs, at the joint; meanwhile,
   there may be many ``way points'' along the pipe to slide the sleeve
   to.  This makes the anaphoric interpretation of \emph{position}
   less ambiguous than that of \emph{slide}; to obtain an equally
   constrained interpretation with \emph{slide}, an additional
   identifying modifier like \emph{into its position} would be needed.
   This favors \emph{position} over \emph{slide}---exactly what we
   observe in our corpus of instructions.  The example illustrates how
   \spud's meaning specifications can be developed step by step, with
   a close connection between the semantic distinctions we introduce
   in lexical entries and their consequences for generation.

   	With differences in pragmatics come differences in the fit
   between utterance and context.  The more specific the pragmatics
   the better the fit; this gives another reason to prefer one verb
   over another.  We did not find such cases among the motion verbs we
   studied, because the contextual links we identified all had effects
   on reference resolution and thus were specified as presuppositions.
   However, we anticipate that pragmatics will prove important when
   differences in meaning involve the perspective taken by the speaker
   on an event, as in the contrast of \ling{buy} and \ling{sell}.

	Appendix~\ref{motion-sample} details our results for the ten
   verbs we studied; \sref{slide} presents the final sample entry for
   \emph{slide}.  The tree gives the syntax for one element in the
   tree family associated with \emph{slide}, with its associated
   semantic indices; the associated formulas describe the semantics of
   the entry in terms of presuppositions and assertions about the
   individuals referenced in the tree.
\begin{examples}{slide}
\item Syntax and syntax--semantics interface: 
\begin{center}
\mbox{\small
	\leaf{$\term{np}(H)$}
	\leaf{\term{v}$\Diamond 1$}
	\leaf{$\term{np}(O)\downarrow$}
	\branch{2}{$\term{vp}_{\m{path}}(A,O,P)$}
	\branch{1}{$\term{vp}_{\m{dur}}(A)$}
	\branch{1}{$\term{vp}_{\m{purp}}(A,H)$}
	\branch{2}{$\term{s}(A)$}
	\tree
}	
\end{center}
\item Assertion:
	$\m{move}(A,H,O,P)$
\item Presupposition: 
	$\m{start-at}(P,O) \wedge \m{surf}(P)$
\end{examples}
	Of course, the corresponding entries \sref{syntax-1} and
   \sref{lex-1} that we used in assembling concrete communicative
   intent for \sref{instruction-eg} in
   Figure~\ref{instruct-intent-spud-fig} refine \sref{slide} only in
   adopting the specific syntactic and semantic refinements of an
   imperative use of the verb.  The entries are provided as
   \sref{syntax-1} and \sref{lex-1} in Appendix~\ref{grammar-sample}.

\section{Previous Work}
\label{related-work-section}

   	In the discussion so far, we have been able to contrast
   \spud\ with a range of research from the sentence planning
   literature.  As first observed in Section~\ref{spud-intro-subsec}
   and substantiated subsequently, \spud's representations and
   algorithms, and the specification strategies they afford, greatly
   improve on prior proposals for communicative-intent--based
   microplanning such as
   \cite{appelt:planning,thomason/hobbs:abduction}.  Meanwhile, as
   catalogued in Section~\ref{example-section}, \spud\ captures the
   essence of techniques for referring expression generation, such as
   \cite{dale/haddock:referring}; for syntactic choice, such as
   \cite{mathiessen:nigel,yang:systemic/tag}; for lexical choice, such
   as \cite{nogier/zock:91,elhadad:constraints,stede:cl98}; and for
   aggregation, such as \cite{dalianis:phd,shaw:inlg98}.  

	At the same time, \spud\ goes beyond these pipelined
   approaches in modeling and exploiting interactions among
   microplanning subtasks, and \spud\ captures these efficiencies
   using a uniform model of communicative intent.  In contrast, other
   research has succeeded in capturing particular descriptive
   efficiencies only by specialized mechanisms.  For example, Appelt's
   planning formalism includes plan-critics that can detect and
   collapse redundancies in sentence plans \cite{appelt:planning}.
   This framework treats subproblems in generation as independent by
   default; and writing tractable and general critics is hampered by
   the absence of abstractions like those used in \spud\ to
   simultaneously model the syntax and the interpretation of a whole
   sentence.  Meanwhile, in \cite{mcdonald:redundancy}, McDonald
   considers descriptions of events in domains which impose strong
   constraints on what information about events is semantically
   relevant.  He shows that such material should and can be omitted,
   if it is both syntactically optional and inferentially derivable:
\begin{quote}
{\sc fairchild} Corporation (Chantilly VA) Donald E Miller was named
senior vice president and general counsel, succeeding Dominic A Petito,
who resigned in November, at this aerospace business. Mr. Miller, 43 years
old, was previously principal attorney for Temkin \& Miller Ltd.,
Providence RI.
\end{quote}
        Here, McDonald points out that one does not need to explicitly
   mention the position that Petito resigned from in specifying the
   resignation sub-event, since it must be the same as the one that
   Miller has been appointed to.  Whereas McDonald adopts
   special-purpose module to handle this, we regard it as a special
   case of pragmatic overloading.
 
	More generally, like many sentence planners, \spud\ achieves a
   flexible association between the content input to a sentence
   planner and the meaning that comes out.  Other researchers
   \cite{Nicolov-etal,rubinoff:choice} have assumed that this
   flexibility comes from a mismatch between input content and
   grammatical options.  In \spud, such differences arise from the
   referential requirements and inferential opportunities that are
   encountered.

        Previous authors \cite{McD-Pust85,joshi87-gen} have noted that
   TAG has many advantages for generation as a syntactic formalism,
   because of its localization of argument
   structure. \cite{joshi87-gen} states that adjunction is a powerful
   tool for elaborating descriptions.  These aspects of TAGs are
   crucial for us; for example, lexicalization allows us to easily
   specify local semantic and pragmatic constraints imposed by the
   lexical item in a particular syntactic frame.

        Various efforts at using TAG for
   generation~\cite{McD-Pust85,joshi87-gen,yang:systemic/tag,danlos:gtag,Nicolov-etal,wahlster:wip}
   enjoy many of these advantages.  They vary in the organization of
   the linguistic resources, the input semantics and how they evaluate
   and assemble alternatives.  Furthermore,
   \cite{shieber/et/al:semantic-head,shieber:uniform,prevost/steedman:generate,hoffman:inlg}
   exploit similar benefits of lexicalization and localization.  Our
   approach is distinguished by its declarative synthesis of a
   representation of communicative intent, which allows \spud\ to
   construct a sentence and its interpretation simultaneously.

\section{Conclusion}

       Most generation systems pipeline pragmatic, semantic, lexical
   and syntactic decisions~\cite{reiter:consensus}.  With the right
   formalism---an explicit, declarative representation of
   \textsc{communicative intent}---it is easier and better to
   construct pragmatics, semantics and syntax simultaneously.  The
   approach elegantly captures the interaction between pragmatic and
   syntactic constraints on descriptions in a sentence, and the
   inferential interactions between multiple descriptions in a
   sentence.  At the same time, it exploits linguistically motivated,
   declarative specifications of the discourse functions of syntactic
   constructions to make contextually appropriate syntactic choices.

   	Realizing a microplanner based on communicative intent
   involves challenges in implementation and specification.  In the
   past \cite{appelt:planning}, these challenges may have made
   communicative-intent--based microplanning seem hopeless and
   intractable.  Nevertheless, in this paper, we have described an
   effective implementation, \spud, that constructs representations of
   communicative intent through top-down LTAG derivation,
   logic-programming and constraint-satisfaction models of
   interpretation, and greedy search; and we have described a
   systematic, step-by-step methodology for designing generation
   grammars for \spud.

	With these results, the challenges that remain for the program
   of microplanning based on communicative intent offer fertile ground
   for further research.  \spud's model of interpretation omits
   important features of natural language, such as plurality
   \cite{stone:inlg00}, discourse connectivity \cite{webber:acl99} and
   such defeasible aspects of interpretation as
   presupposition-accommodation \cite{lewis:accommodation}.  \spud's
   search procedure is simplistic, and is vulnerable to stalled states
   where lookahead is required to recognize the descriptive effect of
   a combination of lexical items.  \cite{gardent/striegnitz:iwcs4}
   illustrate how refinements in \spud's models of interpretation and
   search can lead to interesting new possibilities for NLG.  At the
   same time, the construction of lexicalized grammars for generation
   with effective representations of semantics calls out for
   automation, using techniques that make lighter demands on
   developers and make better use of machine learning.

\section*{Acknowledgments}

	This paper has been brewing for a long time---during which the
   investigation has been supported by: NSF and IRCS graduate
   fellowships, a RuCCS postdoctoral fellowship, NSF grant NSF-STC SBR
   8920230, NSF Resarch Instrumentation award 9818322, ARPA grant
   N6600194C6-043, and ARO grant DAAH04-94-G0426.  We are grateful for
   discussion and comments from Gann Bierner, Betty Birner, Julie
   Bourne, Justine Cassell, Aravind Joshi, Alistair Knott, Ellen
   Prince, Owen Rambow, Joseph Rosenzweig, Anoop Sarkar, Mark
   Steedman, Mike White and Hao Yan.

\bibliographystyle{apalike}
\bibliography{/fac/u/mdstone/info/biblio}

\begin{thebibliography}{}

\bibitem[Appelt, 1985]{appelt:planning}
Appelt, D. (1985).
\newblock {\em Planning English Sentences}.
\newblock Cambridge University Press, Cambridge England.

\bibitem[Bach, 1989]{bach:lectures}
Bach, E. (1989).
\newblock {\em Informal Lectures on Formal Semantics}.
\newblock State University of New York Press, Albany, NY.

\bibitem[Badler et~al., 2000]{badler:agents00}
Badler, N., Bindiganavale, R., Allbeck, J., Schuler, W., Zhao, L., Lee, S.-J.,
  Shin, H., and Palmer, M. (2000).
\newblock Parameterized action representation and natural language instructions
  for dynamic behavior modification of embodied agents.
\newblock In {\em Agents}.

\bibitem[Badler et~al., 1999]{badler:cacm99}
Badler, N., Palmer, M., and Bindiganavale, R. (1999).
\newblock Animation control for real-time virtual humans.
\newblock {\em Communications of the {ACM}}, 42(7):65--73.

\bibitem[Baldoni et~al., 1998]{baldoni:jlc98}
Baldoni, M., Giordano, L., and Martelli, A. (1998).
\newblock A modal extension of logic programming: Modularity, beliefs and
  hypothetical reasoning.
\newblock {\em Journal of Logic and Computation}, 8(5):597--635.

\bibitem[Birner, 1992]{birner:thesis}
Birner, B. (1992).
\newblock {\em The Discourse Function of Inversion in English}.
\newblock PhD thesis, Northwestern University.

\bibitem[Bourne, 1998]{bourne:phd}
Bourne, J. (1998).
\newblock Generating effective instructions: Knowing when to stop.
\newblock PhD Thesis Proposal, Department of Computer \& Information Science,
  University of Pennsylvania.

\bibitem[Bratman, 1987]{bratman:book}
Bratman, M.~E. (1987).
\newblock {\em Intention, Plans, and Practical Reason}.
\newblock Harvard University Press, Cambrdige, MA.

\bibitem[Cahill and Reape, 1999]{cahill/reape:99-05}
Cahill, L. and Reape, M. (1999).
\newblock Component tasks in applied {NLG} systems.
\newblock Technical Report ITRI-99-05, {ITRI}, University of Brighton.

\bibitem[Candito and Kahane, 1998]{candito/kahane:tag+98a}
Candito, M. and Kahane, S. (1998).
\newblock Can the {TAG} derivation tree represent a semantic graph? an answer
  in the light of {Meaning-Text Theory}.
\newblock In {\em {TAG+4}}.

\bibitem[Carberry and Lambert, 1999]{carberry/lambert:cl99}
Carberry, S. and Lambert, L. (1999).
\newblock A process model for recognizing communicative acts and modeling
  negotiation subdialogues.
\newblock {\em Computational Linguistics}, 25:1--53.

\bibitem[Cassell, 2000]{cassell:cacm}
Cassell, J. (2000).
\newblock Embodied conversational interface agents.
\newblock {\em Communications of the {ACM}}, 43(4):70--78.

\bibitem[Cassell et~al., 2000]{cassell:inlg00}
Cassell, J., Stone, M., and Yan, H. (2000).
\newblock Coordination and context-dependence in the generation of embodied
  conversation.
\newblock In {\em First International Confernence on Natural Language
  Generation}, pages 171--178.

\bibitem[Chen and Vijay-Shanker, 2000]{chen/vijay:pt00}
Chen, J. and Vijay-Shanker, K. (2000).
\newblock Automated extraction of {TAGs} from the {Penn} treebank.
\newblock In {\em Proceedings of the 6th International Workshop on Parsing
  Technologies}, Trento, Italy.

\bibitem[Cheng and Mellish, 2000]{cheng/mellish:inlg00}
Cheng, H. and Mellish, C. (2000).
\newblock An empirical analysis of constructing non-restrictive {NP} components
  to express semantic relations.
\newblock In {\em First International Conference on Natural Language
  Generation}, pages 108--115.

\bibitem[Clark, 1996]{clark:book}
Clark, H.~H. (1996).
\newblock {\em Using Language}.
\newblock Cambridge University Press, Cambridge, UK.

\bibitem[Clark and Marshall, 1981]{clark/marshall:mutual}
Clark, H.~H. and Marshall, C.~R. (1981).
\newblock Definite reference and mutual knowledge.
\newblock In Joshi, A.~K., Webber, B.~L., and Sag, I., editors, {\em Elements
  of Discourse Understanding}, pages 10--63. Cambridge University Press,
  Cambridge.

\bibitem[Dale, 1992]{dale:expression/book}
Dale, R. (1992).
\newblock {\em Generating Referring Expressions: Constructing Descriptions in a
  Domain of Objects and Processes}.
\newblock MIT Press, Cambridge MA.

\bibitem[Dale and Haddock, 1991]{dale/haddock:referring}
Dale, R. and Haddock, N. (1991).
\newblock Content determination in the generation of referring expressions.
\newblock {\em Computational Intelligence}, 7(4):252--265.

\bibitem[Dale and Reiter, 1995]{dale/reiter:cogsci}
Dale, R. and Reiter, E. (1995).
\newblock Computational interpretations of the {Gricean} maxims in the
  generation of referring expressions.
\newblock {\em Cognitive Science}, 18:233--263.

\bibitem[Dalianis, 1996]{dalianis:phd}
Dalianis, H. (1996).
\newblock {\em Concise Natural Language Generation from Formal Specifications}.
\newblock PhD thesis, Royal Institute of Technology, Stockholm.
\newblock Department of Computer and Systems Sciences.

\bibitem[Danlos, 1996]{danlos:gtag}
Danlos, L. (1996).
\newblock {G-TAG: A formalism for Text Generation inspired from Tree Adjoining
  Grammar: TAG issues}.
\newblock Unpublished manuscript, TALANA, Universit\'e Paris 7.

\bibitem[Davidson, 1980]{davidson:lf/action}
Davidson, D. (1980).
\newblock The logical form of action sentences.
\newblock In {\em Essays on actions and events}, pages 105--148. Clarendon
  Press, Oxford.

\bibitem[Debart et~al., 1992]{debart:modal/lp}
Debart, F., Enjalbert, P., and Lescot, M. (1992).
\newblock Multimodal logic programming using equational and order-sorted logic.
\newblock {\em Theoretical Computer Science}, 105:141--166.

\bibitem[{Di Eugenio} and Webber, 1996]{dieugenio/webber:overloading}
{Di Eugenio}, B. and Webber, B. (1996).
\newblock Pragmatic overloading in natural language instructions.
\newblock {\em Internationl Journal of Expert Systems}, 9(2):53--84.

\bibitem[Doran et~al., 1994]{doran:xtag}
Doran, C., Egedi, D., Hockey, B.~A., Srinivas, B., and Zaidel, M. (1994).
\newblock {XTAG System} - a wide coverage grammar for {English}.
\newblock In {\em Proceedings of {COLING}}.

\bibitem[Elhadad et~al., 1997]{elhadad:constraints}
Elhadad, M., McKeown, K., and Robin, J. (1997).
\newblock Floating constraints in lexical choice.
\newblock {\em Computational Linguistics}, 23(2):195--240.

\bibitem[Elhadad and Robin, 1992]{elhadad/robin:fuf}
Elhadad, M. and Robin, J. (1992).
\newblock Controlling content realization with functional unification grammars.
\newblock In Dale, R., Hovy, E., {R\"osner}, D., and Stock, O., editors, {\em
  Aspects of Automated Natural Language Generation: 6th International Workshop
  on Natural Language Generation}, Lecture Notes in Artificial Intelligence
  587, pages 89--104. Springer Verlag, Berlin.

\bibitem[{Fari\~nas del Cerro}, 1986]{cerro:molog}
{Fari\~nas del Cerro}, L. (1986).
\newblock {MOLOG}: A system that extends {PROLOG} with modal logic.
\newblock {\em New Generation Computing}, 4:35--50.

\bibitem[Gardent and Striegnitz, 2001]{gardent/striegnitz:iwcs4}
Gardent, C. and Striegnitz, K. (2001).
\newblock Generating indirect anaphora.
\newblock In {\em Proceedings of IWCS}, pages 138--155.

\bibitem[Grice, 1957]{grice:meaning}
Grice, H.~P. (1957).
\newblock Meaning.
\newblock {\em The Philosophical Review}, 66:377--388.

\bibitem[Gundel et~al., 1993]{gundel/hz:referring}
Gundel, J.~K., Hedberg, N., and Zacharski, R. (1993).
\newblock Cognitive status and the form of referring expressions in discourse.
\newblock {\em Language}, 69(2):274--307.

\bibitem[Haddock, 1989]{haddock:thesis}
Haddock, N. (1989).
\newblock {\em Incremental Semantics and Interactive Syntactic Processing}.
\newblock PhD thesis, University of Edinburgh.

\bibitem[Hardt, 1999]{hardt:dynamic/vpe}
Hardt, D. (1999).
\newblock Dynamic interpretation of verb phrase ellipsis.
\newblock {\em Linguistics and Philosophy}, 22(2):187--221.

\bibitem[Hart et~al., 1968]{hart:astar}
Hart, P.~E., Nilsson, N.~J., and Raphael, B. (1968).
\newblock A formal basis for the heuristic determination of minimum cost paths.
\newblock {\em {IEEE} Transactions on {SSC}}, 4:100--107.
\newblock Correction in \emph{{SIGART} Newsletter} 37:28--29.

\bibitem[Heeman and Hirst, 1995]{heeman/hirst:collab}
Heeman, P. and Hirst, G. (1995).
\newblock Collaborating on referring expressions.
\newblock {\em Computational Linguistics}, 21(3).

\bibitem[Hirschberg, 1985]{hirschberg:thesis}
Hirschberg, J. (1985).
\newblock {\em A Theory of Scalar Implicature}.
\newblock PhD thesis, University of Pennsylvania.

\bibitem[Hobbs et~al., 1993]{hobbs:abduction/journal}
Hobbs, J., Stickel, M., Appelt, D., and Martin, P. (1993).
\newblock Interpretation as abduction.
\newblock {\em Artificial Intelligence}, 63:69--142.

\bibitem[Hobbs, 1985]{hobbs:promiscuity}
Hobbs, J.~R. (1985).
\newblock Ontological promiscuity.
\newblock In {\em Proceedings of ACL}, pages 61--69.

\bibitem[Hobbs et~al., 1988]{hsam:abduction}
Hobbs, J.~R., Stickel, M., Appelt, D., and Martin, P. (1988).
\newblock Interpretation as abduction.
\newblock In {\em Proceedings of ACL}, pages 95--103.

\bibitem[Hockenmaier et~al., 2001]{hockenmaier:jlac01}
Hockenmaier, J., Bierner, G., and Baldridge, J. (2001).
\newblock Providing robustness for a {CCG} system.
\newblock Submitted.

\bibitem[Hoffman, 1994]{hoffman:inlg}
Hoffman, B. (1994).
\newblock Generating context-appropriate word orders in {Turkish}.
\newblock In {\em Proceedings of the Seventh International Generation
  Workshop}.

\bibitem[Horacek, 1995]{horacek:referring}
Horacek, H. (1995).
\newblock More on generating referring expressions.
\newblock In {\em Proceedings of the Fifth European Workshop on Natural
  Language Generation}, pages 43--58, Leiden.

\bibitem[Jackendoff, 1977]{jackendoff:xbar}
Jackendoff, R. (1977).
\newblock {\em $\bar{X}$ Syntax: A Study of Phrase Structure}.
\newblock MIT.

\bibitem[Jackendoff, 1990]{jackendoff:structures}
Jackendoff, R.~S. (1990).
\newblock {\em Semantic Structures}.
\newblock MIT Press, Cambridge, MA.

\bibitem[Joshi and Vijay-Shanker, 1999]{joshi:1999}
Joshi, A. and Vijay-Shanker, K. (1999).
\newblock Compositional semantics with lexicalized tree-adjoining grammar
  {(LTAG)}.
\newblock In {\em International Workshop on Computational Semantics}, pages
  131--145.

\bibitem[Joshi, 1987]{joshi87-gen}
Joshi, A.~K. (1987).
\newblock The relevance of tree adjoining grammar to generation.
\newblock In Kempen, G., editor, {\em Natural Language Generation}, pages
  233--252. Martinus Nijhoff Press, Dordrecht, The Netherlands.

\bibitem[Joshi et~al., 1975]{joshi:tag}
Joshi, A.~K., Levy, L., and Takahashi, M. (1975).
\newblock Tree adjunct grammars.
\newblock {\em Journal of the Computer and System Sciences}, 10:136--163.

\bibitem[Kallmeyer and Joshi, 1999]{kallmeyer/joshi:ams99}
Kallmeyer, L. and Joshi, A. (1999).
\newblock Factoring predicate argument and scope semantics: underspecified
  semantics with {LTAG}.
\newblock In {\em 12th Amsterdam Colloquium}.

\bibitem[Kamp and Rossdeutscher, 1994]{kamp/rossdeutscher:tl94}
Kamp, H. and Rossdeutscher, A. (1994).
\newblock {DRS}-construction and lexically driven inference.
\newblock {\em Theoretical Linguistics}, 20:97--164.

\bibitem[Kittredge et~al., 1991]{kittredge:domain}
Kittredge, R., Korelsky, T., and Rambow, O. (1991).
\newblock On the need for domain communication knowledge.
\newblock {\em Computational Intelligence}, 7(4):305--314.

\bibitem[Lascarides and Asher, 1991]{lascarides:relations}
Lascarides, A. and Asher, N. (1991).
\newblock Discourse relations and defeasible knowledge.
\newblock In {\em Proceedings of ACL 29}, pages 55--62.

\bibitem[Levelt, 1989]{levelt:speaking}
Levelt, W. J.~M. (1989).
\newblock {\em Speaking}.
\newblock MIT, Cambridge, MA.

\bibitem[Levin, 1993]{levin:verbs}
Levin, B. (1993).
\newblock {\em English Verb Classes and Alternations: A preliminary
  investigation}.
\newblock University of Chicago, Chicago.

\bibitem[Lewis, 1979]{lewis:accommodation}
Lewis, D. (1979).
\newblock Scorekeeping in a language game.
\newblock In {\em Semantics from Different Points of View}, pages 172--187.
  Springer Verlag, Berlin.

\bibitem[Mackworth, 1987]{mackworth:constraint}
Mackworth, A. (1987).
\newblock {Constraint Satisfaction}.
\newblock In Shapiro, S., editor, {\em Encyclopedia of Artificial
  Intelligence}, pages 205--211. John Wiley and Sons.

\bibitem[Mathiessen, 1983]{mathiessen:nigel}
Mathiessen, C. M. I.~M. (1983).
\newblock Systemic grammar in computation: the {Nigel} case.
\newblock In {\em Proceedings of EACL}, pages 155--164.

\bibitem[McDonald, 1992]{mcdonald:redundancy}
McDonald, D. (1992).
\newblock Type-driven suppression of redundancy in the generation of
  inference-rich reports.
\newblock In Dale, R., Hovy, E., {R\"osner}, D., and Stock, O., editors, {\em
  Aspects of Automated Natural Language Generation: 6th International Workshop
  on Natural Language Generation}, Lecture Notes in Artificial Intelligence
  587, pages 73--88. Springer Verlag, Berlin.

\bibitem[McDonald and Pustejovsky, 1985]{McD-Pust85}
McDonald, D.~D. and Pustejovsky, J.~D. (1985).
\newblock {TAG}'s as a grammatical formalism for generation.
\newblock In {\em Proceedings of the 23$^{rd}$ Annual Meeting of the
  Association for Computational Linguistics}, pages 94--103, Chicago, IL.

\bibitem[Mellish et~al., 1998]{ilex:inlg98a}
Mellish, C., O'Donnell, M., Oberlander, J., and Knott, A. (1998).
\newblock An architecture for opportunistic text generation.
\newblock In {\em 9th International Workshop on Natural Language Generation}.

\bibitem[Mellish, 1985]{mellish:descriptions}
Mellish, C.~S. (1985).
\newblock {\em Computer Interpretation of Natural Language Descriptions}.
\newblock Ellis Horwood, Chichester, UK.

\bibitem[Miller et~al., 1991]{mnps:uniform}
Miller, D., Nadathur, G., Pfenning, F., and Scedrov, A. (1991).
\newblock Uniform proofs as a foundation for logic programming.
\newblock {\em Annals of Pure and Applied Logic}, 51:125--157.

\bibitem[Moore, 1994]{j.moore:phd}
Moore, J. (1994).
\newblock {\em Participating in Explanatory Dialogues}.
\newblock MIT Press, Cambridge MA.

\bibitem[Moore and Paris, 1993]{moore/paris:cl/planning}
Moore, J.~D. and Paris, C.~L. (1993).
\newblock Planning text for advisory dialogues: capturing intentional and
  rhetorical information.
\newblock {\em Computational Linguistics}, 19(4):651--695.

\bibitem[Nicolov et~al., 1995]{Nicolov-etal}
Nicolov, N., Mellish, C., and Ritchie, G. (1995).
\newblock Sentence generation from conceptual graphs.
\newblock In G.~Ellis, R.~Levinson, W.~R. and Sowa, F., editors, {\em
  Conceptual Structures: Applications, Implementation and Theory (Proceedings
  of Third International Conference on Conceptual Structures)}, pages 74--88.
  Springer.

\bibitem[Nilsson, 1971]{nilsson:ai71}
Nilsson, N. (1971).
\newblock {\em Problem-solving Methods in Artificial Intelligence}.
\newblock McGraw-Hill, New York.

\bibitem[Nogier and Zock, 1991]{nogier/zock:91}
Nogier, J. and Zock, M. (1991).
\newblock Lexical choice as pattern-matching.
\newblock In Nagle, T., Nagle, J., Gerholz, L., and Eklund, P., editors, {\em
  Current Directions in Conceptual Structures Research}. Springer.

\bibitem[Palmer, 1990]{palmer90}
Palmer, M. (1990).
\newblock {\em {Semantic Processing for Finite Domains}}.
\newblock Cambridge Univeristy Press.

\bibitem[Pereira and Shieber, 1987]{pereira/shieber:pnlp}
Pereira, F. C.~N. and Shieber, S.~M. (1987).
\newblock {\em Prolog and Natural Language Analysis}.
\newblock CSLI, Stanford CA.

\bibitem[Pollack, 1991]{pollack:overloading}
Pollack, M. (1991).
\newblock Overloading intentions for efficient practical reasoning.
\newblock {\em No{\^u}s}, 25:513--536.

\bibitem[Pollack, 1992]{pollack:uses}
Pollack, M.~E. (1992).
\newblock The uses of plans.
\newblock {\em Artificial Intelligence}, 57:43--68.

\bibitem[Prevost and Steedman, 1993]{prevost/steedman:generate}
Prevost, S. and Steedman, M. (1993).
\newblock Generating contextually appropriate intonation.
\newblock In {\em Proceedings of the Sixth Conference of the European Chapter
  of ACL}, pages 332--340, Utrecht.

\bibitem[Prince and Smolensky, 1997]{prince/smolensky:science97}
Prince, A. and Smolensky, P. (1997).
\newblock Optimality: From neural networks to universal grammar.
\newblock {\em Science}, 275:1604--1610.

\bibitem[Prince, 1986]{prince:open-prop}
Prince, E. (1986).
\newblock On the syntactic marking of presupposed open propositions.
\newblock In {\em Proceedings of the 22nd Annual Meeting of the Chicago
  Linguistic Society}, pages 208--222, Chicago. CLS.

\bibitem[Rambow and Korelsky, 1992]{rambow/korelsky:sp}
Rambow, O. and Korelsky, T. (1992).
\newblock Applied text generation.
\newblock In {\em Applied Natural Language Processing Conference}, pages
  40--47.

\bibitem[Reiter, 1994]{reiter:consensus}
Reiter, E. (1994).
\newblock Has a consensus {NL} generation architecture appeared, and is it
  psycholinguistically plausible?
\newblock In {\em Seventh International Workshop on Natural Language
  Generation}, pages 163--170.

\bibitem[Reiter and Dale, 2000]{reiter/dale:bnlg}
Reiter, E. and Dale, R. (2000).
\newblock {\em Building Natural Language Generation Systems}.
\newblock Cambridge.

\bibitem[Rich et~al., 2001]{rich:ai01}
Rich, C., Sidner, C.~L., and Lesh, N. (2001).
\newblock {COLLAGEN:} applying collaborative discourse theory to human-computer
  interaction.
\newblock {\em {AI} Magazine}.
\newblock to appear.

\bibitem[Rubinoff, 1992]{rubinoff:choice}
Rubinoff, R. (1992).
\newblock Integrating text planning and linguistic choice by annotating
  linguistic structures.
\newblock In Dale, R., Hovy, E., {R\"osner}, D., and Stock, O., editors, {\em
  Aspects of Automated Natural Language Generation: 6th International Workshop
  on Natural Language Generation}, Lecture Notes in Artificial Intelligence
  587, pages 45--56. Springer Verlag, Berlin.

\bibitem[Saeboe, 1996]{saeboe:presup}
Saeboe, K.~J. (1996).
\newblock Anaphoric presuppositions and zero anaphora.
\newblock {\em Linguistics and Philosophy}, 19(2):187--209.

\bibitem[Sarkar, 2001]{sarkar:naacl01}
Sarkar, A. (2001).
\newblock Applying co-training methods to statistical parsing.
\newblock In {\em Proceedings of the North Americal Association for
  Computational Linguistics}.

\bibitem[Schabes, 1990]{schabes90}
Schabes, Y. (1990).
\newblock {\em {Mathematical and Computational Aspects of Lexicalized
  Grammars}}.
\newblock PhD thesis, Computer Science Department, University of Pennsylvania.

\bibitem[Shaw, 1998]{shaw:inlg98}
Shaw, J. (1998).
\newblock Clause aggregation using linguistic knowledge.
\newblock In {\em Ninth International Workshop on Natural Language Generation},
  pages 138--148.

\bibitem[Shieber et~al., 1990]{shieber/et/al:semantic-head}
Shieber, S., van Noord, G., Pereira, F., and Moore, R. (1990).
\newblock Semantic-head-driven generation.
\newblock {\em Computational Linguistics}, 16:30--42.

\bibitem[Shieber, 1991]{shieber:uniform}
Shieber, S.~M. (1991).
\newblock A uniform architecture for parsing and generation.
\newblock In {\em ICLP}, pages 614--619.

\bibitem[Stede, 1998]{stede:cl98}
Stede, M. (1998).
\newblock A generative perspective on verb alternations.
\newblock {\em Computational Linguistics}, 24(3):401--430.

\bibitem[Steedman, 1997]{steedman:temporality}
Steedman, M. (1997).
\newblock Temporality.
\newblock In van Benthem, J. and ter Meulen, A., editors, {\em Handbook of
  Logic and Language}, pages 895--935. Elsevier.

\bibitem[Stone, 1998]{stone:phdthesis}
Stone, M. (1998).
\newblock {\em Modality in Dialogue: Planning, Pragmatics and Computation}.
\newblock PhD thesis, University of Pennsylvania.

\bibitem[Stone, 1999]{stone:modal/lp}
Stone, M. (1999).
\newblock Indefinite information in modal logic programming.
\newblock Technical Report RUCCS Report 56, Rutgers University.

\bibitem[Stone, 2000a]{stone:inlg00}
Stone, M. (2000a).
\newblock On identifying sets.
\newblock In {\em First International Confernence on Natural Language
  Generation}, pages 116--123.

\bibitem[Stone, 2000b]{jlac00}
Stone, M. (2000b).
\newblock Towards a computational account of knowledge, action and inference in
  instructions.
\newblock {\em Journal of Language and Computation}, 1:231--246.

\bibitem[Stone et~al., 2000]{stone:tag+00}
Stone, M., Bleam, T., Doran, C., and Palmer, M. (2000).
\newblock Lexicalized grammar and the description of motion events.
\newblock In {\em {TAG+:} Workshop on Tree-Adjoining Grammar and Related
  Formalisms}, pages 199--206.

\bibitem[Stone and Doran, 1996]{colloc-paper}
Stone, M. and Doran, C. (1996).
\newblock Paying heed to collocations.
\newblock In {\em International Natural Language Generation Workshop}, pages
  91--100.

\bibitem[Stone and Doran, 1997]{gen-paper}
Stone, M. and Doran, C. (1997).
\newblock Sentence planning as description using tree-adjoining grammar.
\newblock In {\em Proceedings of ACL}, pages 198--205.

\bibitem[Stone and Webber, 1998]{genw-paper}
Stone, M. and Webber, B. (1998).
\newblock Textual economy through close coupling of syntax and semantics.
\newblock In {\em Proceedings of International Natural Language Generation
  Workshop}, pages 178--187.

\bibitem[Talmy, 1988]{talmy:dynamics}
Talmy, L. (1988).
\newblock Force dynamics in language and cognition.
\newblock {\em Cognitive Science}, 12:49--100.

\bibitem[{The XTAG-Group}, 1995]{xtag}
{The XTAG-Group} (1995).
\newblock {A Lexicalized Tree Adjoining Grammar for English}.
\newblock Technical Report IRCS 95-03, University of Pennsylvania.
\newblock Updated version available at
  http://www.cis.upenn.edu/\~{}xtag/tr/tech-report.html.

\bibitem[Thomason, 1990]{thomason:intentions}
Thomason, R.~H. (1990).
\newblock Accommodation, meaning and implicature.
\newblock In Cohen, P.~R., Morgan, J., and Pollack, M.~E., editors, {\em
  Intentions in Communication}, pages 325--363. MIT Press, Cambridge, MA.

\bibitem[Thomason et~al., 1996]{thomason/hobbs/moore}
Thomason, R.~H., Hobbs, J., and Moore, J. (1996).
\newblock Communicative goals.
\newblock In {\em ECAI Workshop on Gaps and Bridges: New Directions in Planning
  and Natural Language Generation}.

\bibitem[Thomason and Hobbs, 1997]{thomason/hobbs:abduction}
Thomason, R.~H. and Hobbs, J.~R. (1997).
\newblock Interrelating interpretation and generation in an abductive
  framework.
\newblock In {\em AAAI Fall Symposium on Communicative Action}.

\bibitem[van~der Sandt, 1992]{vandersandt:anaphora}
van~der Sandt, R. (1992).
\newblock Presupposition projection as anaphora resolution.
\newblock {\em Journal of Semantics}, 9(2):333--377.

\bibitem[Vijay-Shanker, 1987]{vijay:diss}
Vijay-Shanker, K. (1987).
\newblock {\em {A Study of Tree Adjoining Grammars}}.
\newblock PhD thesis, Department of Computer and Information Science,
  University of Pennsylvania.

\bibitem[Wahlster et~al., 1991]{wahlster:wip}
Wahlster, W., {Andr\'e}, E., Bandyopadhyay, S., Graf, W., and Rist, T. (1991).
\newblock {WIP}: The coordinated generation of multimodal presentations from a
  common representation.
\newblock In Stock, O., Slack, J., and Ortony, A., editors, {\em Computational
  Theories of Communication and their Applications}. Berlin: Springer Verlag.

\bibitem[Wanner and Hovy, 1996]{wanner/hovy:inlg96}
Wanner, L. and Hovy, E. (1996).
\newblock The {HealthDoc} sentence planner.
\newblock In {\em Seventh International Workshop on Natural Language
  Generation}, pages 1--10.

\bibitem[Ward, 1985]{ward:thesis}
Ward, G. (1985).
\newblock {\em The Semantics and Pragmatics of Preposing}.
\newblock PhD thesis, University of Pennsylvania.
\newblock Published 1988 by Garland.

\bibitem[Webber et~al., 1999]{webber:acl99}
Webber, B., Knott, A., Stone, M., and Joshi, A. (1999).
\newblock Discourse relations: A structural and presuppositional account using
  lexicalised {TAG}.
\newblock In {\em Association for Computational Linguistics}, pages 41--48.

\bibitem[Webber, 1988]{webber:tense}
Webber, B.~L. (1988).
\newblock Tense as discourse anaphor.
\newblock {\em Computational Linguistics}, 14(2):61--73.

\bibitem[Webber et~al., 1998]{webber:traumaid/aij98}
Webber, B.~L., Carberry, S., Clarke, J.~R., Gertner, A., Harvey, T., Rymon, R.,
  and Washington, R. (1998).
\newblock Exploiting multiple goals and intentions in decision support for the
  management of multiple trauma: A review of the {TraumAID} project.
\newblock {\em Artificial Intelligence}, 105:263--293.

\bibitem[Xia et~al., 1998]{xia:tag+4}
Xia, F., Palmer, M., Vijay-Shanker, K., and Rosenzweig, J. (1998).
\newblock Consistent grammar development using partial tree-descriptions for
  lexicalized tree-adjoining grammars.
\newblock In {\em TAG+4}.

\bibitem[Yang et~al., 1991]{yang:systemic/tag}
Yang, G., McCoy, K.~F., and Vijay-Shanker, K. (1991).
\newblock From functional specification to syntactic structures: systemic
  grammar and tree-adjoining grammar.
\newblock {\em Computational Intelligence}, 7(4):207--219.

\end{thebibliography}

\appendix

\section{Instruction Grammar Fragment}
\label{grammar-sample}

\subsection{Syntactic Constructions}

\begin{examples}{syntax-1}
\item \term{name:} axnpVnpopp
\item \term{parameters:} $A,H,O,P,S$
\item \term{pragmatics:}
	$\m{obl}(S,H)$
\item \term{tree:}
	\mbox{\small
	\leaf{$\epsilon$}
	\branch{1}{$\term{np}(H)$}
	\leaf{\term{v}$\Diamond 1$}
	\leaf{$\term{np}(O)\downarrow$}
	\branch{2}{$\term{vp}_{\m{path}}(A,O,P)$}
	\branch{1}{$\term{vp}_{\m{dur}}(A)$}
	\branch{1}{$\term{vp}_{\m{purp}}(A,H)$}
	\branch{2}{$\term{s}(A)$}
	\tree
}	
\end{examples}

\begin{examples}{syntax-2}
\item  \term{name:} bvpPsinf
\item  \term{parameters:} $A1,H,A2$
\item \term{pragmatics:} ---
\item  \term{tree:}
\mbox{\small
	\leaf{$\term{vp}_{\m{purp}}(A1,H)*$}
	\leaf{$\term{s}_i(A2,H)\downarrow$}
	\branch{2}{$\term{vp}_{\m{purp}}(A1,H)$}
\tree
}
\end{examples}

\begin{examples}{syntax-3}
\item  \term{name:} anpxVinp
\item  \term{parameters:} $A,H,O$
\item
	\term{pragmatics:} ---
\item  \term{tree:}
\mbox{\small
	\leaf{$\epsilon$ (PRO)}
	\branch{1}{$\term{np}(H)$}
	\leaf{to}
	\leaf{$\term{v}\Diamond 1$}
	\leaf{$\term{np}(O)\downarrow$}
	\branch{2}{$\term{vp}_{\m{dur}}(A,H)$}
	\branch{1}{$\term{vp}_{\m{purp}}(A,H)$}
	\branch{2}{$\term{vp}_{\m{purp}}(A,H)$}
	\branch{2}{$\term{s}_i(A,H)$}
\tree
}
\end{examples}

\begin{examples}{syntax-4}
\item
	\term{name} zeroDefNP
\item
	\term{parameters:} $R$
\item
	\term{pragmatics:} $\m{zero-genre} \wedge \m{def}(R)$
\item
	\term{tree:}
	\mbox{\small
	\leaf{$\term{n}\Diamond$1}
	\branch{1}{$\term{n}'(R)$}
	\branch{1}{$\term{np}(R)$}
	\tree}
\end{examples}

\begin{examples}{syntax-5a}
\item
	\term{name:} bvpPnp
\item
	\term{parameters:} $E,O,P,R$
\item
	\term{pragmatics:}  $\m{zero-genre} \wedge \m{def}(R)$
\item
	\term{tree:}
\mbox{\small
	\leaf{$\term{vp}_{\m{path}}(E,O,P)\footnodemark$}
	\leaf{$\term{p}\Diamond1$}
	\leaf{$\term{np}(R)\downarrow$}
	\branch{2}{$\term{pp}(P)$}
	\branch{2}{$\term{vp}_{\m{path}}(E,O,P)$}
\tree
}
\end{examples}

\begin{examples}{syntax-6}
\item
	\term{name:} bNnn
\item
	\term{parameters:} $A,B$
\item
	\term{pragmatics:} 
	$\m{def}(A)$
\item
	\term{tree:} 
\mbox{\small
	\leaf{$\term{n}\Diamond 1$}
	\branch{1}{$\term{n}'(A)$}
	\branch{1}{$\term{np}(A)$}
	\leaf{$\term{n}'(B)\footnodemark$}
	\branch{2}{$\term{n}'(B)$}
	\tree}
\end{examples}

\subsection{Lexical Entries}

\begin{examples}{lex-1}
\item \term{name:} slide
\item \term{parameters:} $A,H,O,P,S$
\item \term{content:} $\m{move}(A,H,O,P) \wedge \m{next}(A)$
\item \term{presupposition:} $\m{start-at}(P,O) \wedge \m{surf}(P) \wedge \m{partic}(S,H)$
\item \term{pragmatics:} ---
\item \term{target:} $\term{s}(A)$ [complement]
\item \term{tree list:} axnpVnpopp$(A,H,O,P,S)$
\end{examples}

\begin{examples}{lex-2}
\item \term{name:} $\langle$purpose$\rangle$
\item \term{parameters:} $A1,H,A2$
\item \term{content:} $\m{purpose}(A1,A2)$
\item \term{presupposition:} ---
\item \term{pragmatics:} ---
\item \term{target:} $\term{vp}_2(A1,H)$ [modifier]
\item \term{tree list:} bvpPsinf$(A1,H,A2)$
\end{examples}

\begin{examples}{lex-3}
\item \term{name:} uncover
\item \term{parameters:} $A,H,O$
\item \term{content:} $\m{uncover}(A,H,O)$
\item \term{presupposition:} ---
\item \term{pragmatics:} ---
\item \term{target:} $\term{s}_i(A,H)$
\item \term{tree list:} anpxVinp$(A,H,O)$
\end{examples}

\begin{examples}{lex-4}
\item \term{name:} sealing-ring
\item \term{parameters:} $N$
\item \term{content:} $\m{sr}(N)$
\item \term{presupposition:} ---
\item \term{pragmatics:} ---
\item \term{target:} $\term{np}(N)$ [complement]
\item \term{tree list:} zerodefnptree$(N)$
\end{examples}

\begin{examples}{lex-5}
\item \term{name:} coupling-nut
\item \term{parameters:} $N$
\item \term{content:} $\m{cn}(N)$
\item \term{presupposition:} ---
\item \term{pragmatics:} ---
\item \term{target:} $\term{np}(N)$ [complement]
\item \term{tree list:} zerodefnptree$(N)$
\end{examples}

\begin{examples}{lex-6a}
\item \term{name:} onto
\item \term{parameters:} $E,O,P,R$
\item \term{content:} $\m{end-on}(P,R)$
\item \term{presupposition:} ---
\item \term{pragmatics:} ---
\item \term{target:} $\term{vp}_{\m{path}(E,O,P)}$ [modifier]
\item \term{tree list:} bvpPnp$(E,O,P,R)$
\end{examples}

\begin{examples}{lex-6}
\item \term{name:} elbow
\item \term{parameters:} $N$
\item \term{content:} $\m{el}(N)$
\item \term{presupposition:} ---
\item \term{pragmatics:} ---
\item \term{target:} $\term{np}(N)$ [complement]
\item \term{tree list:} zerodefnptree$(N)$
\end{examples}

\begin{examples}{lex-7}
\item \term{name:} fuel-line
\item \term{parameters:} $N,R,X$
\item \term{content:} $\m{fl}(N) \wedge \m{nn}(R,N,X)$
\item \term{presupposition:} ---
\item \term{pragmatics:} ---
\item \term{target:} $\term{n}'(R)$ [modifier]
\item \term{tree list:} bNnn$(N)$
\end{examples}

\section{Motion Verb Entries}
\label{motion-sample}

\subsection{Pure Motion Verbs}

	The verbs \emph{slide}, \emph{rotate}, \emph{turn},
   \emph{push}, \emph{pull}, and \emph{lift} all share a use in which
   they describe an event $A$ in which some agent $H$ moves an object
   $O$ along a path $P$.  Our analysis of this use was presented in
   detail in Section~\ref{build-sec}.  \sref{m-a-frame} gives the
   syntactic frame for this class.
\sentence{m-a-frame}{
\mbox{\small
	\leaf{$\term{np}(H)$}
	\leaf{\term{v}$\Diamond 1$}
	\leaf{{\sc np}(O)$\downarrow$}
	\branch{2}{$\term{vp}_{\m{path}}(A,O,P)$}
	\branch{1}{$\term{vp}_{\m{dur}}(A)$}
	\branch{1}{$\term{vp}_{\m{purp}}(A,H)$}
	\branch{2}{$\term{s}(A)$}
	\tree
}	
}
	Semantically, \ling{slide}, \ling{rotate} and \ling{turn} all
   assert simple motions; the verbs differ in that \ling{slide}
   presupposes motion along a surface while \ling{turn} presupposes a
   circular or helical path around an axis by which an object can
   pivot and \ling{rotate} presupposes a circular path around an axis
   through the center of an object.  \sref{move-mng-1} represents
   this.
\begin{examples}{move-mng-1}
\item
	slide: assert $\m{move}(A,H,O,P)$; presuppose
   $\m{start-at}(P,O) \wedge \m{surf}(P)$
\item
	turn: assert $\m{move}(A,H,O,P)$; presuppose
   $\m{start-at}(P,O) \wedge \m{around}(P,X) \wedge \m{pivot}(O,X)$
\item
	rotate: assert $\m{move}(A,H,O,P)$; presuppose
   $\m{start-at}(P,O) \wedge \m{around}(P,X) \wedge \m{center}(O,X)$
\end{examples}
	The verbs \ling{push}, \ling{pull} and \ling{lift} involve
   force as well as motion; they differ in presuppositions about the
   direction of force and motion: for \ling{push}, it is away from the
   agent; for \ling{pull}, it is towards the agent; \ling{lift} has an
   upward component:
\begin{examples}{move-mng-2}
\item
	push: assert $\m{forced-move}(A,H,O,P)$; presuppose
   $\m{start-at}(P,O) \wedge \m{away}(P,H)$
\item
	pull: assert $\m{forced-move}(A,H,O,P)$; presuppose
   $\m{start-at}(P,O) \wedge \m{towards}(P,H)$
\item
	lift: assert $\m{forced-move}(A,H,O,P)$; presuppose
   $\m{start-at}(P,O) \wedge \m{upwards}(P)$
\end{examples}

\subsection{Pure Change-of-state Verbs}

   	This category of verbs describes an event $A$ in which an
   agent $H$ changes of state of an object $O$; these verbs appeal to
   a single optional semantic argument $U$ which helps to specify what
   the change of state is.  Examples of this class are \emph{remove
   [from $U$]}, \emph{disconnect [from $U$]} and \emph{connect [to
   $U$]}; $U$ is a landmark object and the change-of-state involves a
   spatial or connection relation between $O$ and $U$.

	Our diagnostic tests give a number of reasons to think of the
   parameter $U$ as a semantic argument that is referenced in the tree
   but described by syntactic adjuncts.  Here are illustrations of
   these tests for the case of \ling{disconnect}.  It is possible to
   extract from it, and impossible to supply it by \ling{do so}
   substitution.
\begin{examples}{basic.cos.diagnostic}
\item
	What did you disconnect the cable from $\epsilon$?
\item
	\questioned Mary disconnected a coupling from system A, and
   John did so from system B.
\end{examples}
	It is possible to take the initial connection between $O$ and
   $U$ as presupposed, and to factor in this constraint in identifying
   $O$ and $U$.  Thus, with many systems and couplings, we might still
   find:
\sentence{basic.cos.diagnostic.2}{
	Disconnect the coupling from system A.
}
	These considerations lead to the syntactic frame of
   \sref{cos-frame}.
\sentence{cos-frame}{
\mbox{\small
	\leaf{$\term{np}(H)$}
	\leaf{\term{v}$\Diamond 1$}
	\leaf{{\sc np}(O)$\downarrow$}
	\branch{2}{$\term{vp}_{\m{arg}}(A,O,U)$}
	\branch{1}{$\term{vp}_{\m{dur}}(A)$}
	\branch{1}{$\term{vp}_{\m{purp}}(A,H)$}
	\branch{2}{$\term{s}(A)$}
	\tree
}}
   	Note that syntactic features can allow the verb to determine
   which preposition is used to specify the optional argument.  That
   is, we can use lexical entries for verbs that indicate that they
   impose feature-value constraints on the syntactic features of the
   anchor \textsc{v}$\Diamond$ node.

   	In order to characterize the semantics of change-of-state
   verbs, we introduce a predicate $\ling{caused-event}(A,H,O)$
   indicating that $A$ is an event in which $H$ has a causal effect on
   $O$; and an operator $\m{result}(A, p)$ indicating that the
   proposition $p$ holds in the state that results from doing $A$.
   (For more on this ontology, see \cite{steedman:temporality}.)
   \sref{cos-sem} uses this notation to describe \ling{connect},
   \ling{disconnect} and \ling{remove}.
\begin{examples}{cos-sem}
\item
	connect: assert $\m{caused-event}(A,H,O) \wedge
   \m{result}(A,\m{connected}(O,U))$; presuppose $\m{free}(O,U)$
\item
	disconnect: assert $\m{caused-event}(A,H,O) \wedge
   \m{result}(A,\m{free}(O,U))$; presuppose $\m{connected}(O,U)$
\item
	remove: assert $\m{caused-event}(A,H,O) \wedge
   \m{result}(A,\m{free}(O,U))$; presuppose $\m{dependent}(O,U)$
\end{examples}
   	That is, \ling{connecting} causes $O$ to be connected to the
   optional argument $U$ where $O$ is presupposed to be presently
   spatially independent of, or \ling{free} of, $U$;
   \ling{disconnecting}, conversely, causes $O$ to be \ling{free} of
   $U$, where $O$ is presupposed to be connected to $U$.  Finally,
   \ling{remove} is more general than \ling{disconnect}.  It
   presupposes only that there is some \ling{dependent} spatial
   relation between $O$ and $U$; $O$ may be attached to $U$, supported
   by $U$, contained in $U$, etc.

\subsection{Near-motion Verbs}

	Distinct from motion verbs and ordinary change-of-state verbs
   is a further class which we have called near-motion verbs:
   near-motion verbs are change-of-state verbs that encode a spatial
   change by evoking the final location where an object comes to rest.
   Semantically, they involve arguments $A$, $H$, $O$, and $L$---the
   fourth, spatial argument $L$ represents a spatial configuration
   rather than a path (as in the case of motion verbs). The canonical
   near-motion verb is \emph{position}; others are \emph{reposition}
   and \emph{install}.  According to our judgments, \emph{turn} and
   \emph{rotate} can be used as near-motion verbs as well as genuine
   motion verbs, whereas \ling{slide}, \emph{push}, \emph{pull} and
   \ling{lift} cannot.

	Now, whenever there is a change of location, there must be
   motion (in our domain); and whenever an object moves to a new
   place, there is a change of location.  This semantic correspondence
   between motion verbs and near-motion verbs is mirrored in similar
   syntactic realizations with prepositional phrases that describe an
   final location.  So we find both:
\begin{examples}{near.genuine.endpoint}
\item
	Push the coupling on the sleeve.
\item
	Position the coupling on the sleeve.
\end{examples}

	The difference between motion verbs and near-motion verbs is
   that motion verbs permit an explicit description of the \term{path}
   the object takes during the motion, while near-motion verbs do not:
\begin{examples}{near.genuine.path}
\item
	Push the coupling to the sleeve.
\item
	\starred Position the coupling to the sleeve.
\end{examples}
   
	Another way to substantiate the contrast is to consider the
   interpretation of ambiguous modifiers.  In
   \msref{downward}{push.downward}, {\em downward} modifies the path
   by describing the direction of motion in the event.  In
   \msref{downward}{position.downward}, with the near-motion verb,
   this path interpretation is not available: the reading of {\em
   downward} instead is that it describes the final orientation of the
   object that is manipulated.
\begin{examples}{downward}
\exitem{push.downward}
	Push handle downward.
\exitem{position.downward}
	Position handle downward.
\end{examples}
	These readings are paraphrased in \sref{downward.para}.
\begin{examples}{downward.para}
\item
	Push handle in a downward direction.
\item
	Position handle so that it is oriented downward.
\end{examples}
	The natural {\em wh}-questions associated with the two
   constructions are also different:
\begin{examples}{downward.wh}
\item
	$\{$ In which direction, \starsp How $\}$ did you push the
   handle?  Downward.
\item
	$\{$ \starsp In which direction, How $\}$ did you position the
   handle?  Downward.
\end{examples}

	\sref{basic.near-motion} schematizes the syntax 
   of near-motion verbs.
\sentence{basic.near-motion}{
\mbox{\small
	\leaf{$\term{np}(H)$}
	\leaf{\term{v}$\Diamond 1$}
	\leaf{{\sc np}(O)$\downarrow$}
	\branch{2}{$\term{vp}_{\m{arg}}(A,O,L)$}
	\branch{1}{$\term{vp}_{\m{dur}}(A)$}
	\branch{1}{$\term{vp}_{\m{purp}}(A,H)$}
	\branch{2}{$\term{s}(A)$}
	\tree
}}
	Like motion verbs, near-motion verbs share a common
   assertion---there is an event $A$ of $H$ acting on $O$ whose result
   is that $O$ is located at place $L$.  The differences among
   near-motion verbs lie in their presuppositions: \ling{position}
   presupposes that $L$ is a position in which $O$ will be able to
   perform its intended function, as in \msref{n-m-sem}{nmsp};
   \ling{reposition} further presupposes a state preceding $A$ where
   $O$ was located at $L$---we write this as $\m{back}(A,O,L)$ in
   \msref{n-m-sem}{nmsrp}; finally, \ling{install} presupposes that
   the spatial position for $O$ is one which fastens $O$ tightly, as
   in \msref{n-m-sem}{nmsi}.
\begin{examples}{n-m-sem}
\exitem{nmsp}
	position: assert $\m{caused-event}(A,H,O) \wedge
   \m{result}(A,\m{loc}(L,O))$; presuppose $\m{position-for}(L,O)$
\exitem{nmsrp}
	reposition: assert $\m{caused-event}(A,H,O) \wedge
   \m{result}(A,\m{loc}(L,O))$; presuppose $\m{position-for}(L,O) \wedge
   \m{back}(A,O,L)$
\exitem{nmsi}
	install: assert $\m{caused-event}(A,H,O) \wedge
   \m{result}(A,\m{loc}(L,O))$; presuppose $\m{position-for}(L,O) \wedge
   \m{fastening}(L,O)$
\end{examples}

\subsection{\ling{Put} Verbs}

	Closely related to the near-motion verbs are the \emph{put}
   verbs.  These differ from near-motion verbs only in that \emph{put}
   verbs take the configuration \textsc{pp} as a syntactic
   complement---rather than as an optional syntactic modifier.
\sentence{basic.put-verbs}{
\mbox{\small
	\leaf{$\term{np}(H)$}
	\leaf{\term{v}$\Diamond 1$}
	\leaf{{\sc np}(O)$\downarrow$}
	\leaf{{\sc pp}(L)$\downarrow$}
	\branch{3}{$\term{vp}(A)$}
	\branch{1}{$\term{vp}_{\m{dur}}(A)$}
	\branch{1}{$\term{vp}_{\m{purp}}(A,H)$}
	\branch{2}{$\term{s}(A)$}
	\tree
}}
	Verbs in this class include not only \ling{put}, but also
   \ling{place}.
\begin{examples}{put-sem}
\item
	put: assert $\m{caused-event}(A,H,O) \wedge
   \m{result}(A,\m{loc}(L,O))$
\item
	place: assert $\m{body-caused-event}(A,H,O) \wedge
   \m{result}(A,\m{loc}(L,O))$; presuppose $\m{place-for}(L,O)$
\end{examples}
	Note that a placement must be performed by hand; the
   presupposition that $L$ be a \ling{place} for $O$ signifies that
   $O$'s specific location at $L$ is required for the success of
   future actions or events.  (\ling{Place} contrasts with
   \ling{position} in that places depend on the action of an agent on
   the object in a particular activity whereas positions are enduring
   regions that depend on the functional properties of the object
   itself; contrast \ling{working place} and \ling{working position}.)
\end{document}